\documentclass[lettersize,journal]{IEEEtran}
\usepackage{amsmath,amsfonts}
\usepackage{algorithmic}
\usepackage{array}
\usepackage[caption=false,font=normalsize,labelfont=sf,textfont=sf]{subfig}
\usepackage{textcomp}
\usepackage{stfloats}
\usepackage{url}
\usepackage{verbatim}
\usepackage{graphicx}
\usepackage{booktabs}
\usepackage{hyperref}
\usepackage[table,xcdraw]{xcolor}
\usepackage{cite}
\usepackage{soul, color, xcolor}
\usepackage{tcolorbox} 
\usepackage{multirow}
\usepackage{overpic}
\usepackage{algorithm,algorithmic}
\usepackage{subfig} 
\usepackage{colortbl}  
\usepackage{xpatch}
\usepackage{balance}
\usepackage{amssymb}
\usepackage{makecell}
\usepackage{siunitx} 
\begin{document}

\title{Tree-Mamba: A Tree-Aware Mamba for Underwater Monocular Depth Estimation}

\author{

Peixian~Zhuang,~\IEEEmembership{Senior Member,~IEEE,}~
Yijian~Wang,~
Zhenqi~Fu,~
Hongliang Zhang, \\
Sam~Kwong,~\IEEEmembership{Fellow,~IEEE}, 
Chongyi~Li,~\IEEEmembership{Senior Member,~IEEE}~

\thanks{Peixian Zhuang is with the Key Laboratory of Knowledge Automation for Industrial Processes, Ministry of Education, the School of Automation and Electrical Engineering, University of Science and Technology Beijing, Beijing 100083, China (e-mail: zhuangpeixian0624@163.com).}
\thanks{Yijian Wang is with the School of Ophthalmology and Optometry, Wenzhou Medical University, Wenzhou 325000, China (e-mail: wangyijian1017@163.com).}
\thanks{Zhenqi Fu is with the Department of Automation, Tsinghua University, Beijing 100084, China (e-mail: fuzhenqi@mail.tsinghua.edu.cn).}
\thanks{Hongliang Zhang is with the Deepinfar Ocean Technology Inc., Tianjin 300450, China (e-mail: hongliang.zhang@deepinfar.com).}
\thanks{Sam Kwong is with the School of Data Science, Lingnan University, Hong Kong SAR (e-mail: samkwong@ln.edu.hk).}
\thanks{Chongyi Li is with the College of Computer Science, Nankai University, Tianjin 300350, China (e-mail: lichongyi25@gmail.com).}
\thanks{Peixian Zhuang and Yijian Wang contributed equally to this work.}
\thanks{*Corresponding authors (Chongyi Li and Peixian Zhuang).}

}

\maketitle

\begin{abstract}
Underwater Monocular Depth Estimation (UMDE) is a critical task that aims to estimate high-precision depth maps from underwater degraded images caused by light absorption and scattering effects in marine environments.
Recently, Mamba-based methods have achieved promising performance across various vision tasks; however, they struggle with the UMDE task because their inflexible state scanning strategies fail to model the structural features of underwater images effectively.
Meanwhile, existing UMDE datasets usually contain unreliable depth labels, leading to incorrect object-depth relationships between underwater images and their corresponding depth maps. 
To overcome these limitations, we develop a novel tree-aware Mamba method, dubbed Tree-Mamba, for estimating accurate monocular depth maps from underwater degraded images. 
Specifically, we propose a tree-aware scanning strategy that adaptively constructs a minimum spanning tree based on feature similarity. The spatial topological features among the tree nodes are then flexibly aggregated through bottom-up and top-down traversals, enabling stronger multi-scale feature representation capabilities.
Moreover, we construct an underwater depth estimation benchmark (called BlueDepth), which consists of 38,162 underwater image pairs with reliable depth labels. This benchmark serves as a foundational dataset for training existing deep learning-based UMDE methods to learn accurate object–depth relationships.
Extensive experiments demonstrate the superiority of the proposed Tree-Mamba over several leading methods in both qualitative results and quantitative evaluations with competitive computational efficiency. 
Code and dataset will be available at https://wyjgr.github.io/Tree-Mamba.html.
\end{abstract}

\begin{IEEEkeywords}
    Underwater image, monocular depth estimation, Mamba, minimum spanning tree, underwater benchmark.
\end{IEEEkeywords}

\vspace{-0.3cm}

\section{Introduction}
\IEEEPARstart{P}{recise} acquisition of underwater depth is essential for underwater tasks \cite{UIM1}, such as marine navigation \cite{PUDE}, 3D reconstruction \cite{UPGformer}, and localization \cite{CD-UDepth}. 
The most straightforward method for underwater depth acquisition is to use sensors, such as light detection and ranging (LiDAR) \cite{SUIM-SDA} or stereo cameras \cite{UPGformer}. 
Unfortunately, stereo cameras \cite{UPGformer} usually produce depth maps with holes due to low-textured areas and occlusions, whereas LiDAR \cite{SUIM-SDA} is affected by substantial laser scattering in water. 
These inherent limitations hinder the applications of these sensors in water. 
In contrast, UMDE offers a more efficient solution by capturing underwater images with a single camera and estimating depth maps using either traditional or deep learning-based approaches.

\begin{figure}[t]
    \centering
    \includegraphics[width=1\linewidth]{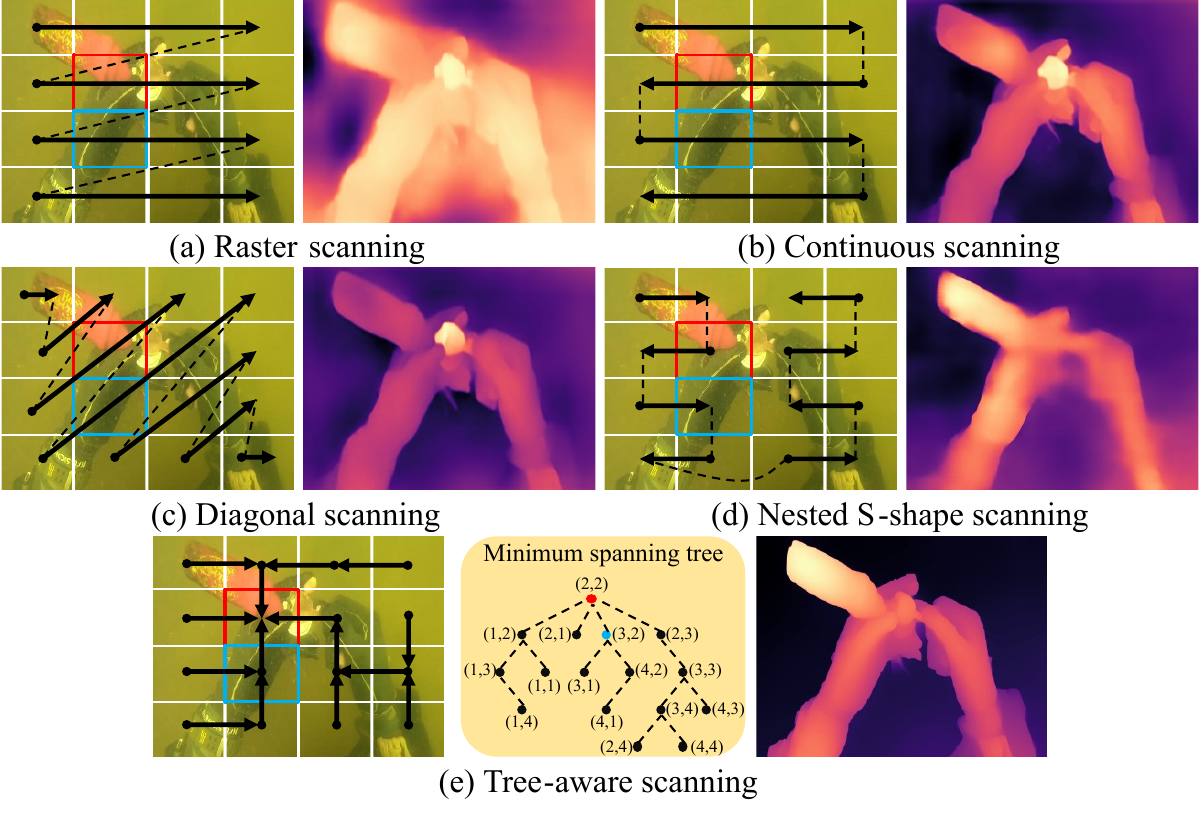}
    
    \addvspace{-10pt}  
    
    \caption{Comparison of different scanning strategies. ($i$,$j$) denotes the subgraph located at the $i$-th row and $j$-th column in an underwater image. Raster \cite{VMamba}, continuous  \cite{Zigma}, diagonal \cite{VmambaIR}, and nested S-shape \cite{MaIR} scanning strategies fail in capturing key spatial features of the image, such as the continuity from the subgraph (2,2) to the subgraph (3,2). In contrast, the proposed tree-aware scanning successfully captures this key continuity by propagating the features along a minimum spanning tree based on feature similarity. Moreover, our tree-aware approach yields better depth map with finer details and sharper geometries than the other four scanning strategies \cite{VmambaIR, MaIR, VMamba, Zigma}.}
    \label{Scans}
    
\end{figure}

\begin{figure}[t]
    \Large
    \centering
    \resizebox{1\linewidth}{!}{
        \begin{tabular}{c@{ }c@{ }c@{ }c@{ }}

            \multicolumn{4}{c}{
                \begin{tabular}{c@{\hspace{0.3em}}c}
                    \includegraphics[height=3cm,width=4cm]{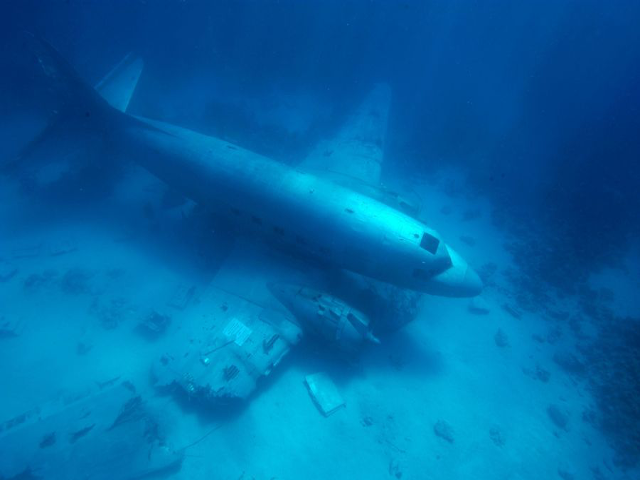} &
                    \includegraphics[height=3cm,width=4cm]{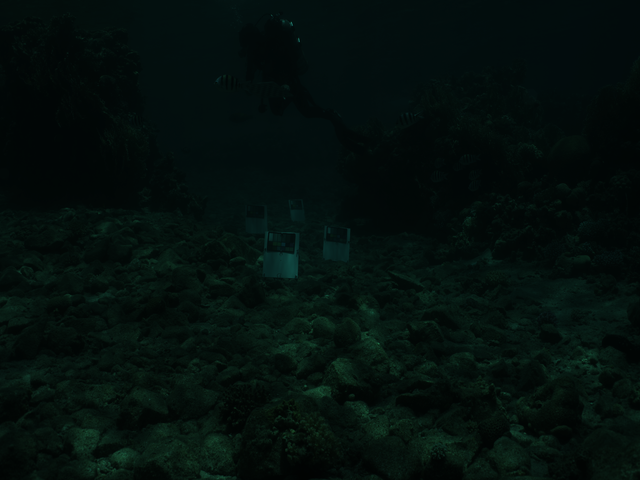} \\
                    \multicolumn{2}{c}{(a) Underwater images }
                \end{tabular}
            } \\
            
            \includegraphics[height=3cm,width=4cm]{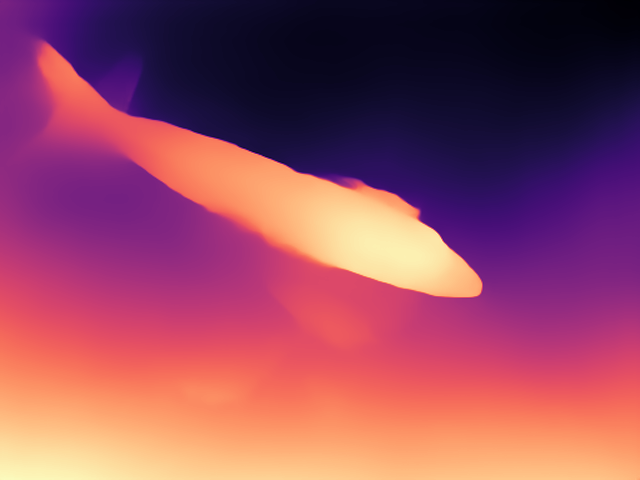} &
            \includegraphics[height=3cm,width=4cm]{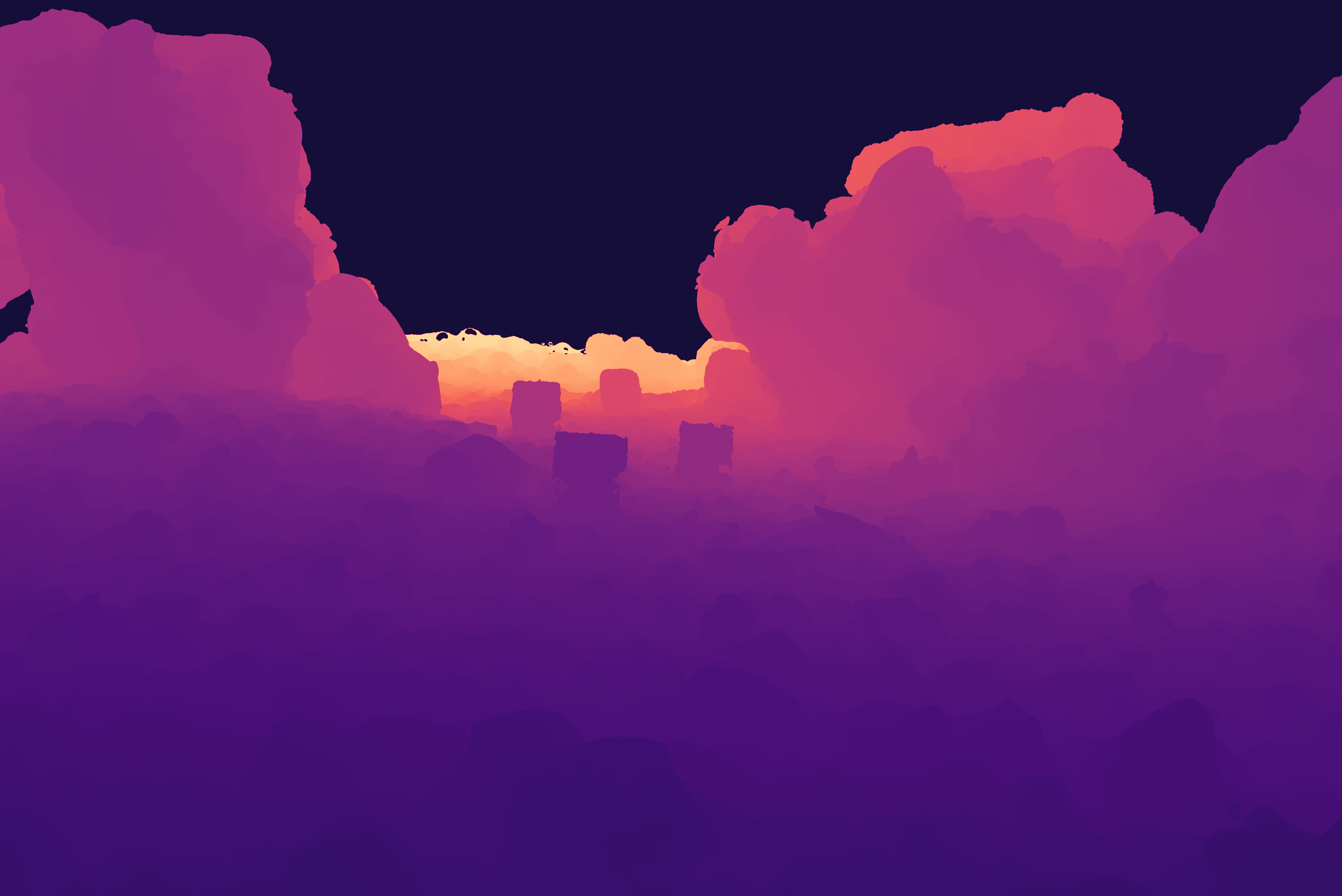} &
            \includegraphics[height=3cm,width=4cm]{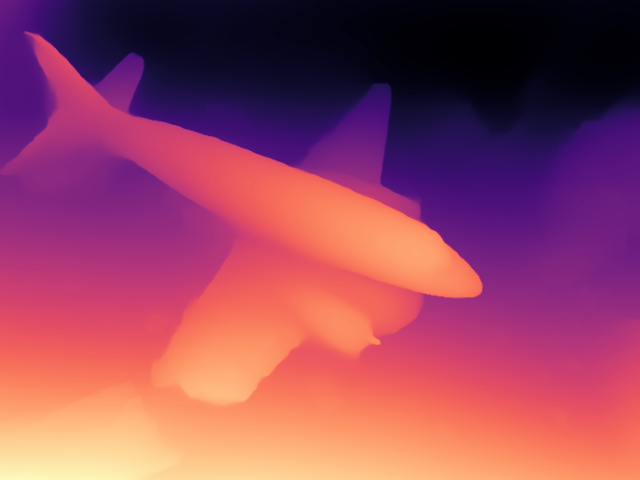} &
            \includegraphics[height=3cm,width=4cm]{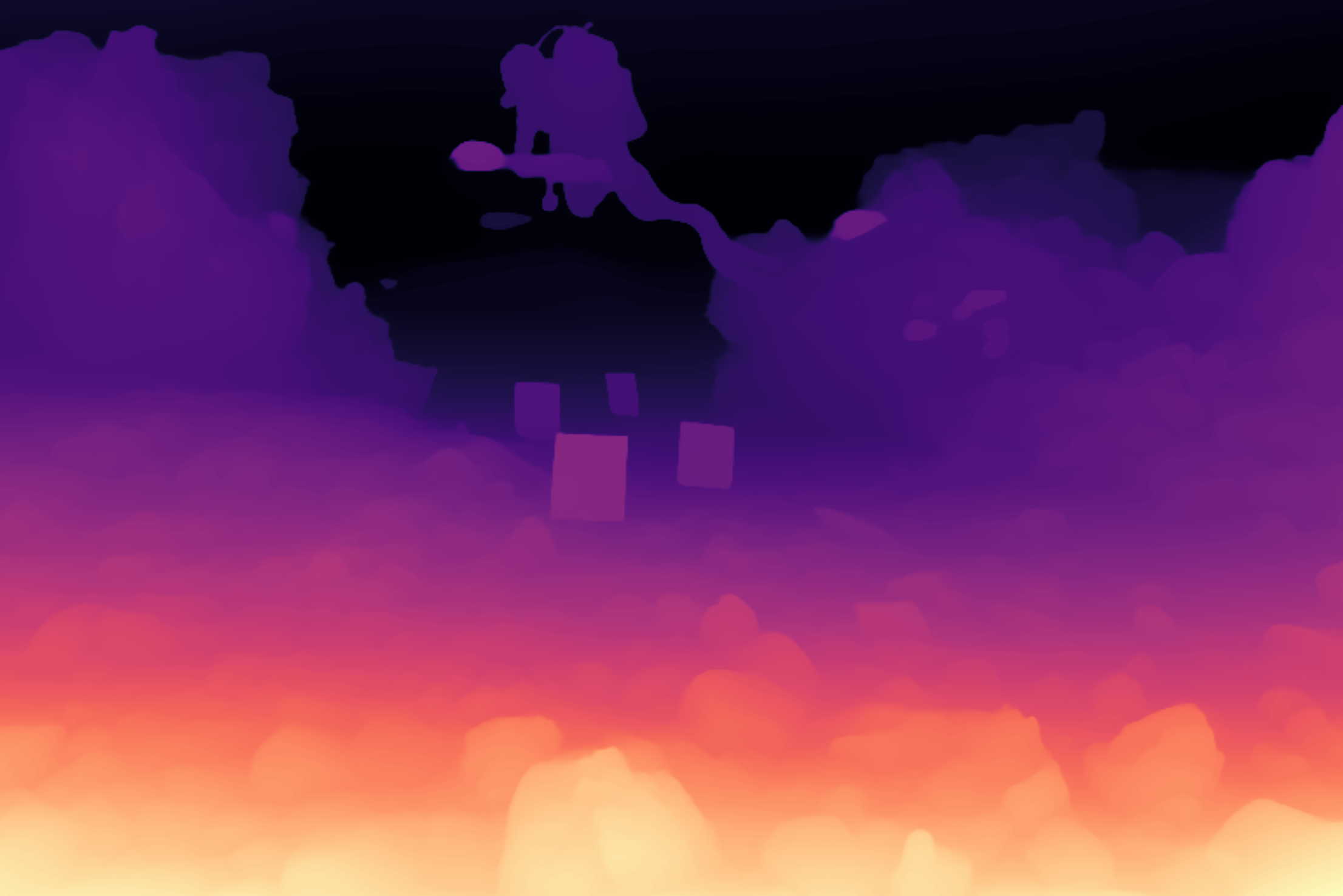} \\
            
            \multicolumn{2}{c}{(b) Depth maps (USOD10K)} & 
            \multicolumn{2}{c}{(c) Depth maps (\textcolor{red}{BlueDepth})} \\

            \includegraphics[height=3cm,width=4cm]{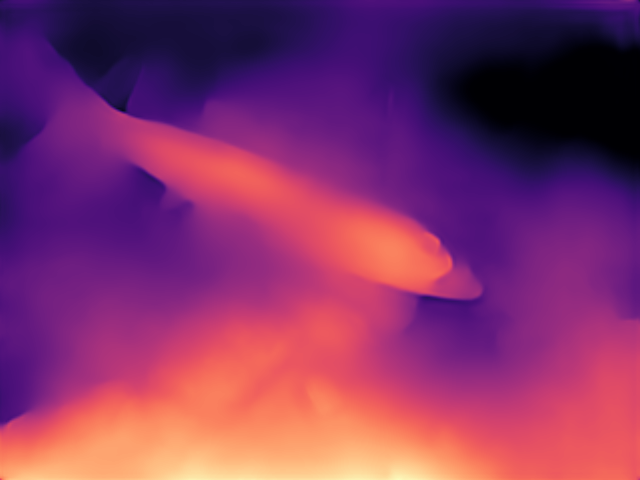} &
            \includegraphics[height=3cm,width=4cm]{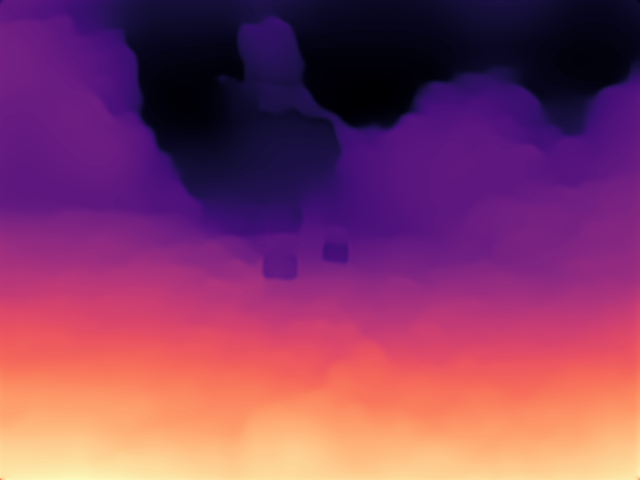} &
            \includegraphics[height=3cm,width=4cm]{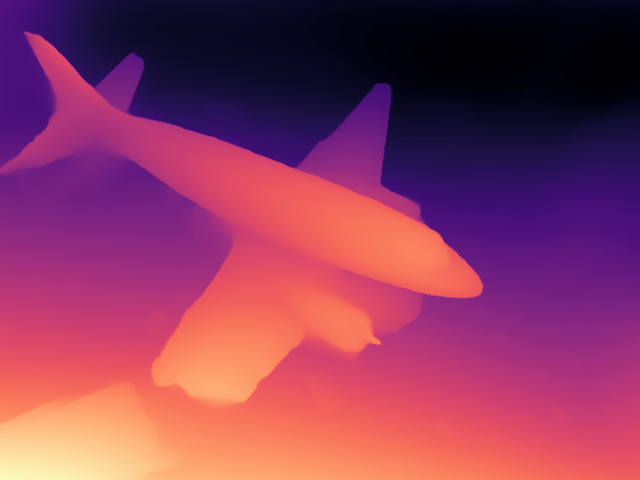} &
            \includegraphics[height=3cm,width=4cm]{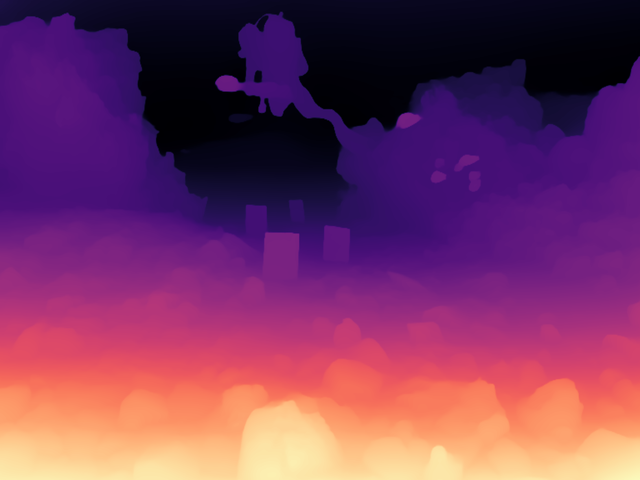} \\
            
            \multicolumn{2}{c}{(d) Tree-Mamba (USOD10K)} & 
            \multicolumn{2}{c}{(e) Tree-Mamba (\textcolor{red}{BlueDepth})} \\
                
        \end{tabular}
    }
    
    \addvspace{-2pt}  

    \caption{Visual comparison of depth maps in USOD10K \cite{USOD10K} and our BlueDepth. (b) and (c) are depth maps in USOD10K \cite{USOD10K} and our BlueDepth, respectively. (d) and (e) are the proposed Tree-Mamba trained on USOD10K \cite{USOD10K} and our BlueDepth, respectively. The depth maps estimated by our Tree-Mamba (bottom) are significantly improved by training on our BlueDepth.}
    \label{Depth Accuracy}
\end{figure}

Traditional UMDE methods \cite{IBLA, GDCP, NUDCP, HazeLine} focus on employing an underwater image formation model (UIFM) \cite{UIM1} to infer depth maps, but they generally produce imprecise results since their assumed priors do not always hold in complex marine environments \cite{UW-Net}.
Numerous deep learning-based approaches \cite{UW-Net, UW-GAN, UDepth, UW-Depth} have been presented to improve the estimation accuracy of underwater depth maps, which can be broadly categorized into CNN-based \cite{UW-Net, UW-GAN} and Transformer-based \cite{UDepth, UW-Depth} methods.
CNN-based methods \cite{UW-Net, UW-GAN} excel at learning local features of images with linear computational complexity, but lack global modeling capability due to the inherent locality of convolution \cite{Mamba}.
Transformer-based methods \cite{UDepth, UW-Depth} possess strong global modeling capability, but incur substantial computational overhead, due to the quadratic complexity of the self-attention mechanism \cite{Mamba}.
To combine both advantages of CNN and Transformer, Dao \emph{et al.} \cite{Mamba} proposed a state space model with a selective mechanism, called Mamba.
Since Mamba was originally designed for sequence modeling, a lot of studies have extended this methodology to various vision tasks, such as image segmentation \cite{VMamba}, image classification \cite{Zigma}, and image restoration \cite{VmambaIR, MaIR}.
However, these Mamba-based methods \cite{VmambaIR, MaIR, VMamba, Zigma} have been directly applied to the UMDE task, resulting in suboptimal performance. This is because their scanning strategies fail to flexibly model the structural features of underwater images. Specifically, they flatten the 2D image into 1D sequences, disrupting the original locality and continuity.
In Fig. \ref{Scans}, these fixed and inflexible scanning strategies \cite{VmambaIR, MaIR, VMamba, Zigma} do not consider the continuity between subgraph (2,2) and subgraph (3,2), whereas our tree-aware approach can capture this continuity and provide more reasonable spatial topology structures.
Moreover, our Tree-Mamba method produces depth maps with finer details and more complete content compared to other scanning strategies \cite{VmambaIR, MaIR, VMamba, Zigma}, demonstrating the superiority of the proposed tree-aware scanning strategy for UMDE.
Besides, existing UMDE benchmarks \cite{SeaThru, NYU-U, HazeLine, Flsea, Atlantis, SUIM-SDA, USOD10K} typically contain imprecise depth maps, which causes existing deep learning-based models \cite{UW-Net, UW-GAN, UDepth, UW-Depth} to learn incorrect object–depth relationships during training.
In Figs. \ref{Depth Accuracy} (b) and (c), the depth maps from our BlueDepth have finer details and sharper geometries than those in the USOD10K \cite{USOD10K}. 
Meanwhile, the proposed Tree-Mamba trained on our BlueDepth produces more accurate depth maps than its counterpart trained on USOD10K \cite{USOD10K}, as shown in Figs. \ref{Depth Accuracy} (d) and (e).

To overcome the above limitations, we first establish an underwater depth estimation benchmark (BlueDepth), which contains 38,162 underwater images with reliable pseudo-labeled depth maps.
Specifically, six leading monocular depth estimation models \cite{AdaBins, DPT, Marigold, DA1, DA2, Lotus} are employed to produce pseudo-labeled depth maps from seven existing datasets \cite{SeaThru, NYU-U, HazeLine, Flsea, Atlantis, SUIM-SDA, USOD10K}.
Then, a fine-tuned image quality assessment model \cite{LAR-IQA} is used to select the most accurate pseudo-depth map from six candidates.
When existing deep learning-based methods \cite{UW-Net, UW-GAN, UDepth, UW-Depth} use the proposed BlueDepth for training, they learn more accurate object-depth relationships of underwater scenes.
The BlueDepth serves as a comprehensive benchmark for evaluating the performance of various UMDE algorithms across diverse underwater scenes and lays a solid foundation for constructing high-quality UMDE datasets.
Based on the proposed BlueDepth dataset, we then develop a Tree-aware Mamba, called Tree-Mamba, which is composed of a CNN-based encoder and tree-aware Mamba blocks that use a tree-aware scanning strategy to model object-depth relationships of underwater scenes.
The CNN-based encoder can be any CNNs for providing rich multi-scale feature maps, while our tree-aware Mamba blocks are designed to refine these multi-scale features.
In the refinement process, the proposed tree-aware scanning strategy dynamically constructs a minimum spanning tree based on feature similarity and performs state propagation among tree nodes according to their spatial topological structure.
The main contributions of our work are summarized as follows:

\begin{itemize}
    \item  We propose an efficient Tree-Mamba method, which is the first work to tackle underwater monocular depth estimation using Mamba with a tree-aware scanning strategy. Benefiting from theoretical convergence guarantees, the proposed strategy enhances feature representation by dynamically constructing a minimum spanning tree to capture spatial topological structures and propagating multi-scale structural features between parent–child tree nodes.
    \item We establish an underwater depth estimation benchmark (dubbed BlueDepth), which contains 36,182 underwater images and corresponding high-precision depth maps covering various underwater scenes, water types, and lighting conditions. BlueDepth alleviates the scarcity of high-quality image–depth pairs for UMDE and provides a comprehensive baseline for evaluating the performance of different UMDE algorithms across diverse underwater environments.
    \item We conduct extensive experiments to demonstrate the superiority of the proposed Tree-Mamba method in both subjective results and objective assessments compared to several leading approaches. Additionally, we show that Tree-Mamba generalizes well by accurately estimating depth maps for hazy, sand-dust, and low-light underwater images.
    
\end{itemize}

\vspace{-0.3cm}

\section{Related Work}

\subsection{Underwater Monocular Depth Estimation Methods}

Existing UMDE methods can be mainly categorized into two types: traditional methods and deep learning-based approaches.
Traditional UMDE methods estimate underwater scene depth by using various manually set priors.
Peng \emph{et al.} \cite{IBLA} presented a depth estimation method based on image blurriness and light absorption priors for underwater scenes.
Based on scene ambient light differential, Peng \emph{et al.} \cite{GDCP} designed a general dark channel prior (GDCP) to estimate underwater depth maps.
Song \emph{et al.} \cite{NUDCP} introduced an underwater dark channel prior (NUDCP) to estimate depth maps based on the histogram distribution and light attenuation of underwater images.
By considering that the attenuation of red wavelengths is the fastest in underwater environments, Galdran \emph{et al.} \cite{RDCP} adopted a red channel prior for predicting underwater scene depth.
However, these manually set priors limit the model's robustness in complex and diverse underwater environments.
Accordingly, deep learning-based methods have been proposed to overcome the aforementioned limitation.
In \cite{CD-UDepth}, Guo \emph{et al.} employed a dual-source fusion framework based on color and light attenuation information, and fused dual-source features by confidence maps to achieve robust depth estimation.
Ebner \emph{et al.} \cite{UW-Depth} introduced a lightweight hybrid framework, called UW-Depth, to integrate a sparse depth prior from triangulated features to address the problem of scale ambiguity.
In \cite{SUIM-SDA}, Li \emph{et al.} presented a multi-task learning framework that integrates features from depth estimation, semantic segmentation, and edge detection to predict underwater scene depth.
Ding \emph{et al.} \cite{WaterMono} used a teacher-guided self-supervised method with a rotated distillation strategy to enhance model robustness for depth estimation. 
Although these deep learning-based methods achieve significant improvements in underwater monocular depth estimation, they struggle to balance prediction accuracy and computational efficiency. 
In contrast, our proposed Tree-Mamba is designed to deliver high-quality depth estimation with fast inference speed by combining the tree-aware Mamba block with a CNN encoder.
Moreover, our proposed tree-aware scanning strategy constructs an input-dependent minimum spanning tree and leverages the structural relationships between parent and child nodes to capture the spatial topology of underwater images, thereby enabling multi-scale feature modeling capabilities.

\vspace{-0.3cm}
\subsection{Underwater Monocular Depth Estimation Datasets}

Existing UMDE datasets can be broadly categorized into two main types: real-label datasets and pseudo-label datasets.

\emph{1) Real-label datasets}: Real-labeled datasets include depth annotations under various real marine scenes.
Li \emph{et al.} \cite{NYU-U} presented the NYU-U dataset, which included 1,449 depth labels, and each depth label contained 10 images with different types of underwater distortions.
Akkaynak \emph{et al.} \cite{SeaThru} established the Sea-Thru dataset, containing 1,205 greenish image pairs under various conditions of underwater visibility.
Berman \emph{et al.} \cite{HazeLine} introduced the SQUID dataset, involving 114 paired underwater images that covered coral, rock, and other scenes.
Randall \emph{et al.} \cite{Flsea} built the FLsea dataset, comprising 22,451 image pairs and covering diverse underwater visibility, ambient light conditions, natural and man-made structures.
The aforementioned datasets are inadequate for high-quality supervised learning due to incomplete scene content and missing or blurred details in their provided depth maps.

\emph{2) Pseudo-label datasets}: Pseudo-labeled datasets contain depth labels of both underwater finer details and complete scene content.
Zhang \emph{et al.} \cite{Atlantis} established the Atlantis dataset, consisting of 3,200 realistic underwater images and their accurate depth maps, which was generated through a pipeline that leverages the Stable Diffusion \cite{SD} and the specialized ControlNet \cite{ControlNet}.
Li \emph{et al.} \cite{SUIM-SDA} constructed the SUIM-SDA dataset that included 1,596 manually annotated sparse depth maps,
Hong \emph{et al.} \cite{USOD10K} built the USOD10K dataset of 10,255 paired underwater images, with each raw image corresponding to depth maps and salient object boundaries.
These pseudo-label datasets overcome the limitations of the aforementioned real-label datasets but introduce numerous unreliable pseudo-labeled depth maps, thereby degrading estimation performance.
To address the above issues, we built an underwater depth estimation benchmark (BlueDepth) with more precise object-depth relationships by leveraging the state-of-the-art MDE models \cite{AdaBins, DPT, Marigold, DA1, DA2, Lotus} to generate pseudo-labeled depth maps from existing seven datasets \cite{SeaThru, NYU-U, HazeLine, Flsea, Atlantis, SUIM-SDA, USOD10K}, and then employing a fine-tuned IQA model \cite{LAR-IQA} to select the finest depth maps.

\vspace{-0.3cm}
\section{Proposed BlueDepth Dataset}

This section details the proposed underwater depth estimation benchmark (BlueDepth), including data collection as well as depth map generation and selection.
\vspace{-0.3cm}
\subsection{Data Collection}
We collect underwater images from the existing seven UMDE datasets: Sea-Thru \cite{SeaThru}, NYU-U \cite{NYU-U}, SQUID \cite{HazeLine}, FLSea \cite{Flsea}, Atlantis \cite{Atlantis}, SUIM-SDA \cite{SUIM-SDA}, and USOD10K \cite{USOD10K}, totaling 53,311 underwater images with 37,470 corresponding depth maps.
The number of images in the proposed BlueDepth dataset compared to seven other UMDE datasets is summarized in Table \ref{Dataset Number}, where these datasets are described in detail in Subsection B of the related works.
As shown, due to the inclusion of two synthetic UMDE datasets (NYU-U \cite{NYU-U} and Atlantis \cite{Atlantis}), their depth maps correspond to multiple underwater images.
For the latest UMDE datasets (Atlantis \cite{Atlantis} and USOD10K \cite{USOD10K}), their depth maps are all pseudo labels, which reflects a prevailing trend toward the usage of pseudo-labeled data in the construction of UMDE datasets.
We detail the generation and selection of high-quality depth maps below.

\begin{table}[t]
    \centering
    \caption{Comparison of image quantities between the proposed BlueDepth dataset and seven existing UMDE datasets.}
    \label{Dataset Number}
    \addvspace{-6pt}  
    \resizebox{0.9\linewidth}{!}{
        \begin{tabular}{c|c|c|c|c|c}
        \hline
        \rowcolor[HTML]{FFCCC9}Datasets &Year&  \multicolumn{2}{c|}{Underwater Images}& \multicolumn{2}{c}{Depth Maps}\\
        \hline \hline   
        Sea-Thru \cite{SeaThru}&    2019&   1205&   Real&       1205&       Real\\
        NYU-U \cite{NYU-U}&         2020&   14490&  Synthetic&  1449&       Real\\
        SQUID \cite{HazeLine}&      2021&   114&    Real&       114&        Real\\
        FLSea \cite{Flsea}&         2023&   22451&  Real&       22451&      Real\\
        Atlantis \cite{Atlantis}&   2024&   3200&   Synthetic&  400&        Pseudo\\
        SUIM-SDA \cite{SUIM-SDA}&   2024&   1596&   Real&       1596&       Pseudo\\
        USOD10K \cite{USOD10K}&     2025&   10255&  Real&       10255&      Pseudo\\  \hline
        
         &  &   
        \textcolor{red}{6010}&   
        \textcolor{red}{Synthetic}&  
        \multicolumn{1}{c|}{\textcolor{red}{6010}}&  \\ \cline{3-5}

        \multirow{-2}{*}{\textcolor{red}{BlueDepth}}& 
        \multirow{-2}{*}{\textcolor{red}{2025}}&   
        \textcolor{red}{32152}&  
        \textcolor{red}{Real}&       
        \multicolumn{1}{c|}{\textcolor{red}{32152}}  & 
        \multirow{-2}{*}{\textcolor{red}{Pseudo}} \\ 
        
        \hline
        \end{tabular}
    }
\end{table}

\begin{figure*}[ht]
    \centering
    \includegraphics[width=0.8\linewidth]{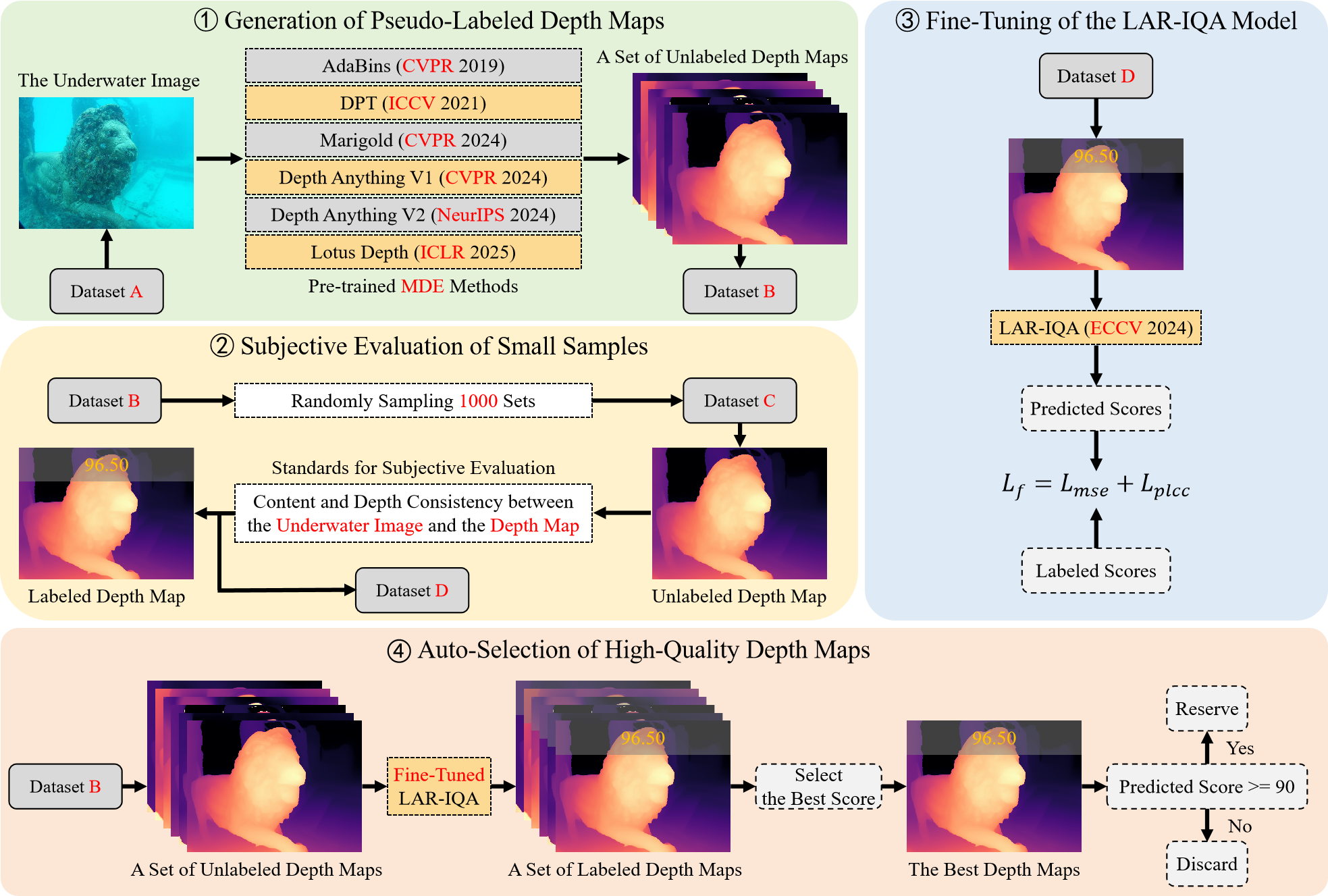}
   
    \addvspace{-8pt}  
    
    \caption{The schematic diagram of constructing our BlueDepth dataset.
    Dataset \textcolor{red}{A} contains all underwater images from the existing seven UMDE datasets \cite{SeaThru, NYU-U, HazeLine, Flsea, Atlantis, SUIM-SDA, USOD10K}, while Dataset \textcolor{red}{B} stores all pseudo-labeled depth maps for each underwater image from Dataset \textcolor{red}{A}. 
    Dataset \textcolor{red}{C} is a subset randomly selected from Dataset \textcolor{red}{B}, while Dataset \textcolor{red}{D} is its subjectively labeled counterpart. The generation and selection of high-quality depth maps includes four main steps.
    \textit{Step 1}: Six state-of-the-art MDE methods (AdaBins \cite{AdaBins}, DPT \cite{DPT}, Marigold \cite{Marigold}, Depth Anything V1 \cite{DA1}, Depth Anything V2 \cite{DA2}, and Lotus Depth \cite{Lotus}) are used to generate pseudo-labeled depth maps from Dataset \textcolor{red}{A}.
    \textit{Step 2}: The depth maps are subjectively evaluated from Dataset \textcolor{red}{C}.
    \textit{Step 3}: The LAR-IQA model \cite{LAR-IQA} is fine-tuned with Dataset \textcolor{red}{D}.
    \textit{Step 4}: The fine-tuned LAR-IQA model \cite{LAR-IQA} is employed to automatically select the best depth maps from Dataset \textcolor{red}{B}.}
\label{Flowchart of UDEB}
\end{figure*}

\vspace{-0.3cm}
\subsection{Depth Map Generation and Selection}
\label{sec_dataset_dmgs}

As illustrated in Fig. \ref{Flowchart of UDEB}, the generation and selection of high-quality depth maps includes four main steps: generation of pseudo-labeled depth maps, subjective evaluation of small samples, fine-tuning of the lightweight, accurate, and robust no-reference image quality assessment (LAR-IQA) model \cite{LAR-IQA},
and auto-selection of high-quality depth maps.

\emph{1) Generation of pseudo-labeled depth maps}:
All underwater images from the existing seven UMDE datasets \cite{SeaThru, NYU-U, HazeLine, Flsea, Atlantis, SUIM-SDA, USOD10K} are first integrated into Dataset A.
Then, a set of unlabeled depth maps for each underwater image in Dataset A is obtained via six state-of-the-art MDE methods \cite{AdaBins, DPT, Marigold, DA1, DA2, Lotus}.
Finally, all sets of unlabeled depth maps are put into Dataset B.

\begin{figure}[t]
    \centering
    \includegraphics[width=0.9\linewidth]{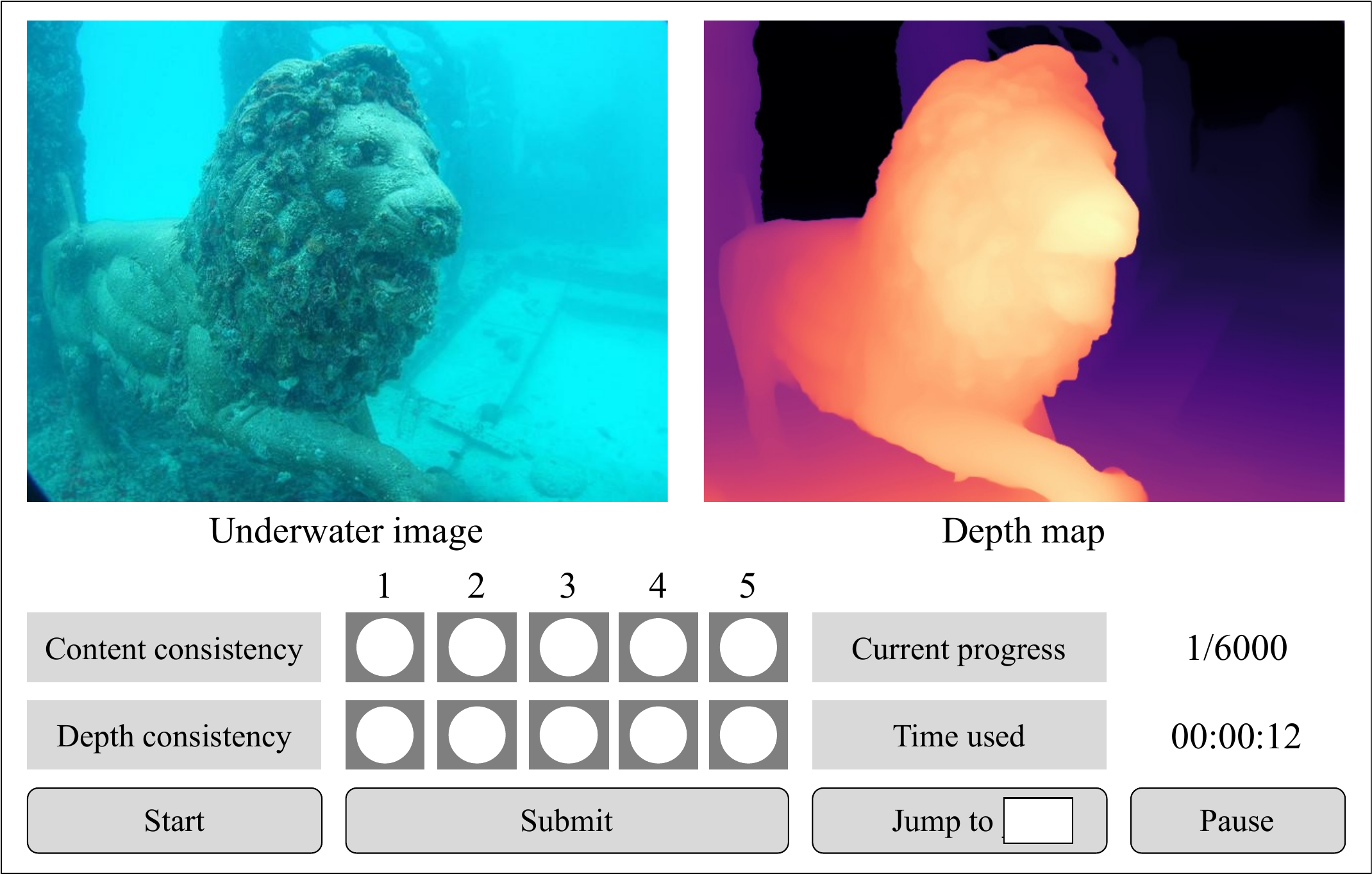}

    \addvspace{-6pt}  
    
    \caption{The scoring interface for our subjective user study.}
    \label{User Interface}
\end{figure}

\emph{2) Subjective evaluation of small samples}:
To obtain small samples used for fine-tuning the LAR-IQA model \cite{LAR-IQA}, 1,000 sets of depth maps from Dataset B are randomly sampled to form Dataset C.
Then, the subjective evaluation is conducted in a controlled laboratory setting.
We use a Dell 24-inch monitor with a resolution of 1920 $\times$ 1080 pixels for display and maintain a viewing distance of twice the monitor’s height. 
The experimental setup complies with the ITU-R BT.500-14 standard \cite{ITUR}.
Twelve expert evaluators, including six males and six females, participated in this subjective study.
Subjects scored depth maps based on the consistency of content and depth between underwater images and depth maps.
The scoring interface is illustrated in Fig. \ref{User Interface}, with an underwater image on the left and the corresponding depth map on the right.
Scoring is conducted using a five-point scale defined by the Absolute Category Rating (ACR) system of ITU-R \cite{ITUR}: 5 (Excellent), 4 (Good), 3 (Fair), 2 (Poor), and 1 (Very Poor).
This interface design enables subjects to easily assess the quality of the depth maps.
To avoid potential biases caused by visual fatigue, subjects are asked to participate in a 30-minute subjective evaluation, followed by a 10-minute rest period.
After 12 subjects scored 6,000 depth maps in two aspects, a total of 12 $\times$ 6,000 $\times$ 2 $=$ 144,000 subjective annotations are collected.
Subsequently, we detect outlier scores and remove invalid subjects by following the recommendation in \cite{ITUR}. 
Subjects with an outlier rate $\leq$ 5$\%$ are considered valid based on the 95$\%$ confidence interval criteria, and two subjects were deemed invalid and removed.
The Z-score is utilized to normalize the subjective scores $z^k_{i,j}$ in two dimensions (content $c$ and depth $d$ consistency) as follows:
\begin{equation}
z^k_{i,j}=\frac{s^k_{i,j}-\mu^k_i}{\sigma^k_i}, k\in\{c,d\},
\label{dataset_eq1}
\end{equation}
where $s_{i,j}$ denotes scores of the $i$-th subject on the $j$-th depth map.
$\mu_i$ and $\sigma_i$ represent the mean and standard deviation of the $i$-th subject’s scores for all depth maps.
Then, we mapped $z^k_{i,j}$ into the range [0, 100]:
\begin{equation}
\hat{z}^k_{i,j}=(z^k_{i,j} - z^k_{min}) \times \frac{100 - 1}{ z^k_{max} -  z^k_{min}}+ 1, k\in\{c,d\},
\label{dataset_eq2}
\end{equation}
where $z^k_{max}$ and $z^k_{min}$ denote the maximum and minimum of $z^k_{i,j}$.
Next, the final subjective score $m_j$ for the $j$-th depth map can be derived as follows:
\begin{equation}
\begin{aligned}
 m_j&=\frac{m^c_{j} + m^d_{j}}{2},\\
m^c_j&=\frac{\sum_{i=1}^{N} \hat{z}^c_{i,j}}{N},\\
m^d_j&=\frac{\sum_{i=1}^{N} \hat{z}^d_{i,j}}{N},\\
\end{aligned}
\label{dataset_eq3}
\end{equation}
where $m^c_j$ and $m^d_j$ denote final subjective scores on the content $c$ and depth $d$ consistency for the $j$-th depth map.
$N$ represents the count of subjects after exclusions.
At last, all labeled depth maps are put into Dataset D.

\begin{figure}[t]
    \centering
    \includegraphics[width=1\linewidth]{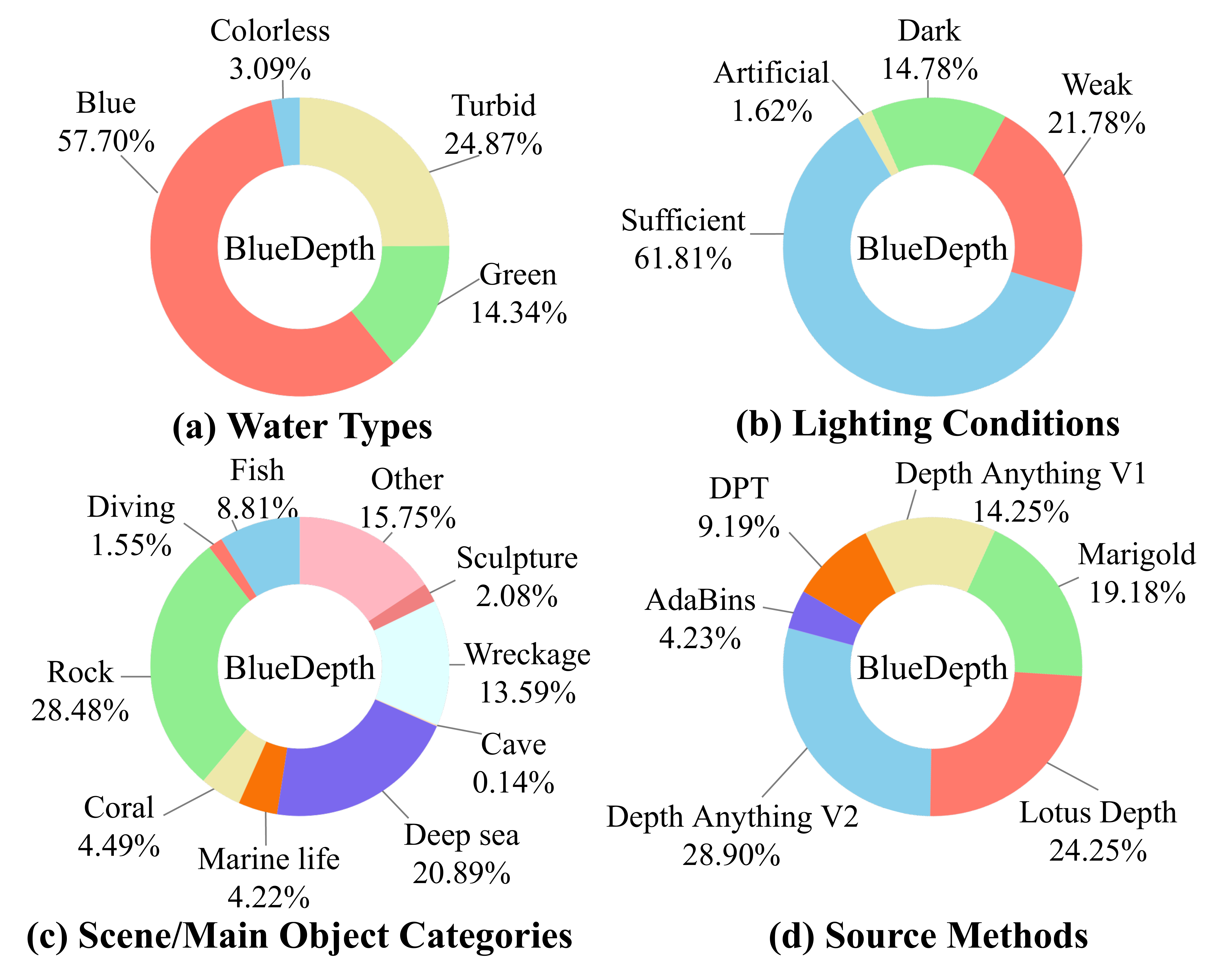}

    \addvspace{-6pt}  

    \caption{Statistics of our constructed BlueDepth. The BlueDepth dataset contains 38,162 underwater images with diverse scenes and various quality degradations.}
    \label{UDEB Pie}
\end{figure}

\emph{3) Fine-tuning of the LAR-IQA model}:
To assess no-reference depth map quality, we use Dataset D to fine-tune the LAR-IQA model \cite{LAR-IQA}.
Following the recommendation in \cite{LAR-IQA}, we employ the linear combination of the mean squared error loss $L_{mse}$ and the Pearson linear correlation coefficient loss $L_{plcc}$ as the final loss function $L_f$:
\begin{equation}
L_f=L_{mse}+L_{plcc}.
\label{dataset_eq4}
\end{equation}
Specifically,  $L_{mse}$ measures the Euclidean distance between the predicted scores $\hat{y}$ and the labeled scores $ y$:
\begin{equation}
L_{mse}=\frac{\sum_{i = 1}^{n}(\hat{y}_i - y_i)^2}{n},
\label{dataset_eq5}
\end{equation}
where $n$ denotes the number of scores. 
$L_{plcc}$ evaluates the linear correlation between $\hat{y}$ and $ y$:
\begin{equation}
L_{plcc}=1 - \frac{\sum_{i = 1}^{n}(\hat{y}_i - \bar{\hat{y}})(y_i - \bar{y})}{\sqrt{\sum_{i = 1}^{n}(\hat{y}_i - \bar{\hat{y}})^2}\sqrt{\sum_{i = 1}^{n}(y_i - \bar{y})^2}},
\label{dataset_eq6}
\end{equation}
where $\bar{\hat{y}}$ and  $\bar{y}$ denote the mean value of $\hat{y}_i$ and $y_i$, respectively.
After fine-tuning for 50 epochs, the LAR-IQA model \cite{LAR-IQA} with the minimum training loss value is retained.

\emph{4) Auto-selection of high-quality depth maps}:
The fine-tuned LAR-IQA model \cite{LAR-IQA} is adopted to score each set of unlabeled depth maps from Dataset B.
Then, the highest-scoring depth map in each set is selected to check if its score is $\ge$ 90.
If the score is $\ge$ 90, the depth map is retained; otherwise, it is discarded.
At last, 38,162 high-quality underwater image pairs are obtained.
Moreover, Fig. \ref{UDEB Pie} shows the statistics of our BlueDepth dataset, covering four types of water types: bluish (57.70\%), turbid (24.87\%), greenish (14.34\%), and colorless (3.09\%); 
four lighting conditions: sufficient (61.81\%), weak(21.78\%), dark (14.78\%), and artificial (1.62\%); 
diverse underwater scene/main object categories: rock (28.48\%), deep-sea (20.89\%), wreckage (13.59\%), other (15.75\%), fish (8.81\%), marine life (4.22\%), coral (4.49\%), sculpture (2.08\%), diving (1.55\%), cave (0.14\%);
six source methods: Depth Anything V2 \cite{DA2} (28.90\%), Lotus Depth \cite{Lotus} (24.25\%), Marigold \cite{Marigold} (19.18\%), Depth Anything V1 \cite{DA1} (14.25\%), DPT \cite{DPT} (9.19\%), and AdaBins \cite{AdaBins} (4.23\%).
This suggests that our BlueDepth contains rich underwater scene/object categories, water types, and lighting conditions.

\begin{figure*}[ht]
    \centering
    \includegraphics[width=0.98\linewidth]{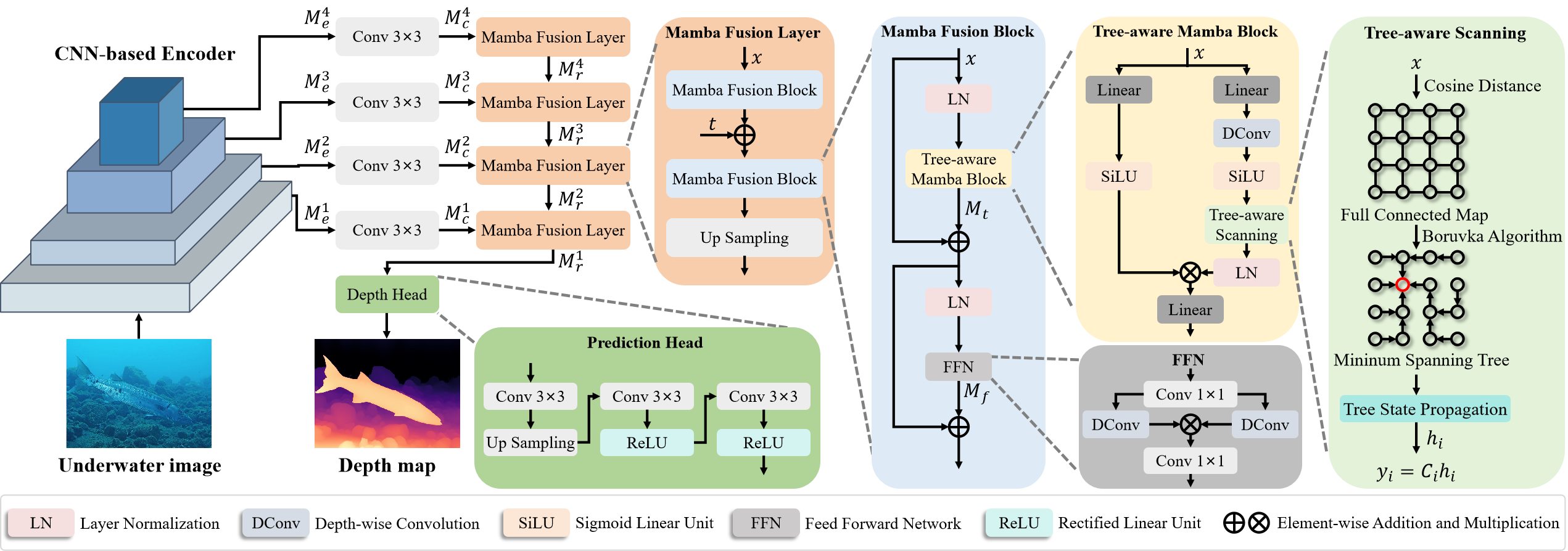}
    
    \addvspace{-6pt}  
    
    \caption{Overview of the proposed Tree-Mamba architecture. An underwater image first passes through a CNN-based encoder to produce multi-scale feature maps. These feature maps are then fed into 3$\times$3 convolutional layers to reduce the feature channels. Next, Mamba fusion layers are used to refine the feature maps, where the proposed tree-aware scanning strategy adaptively constructs a minimum spanning tree with a spatial topological structure from the input feature maps and propagates multi-scale structural features via tree state propagation. Finally, the depth map is produced through the prediction head \cite{MiDas}.}
    \label{Framework}
   
\end{figure*}

\vspace{-0.3cm}
\section{Proposed Tree-Mamba Method}

In this section, we first present the overall architecture of the proposed Tree-Mamba,
and we then detail the proposed tree-aware scanning strategy and its theoretically convergent proof, and the loss function.

\begin{figure}[!ht]
    \Large
    \centering
    \resizebox{0.9\linewidth}{!}{
        \begin{tabular}{c@{ }c@{ }c@{ }}
            \includegraphics[height=3cm,width=4cm]{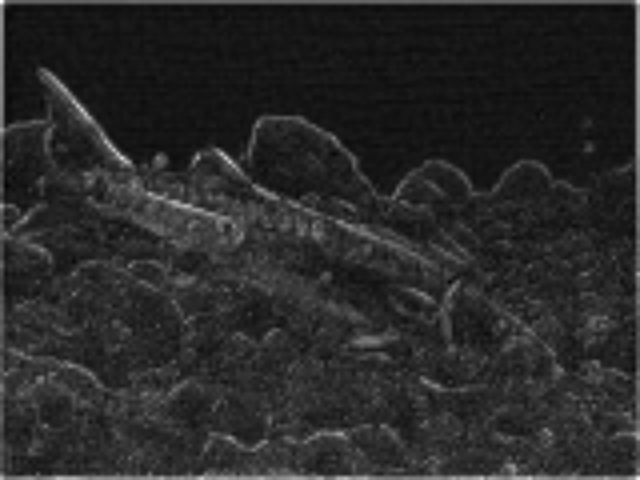} &
            \includegraphics[height=3cm,width=4cm]{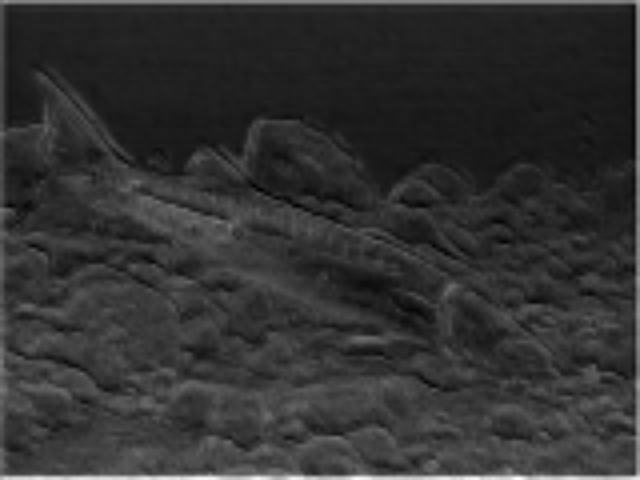} &
            \includegraphics[height=3cm,width=4cm]{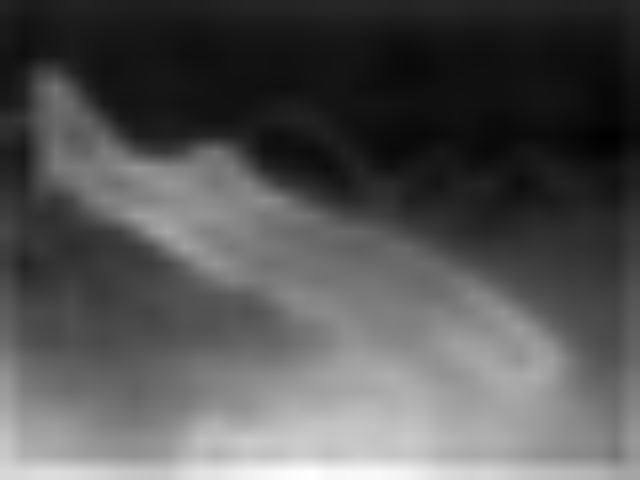} \\

            (a) $M_e^1$ & 
            (b) $M_c^1$ &
            (c) $M_r^1$ \\

            \includegraphics[height=3cm,width=4cm]{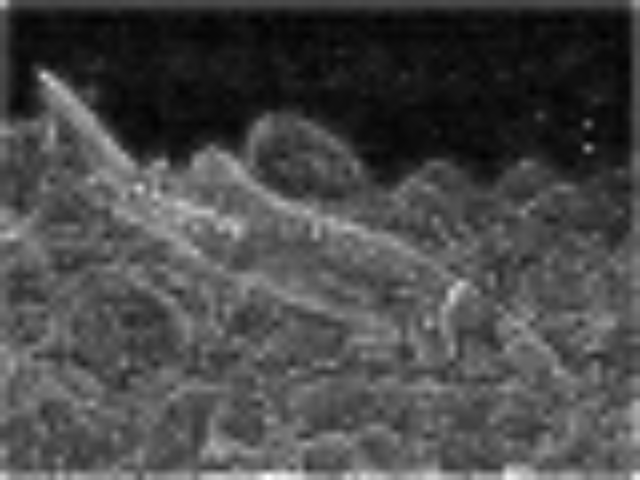} &
            \includegraphics[height=3cm,width=4cm]{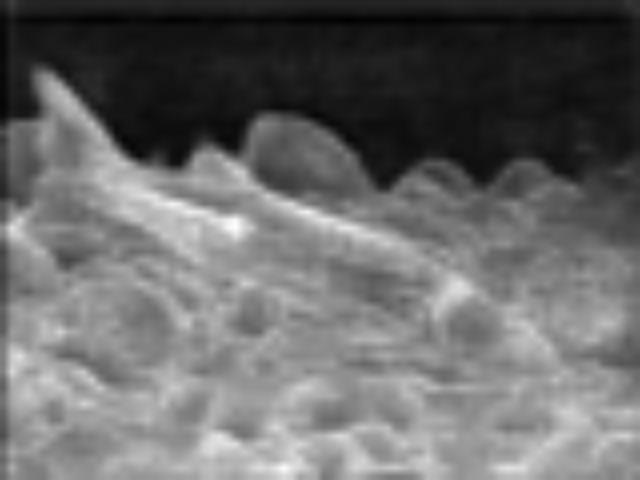} &
            \includegraphics[height=3cm,width=4cm]{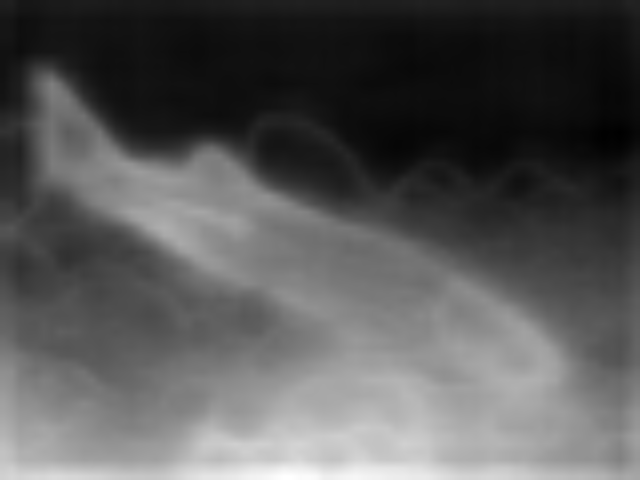} \\

            (d) $M_e^2$ & 
            (e) $M_c^2$ &
            (f) $M_r^2$ \\

            \includegraphics[height=3cm,width=4cm]{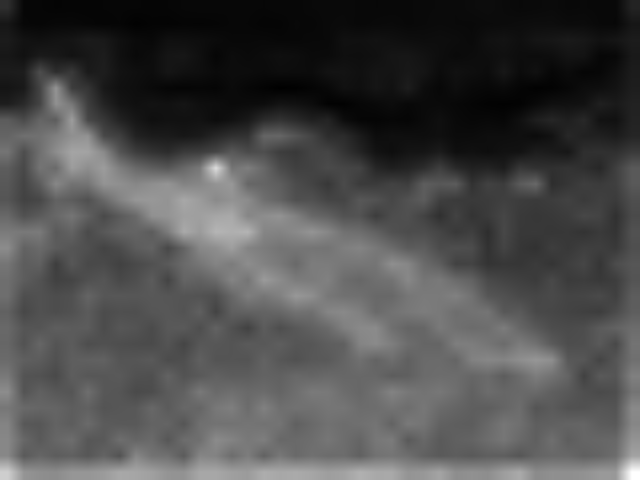} &
            \includegraphics[height=3cm,width=4cm]{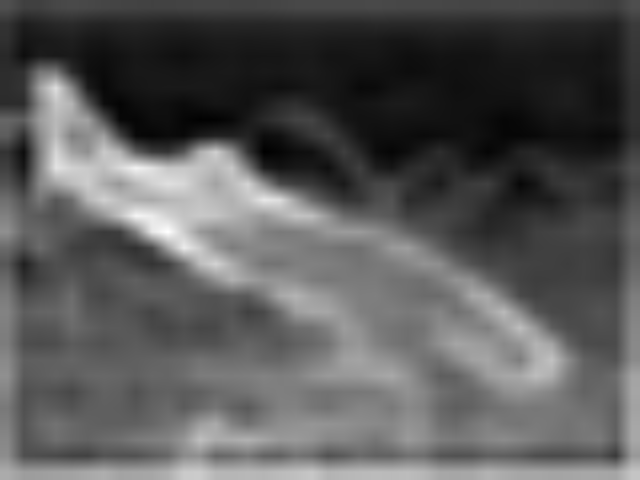} &
            \includegraphics[height=3cm,width=4cm]{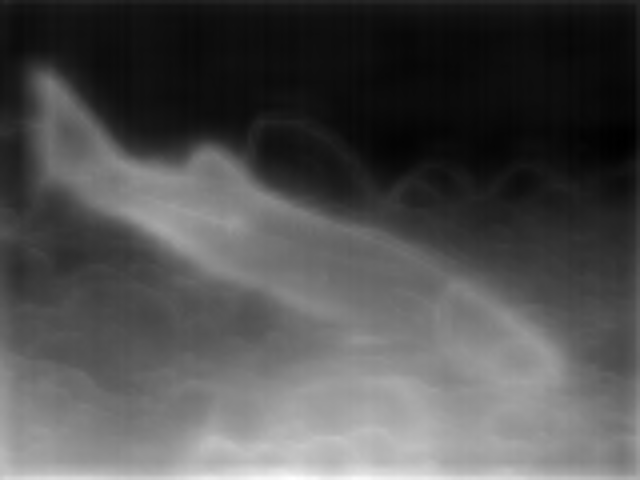} \\

            (g) $M_e^3$ & 
            (h) $M_c^3$ &
            (i) $M_r^3$ \\

            \includegraphics[height=3cm,width=4cm]{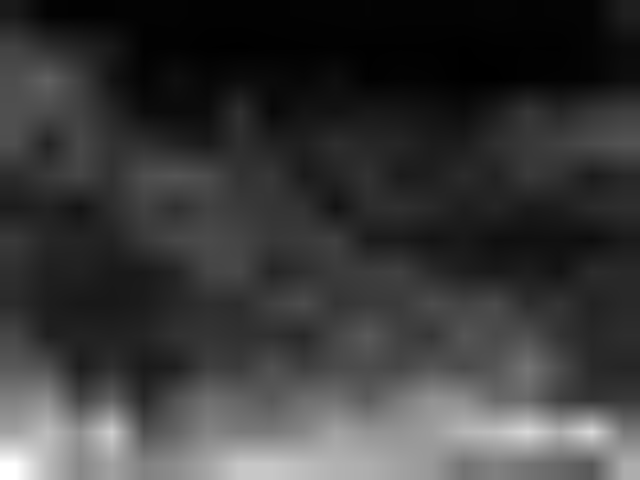} &
            \includegraphics[height=3cm,width=4cm]{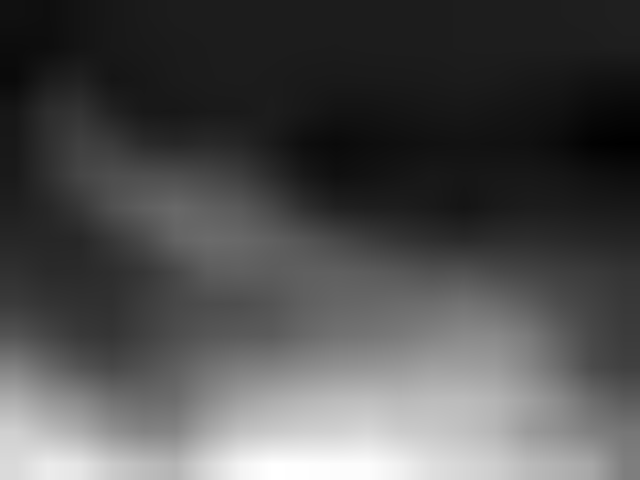} &
            \includegraphics[height=3cm,width=4cm]{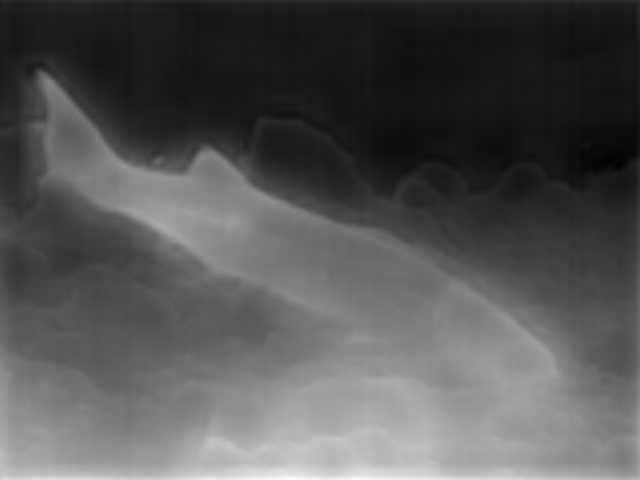} \\

            (j) $M_e^4$ & 
            (k) $M_c^4$ &
            (l) $M_r^4$ \\

            \multicolumn{3}{c}{
                \begin{tabular}{c@{\hspace{0.3em}}c}
                    \includegraphics[height=3cm,width=4cm]{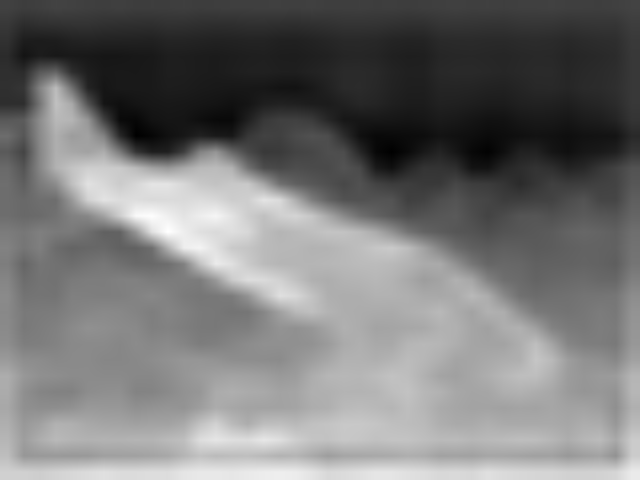} &
                    \includegraphics[height=3cm,width=4cm]{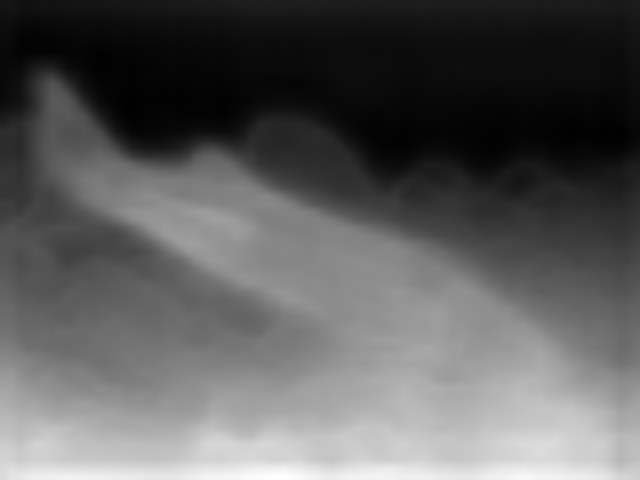} \\
                    (m) $M_t$ & 
                    (n) $M_f$
                \end{tabular}
            } \\

        \end{tabular}
    }
    \caption{Visualization results of feature maps from $M_e^{1}$ to $M_e^{4}$, from $M_c^{1}$ to $M_c^{4}$, from $M_r^{1}$ to $M_r^{4}$, $M_t$ and $M_f$ within the second Mamba fusion block of the second Mamba fusion layer. The higher brightness value denotes a greater weight. The brightness value is ranged in [0, 1].}
    \label{Mid_F}
\end{figure}

\vspace{-0.3cm}
\subsection{Overall Architecture}

The overall architecture of our proposed Tree-aware Mamba (Tree-Mamba) is shown in Fig. \ref{Framework}. 
Tree-Mamba is composed of a CNN-based encoder, four standard convolution layers with a 3$\times$3 kernel (Conv 3$\times$3), four Mamba fusion layers, and a depth head \cite{MiDas}. 
Given an underwater image, we first use a pre-trained encoder, which can be any CNNs, to extract multi-scale feature maps $\left \{M_e^{1},...,M_e^{4}\right \}$, since the feature extraction ability of a CNN can boost the performance of MDE \cite{MiDas, DA1, DA2}. 
As illustrated in Fig. \ref{Mid_F}, $M_e^1$ to $M_e^4$ contain various scales of structure and texture information.
Then, Conv 3$\times$3 layers are used to reduce the channel of these feature maps and obtain the initial refined feature maps $\left \{M_c^{1},...,M_c^{4}\right \}$.
As depicted in Fig. \ref{Mid_F}, $M_c^1$ to $M_c^4$ show richer scene structures than those of $M_e^1$ to $M_e^4$.
Subsequently, the Mamba fusion layer consisting of two Mamba fusion blocks and an up-sampling is used to further refine these feature maps $\left \{M_r^{1},...,M_r^{4}\right \}$.
In Fig. \ref{Mid_F}, $M_r^1$ to $M_r^4$ exhibit clearer scene structures and a rough depth trend.
Within a Mamba fusion block, a tree-aware Mamba block and FFN module are introduced to enhance global and non-linear modeling capability of our Tree-Mamba, respectively.
In Fig. \ref{Mid_F}, output feature maps of our tree-aware Mamba block $M_t$ contain clear global structures, while output feature maps of our FFN module $M_f$ show finer ones.
Besides, our tree-aware Mamba block is built by following the architecture of the Mamba block \cite{Mamba}, to replace the naive scanning strategy with our tree-aware scanning strategy.
The proposed tree-aware scanning strategy is used to capture long-range dependencies of feature maps while maintaining locality and continuity of features, which will be detailed in the next subsection.
At last, the depth head \cite{MiDas} is used to infer the depth map.

\vspace{-0.3cm}
\subsection{Tree-aware Scanning Strategy}

Mamba achieves remarkable performance in temporal sequence modeling; however, it faces significant challenges in visual tasks because it unfolds 2D feature maps into 1D sequences for causal modeling, which does not consider connections between semantically related but non-neighboring patches.
Various vision Mamba-based methods \cite{VMamba, Zigma, VmambaIR, MaIR} have been developed to deal with non-causal image sequences.
However, their scanning strategies fail to effectively preserve inherent spatial context within 2D feature maps, since they perform directional scanning along different orientations fixedly.
To overcome this limitation, we propose a novel tree-aware scanning strategy, as shown in Fig. \ref{Framework}. 
Specifically, given a 2D feature map $x$, an undirected fully-connected graph is first constructed, where the edge weight is the feature similarity between adjacent feature vertices ($f_1, f_2$) calculated by the cosine distance $d_{cos}(f_1, f_2)$:
\begin{equation}
     d_{cos}(f_1, f_2)=1-\frac{\sum_{i=1}^{ch} f_1^i f_2^i}{\sqrt{\sum_{i=1}^{ch} (f_1^i)^2} \cdot \sqrt{\sum_{i=1}^{ch} (f_2^i)^2}},\\
\end{equation}
where $ch$ denotes the number of feature channels.
Then, the Boruvka algorithm is used to prune the edges with significant similarity and generate a minimum spanning tree.
After that, the tree state propagation is conducted according to the propagation path provided by the breadth-first search:
\begin{equation}
     h^{lr}_i  = \sum_{j\in\left\{ch(i)\right\}}\bar{A}_j h^{lr}_j + \bar{B}_i x_i, i \in \Omega_{lr},\\
\label{method_eq4}
\end{equation}
where $\Omega_{lr}$ is the index set of all vertices along the leaf-to-root $lr$ propagation path.
$ch(i)$ denotes all child nodes of the $i$-th node.
$\{ h^{lr}_i \}^L_{i=1}$ represents the latent state of the $i$-th node, and their initial values are the input sequence $\{ x_i \}^L_{i=1}$ with length $L$.
$\{ \bar{A}_i \}^L_{i=1}$ and $\{ \bar{B}_i \}^L_{i=1}$ determine the influences of child nodes and current input node on $\{ h^{lr}_i \}^L_{i=1}$, respectively.
After the state propagation in Eq. (\ref{method_eq4}), all nodes contain the states of their child nodes, but lack the states of their sibling nodes.
Therefore, the propagation along the path from root to leaf is conducted as follows:
\begin{equation}
     h_i = \bar{A}_i (h^{lr}_{pr(i)}-h^{lr}_i) + h^{lr}_i, i \in \Omega_{rl},\\
\label{method_eq5}
\end{equation}
where $\Omega_{rl}$ is the index set of all vertices along the root-to-leaf $rl$ propagation path.
$h^{lr}_{pr(i)}$ is the latent state of the parent node of the $i$-th node.
$(h^{lr}_{pr(i)}-h^{lr}_i)$ denotes aggregated states for all sibling nodes of the \( i \)-th node.
The final latent state vector $h$ is obtained, and the output vector $y$ is defined by
\begin{equation}
     y = C \otimes h,\\
\label{method_eq6}
\end{equation}
where $\{ C_i\}^L_{i=1}$ determine how $h$ is transformed into $y$.
The definitions of $\{ \bar{A}_i \}^L_{i=1}$, $\{ \bar{B}_i \}^L_{i=1}$, and $\{ C_i\}^L_{i=1}$ are available at the supplementary material\footnote{https://wyjgr.github.io/Tree-Mamba.html\label{sm}}.
The main steps of our tree-aware scanning strategy are summarized in \textbf{Algorithm \ref{algorithm_tss}}.

\begin{algorithm}[!ht]
    \renewcommand{\algorithmicrequire}{\textbf{Input}:}
    \renewcommand{\algorithmicensure}{\textbf{Output}:}
    \caption{Outline of Our Tree-aware Scanning Strategy}
    \label{algorithm_tss}
    \small
    \begin{algorithmic}[1]
        \REQUIRE Feature maps $\{ x_i \}^L_{i=1}$.
        \STATE Initial latent state $\{ h^{lr}_i \}^L_{i=1}$ = $\{ x_i \}^L_{i=1}$.
        \STATE Update initial latent state $\{ h^{lr}_i \}^L_{i=1}$ by Eq. (\ref{method_eq4}).
        \STATE Compute final latent state $h$ by Eq. (\ref{method_eq5}).
        \STATE Obtain the output $y$ by Eq. (\ref{method_eq6}).
        \ENSURE $y$.
\end{algorithmic}
\end{algorithm}

\vspace{-0.5cm}
\subsection{Convergence Proof}

 The convergence of our proposed tree-aware scanning method is theoretically proven.
Specifically, we prove the convergence of Eqs. (\ref{method_eq4}) and (\ref{method_eq5}) by reformulating the original expressions in matrix form and solving for their unique solutions.
Eq. (\ref{method_eq4}) can be reformulated in matrix form as follows:
\begin{equation}
\begin{aligned}
H^{lr}&=\bar{A}CH^{lr}+\bar{B}X\\
&=(I-\bar{A}C)^{-1}\bar{B}X,\\
\end{aligned}
\label{method_eq7}
\end{equation}
where $X=[x_1,x_2,...,x_L]^T$, $\bar{B}=diag(\bar{B}_1,\bar{B}_2,...,\bar{B}_L)$, and $\bar{A}=diag(\bar{A}_1,\bar{A}_2,...,\bar{A}_L)$.
$C=[C^1_{ij},C^2_{ij},...,C^L_{ij}]$ is the child node matrix, where $C^l_{ij}=1$ denotes that the $i$ node is the child node of $j$-th node. 
After expanding the Eq. (\ref{method_eq5}), the following matrix expression can be obtained by 
\begin{equation}
\begin{split}
H&=\bar{A}SH+(I-\bar{A})H^{lr}\\
&=\bar{A}SH+(I-\bar{A})(I-\bar{A}C)^{-1}\bar{B}X\\
&=(I-\bar{A}S)^{-1}+(I-\bar{A})(I-\bar{A}C)^{-1}\bar{B}X,
\end{split}
\label{method_eq8}
\end{equation}
where $S=[S^1_{ij},S^2_{ij},...,S^L_{ij}]$ is the parent node matrix, and $S^l_{ij}=1$ denotes that the $i$ node is the parent node of $j$-th node. 
When  $(I-\bar{A}S)^{-1}$ and $(I-\bar{A}C)^{-1}$ exist, Eqs. (\ref{method_eq7}) and (\ref{method_eq8}) have unique solutions.

According to the matrix theory, for any matrix $M$, its $(I - M)^{-1}$ exists if the spectral radius $\rho(M) < 1$, and $\rho(M) \leq \Vert M \Vert$, where $ \Vert \cdot \Vert$ denotes any induced norm. 
Based on the infinity norm $\Vert M \Vert_\infty = \max\limits_{i} \sum_{j}\left|M_{i j}\right|$ (maximum row sums), $\Vert S \Vert_\infty$, $\Vert C \Vert_\infty$, and $\Vert \bar{A} \Vert_\infty$ are analyzed respectively.
Since each node in the constructed minimum spanning tree has at most four child nodes and four parent nodes, both $\Vert S \Vert_\infty$ and $\Vert C \Vert_\infty$ satisfies:
\begin{equation}
\Vert S \Vert_\infty \leq 4,\Vert C \Vert_\infty \leq 4.\\
\label{method_eq9}
\end{equation}
$\Vert \bar{A} \Vert_\infty$ is equal to the absolute value of its largest diagonal element:
\begin{equation}
\| \bar{A} \|_{\infty} =  \max_i \left|\bar{A}_{i}\right|.
\label{method_eq10}
\end{equation}
Using the norm inequality in matrix multiplication, we have:
\begin{equation}
\begin{split}
\rho(\bar{A}S) \leq \| \bar{A}S \|_{\infty} & \leq \| \bar{A} \|_{\infty} \cdot \| S \|_{\infty}  \leq \ 4\max_i \left|\bar{A}_{i}\right|,\\
\rho(\bar{A}C) \leq \| \bar{A}C \|_{\infty} & \leq \| \bar{A} \|_{\infty} \cdot \| C \|_{\infty}  \leq \ 4\max_i \left|\bar{A}_{i}\right|.\\
\end{split}
\label{method_eq11}
\end{equation}
To ensure $\rho(\bar{A}S) < 1$ and $\rho(\bar{A}C) < 1$, the following condition must be satisfied:
\begin{equation*}
4 \max_i \left| \bar{A}_i \right| < 1 
\quad \Rightarrow \quad 
\max_i \left| \bar{A}_i \right| < \frac{1}{4}.
\label{method_eq12}
\end{equation*}
Consequently, when $\bar{A} \in (-\tfrac{1}{4}, \tfrac{1}{4})$, both $(I - AS)^{-1}$ and $(I - AC)^{-1}$ exist, thus Eqs. (\ref{method_eq7}) and (\ref{method_eq8}) have unique solutions.
The proof of Eqs. (\ref{method_eq4}) and (\ref{method_eq5}) is completed.
\hfill $\square$

To sum up, it is necessary to constrain the value range of the initialized  $\bar{A}$ through an extremum-based linear transformation:
\begin{equation}
\bar{A} = \frac{\bar{A}}{4 \cdot \max(|\bar{A}|)},
\label{method_eq13}
\end{equation}
where $\max(|\bar{A}|)$ denotes the maximum absolute value among all elements in $\bar{A}$.
This transformation ensures that the entries of $\bar{A}$ lie within the interval $(-\tfrac{1}{4}, \tfrac{1}{4})$.

\vspace{-0.4cm}
\subsection{Loss Function}

Following the suggestion in \cite{LAPNet}, we use a linear combination of the mean absolute error $\mathcal{L}_{mae}$, the gradient loss $\mathcal{L}_{grad}$, and the structural similarity loss $\mathcal{L}_{ssim}$ to evaluate the predicted depth map and the reference depth map as follows:
\begin{equation}
\mathcal{L}_{final}= \mathcal{L}_{mae} + \mathcal{L}_{grad} + \mathcal{L}_{ssim}.
\end{equation}
Specifically, $\mathcal{L}_{mae}$ evaluates the pixel-wise difference between the predicted depth map $x$ and the reference depth map $y$ by
\begin{equation}
\mathcal{L}_{mae}=\frac{1}{HW}\sum_{i,j}^{H,W}|y_{i,j}-x_{i,j}|,
\end{equation}
where $H$ and $W$ represent the height and width of a depth map, respectively. 
$\mathcal{L}_{grad}$ penalizes the errors around edges below:
\begin{equation}
\begin{split}
    \mathcal{L}_{grad}&=\frac{1}{HW}\sum_{i,j}^{H,W}\Delta_g^h(i,j)+\Delta_g^v(i,j),\\
    \Delta_g^h(i,j) &= |y_{i,j+1}-y_{i,j-1}|-|x_{i,j+1}-x_{i,j-1}| ,\\
   \Delta_g^v(i,j) &= |y_{i+1,j}-y_{i-1,j}|-|x_{i+1,j}-x_{i-1,j}|,\\
\end{split}
\end{equation}
where $\Delta_g^h(i,j)$ and $\Delta_g^v(i,j)$ represent the errors on the horizontal and vertical gradients, respectively.
$\mathcal{L}_{ssim}$ measures tiny structural errors:
\begin{equation}
\mathcal{L}_{ssim}=1-SSIM(y,x),
\end{equation}
where $SSIM$ is the structural similarity index measure \cite{SSIM}.

\vspace{-0.3cm}
\section{Experimental Results and Analysis}

To demonstrate the superior performance of the proposed Tree-Mamba, qualitative evaluation, quantitative assessment, ablation study, and computational efficiency are conducted, respectively.
\emph{Due to the limited space, more experimental results and implementation details can be found in the supplementary material\textsuperscript{\ref {sm}}}.

\vspace{-0.4cm}
\subsection{Experiment Settings}

\noindent
\textbf{Compared Methods.} 
We compare our Tree-Mamba method against thirteen MDE approaches, including 
five conventional UMDE methods (IBLA \cite{IBLA}, GDCP \cite{GDCP}, NUDCP \cite{NUDCP}, HazeLine \cite{HazeLine}, and ADPCC \cite{ADPCC}), 
six deep learning-based UMDE methods (UW-Net \cite{UW-Net}, UW-GAN \cite{UW-GAN}, UDepth \cite{UDepth}, UW-Depth \cite{UW-Depth}, WsUID-Net \cite{SUIM-SDA}, and WaterMono \cite{WaterMono}), 
and two deep learning-based MDE methods (MiDas \cite{MiDas} and Lite-Mono \cite{Lite-Mono}).  
For a fair comparison, we use the source code provided by the authors, retrain each method on the training set, and produce the best prediction results.

\begin{figure*}[!ht]
    \Large
    \centering
    \resizebox{0.84\linewidth}{!}{
        \begin{tabular}{c@{ }c@{ }c@{ }c@{ }c@{ }c@{ }c@{ }c@{ }}
            \includegraphics[height=3cm,width=4cm]{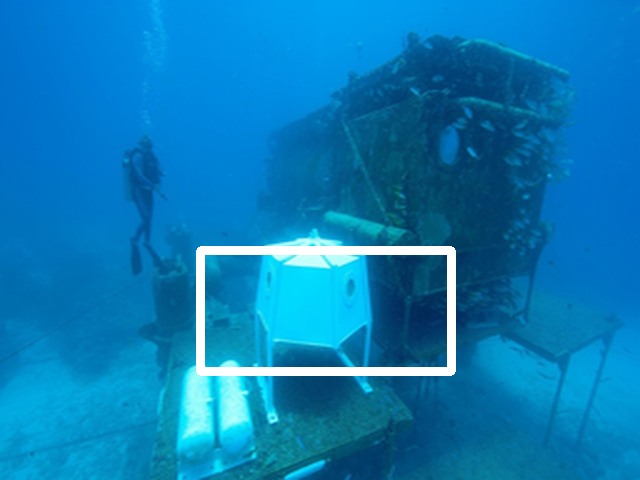} &
            \includegraphics[height=3cm,width=4cm]{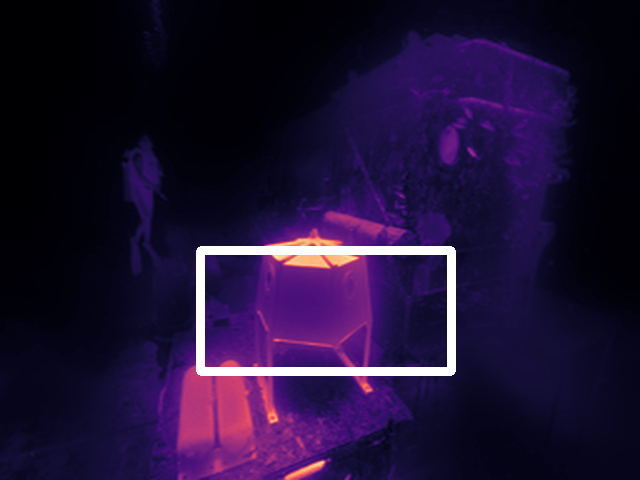} &
            \includegraphics[height=3cm,width=4cm]{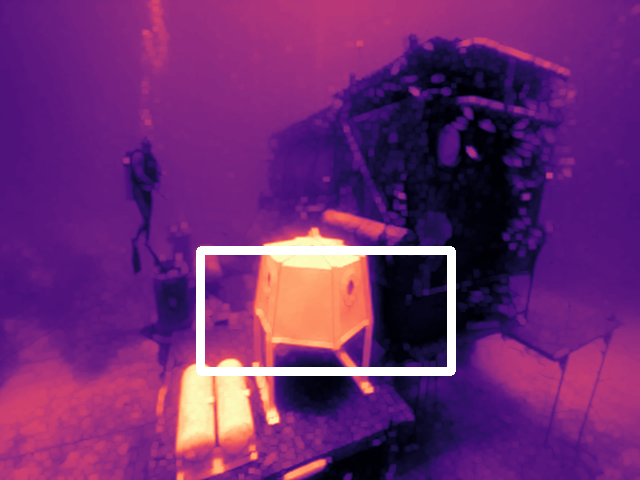} &
            \includegraphics[height=3cm,width=4cm]{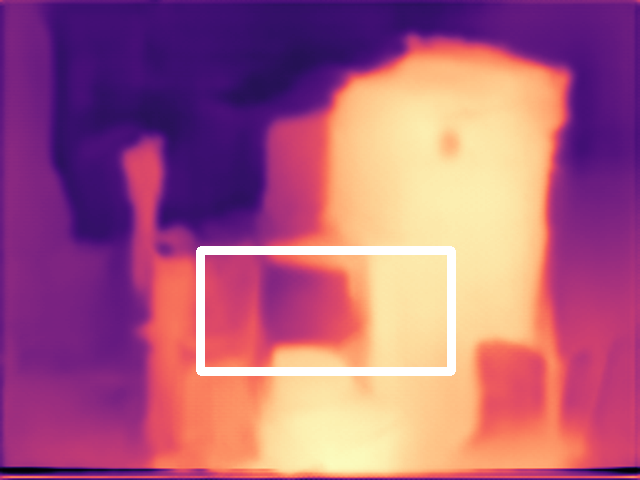} &
            \includegraphics[height=3cm,width=4cm]{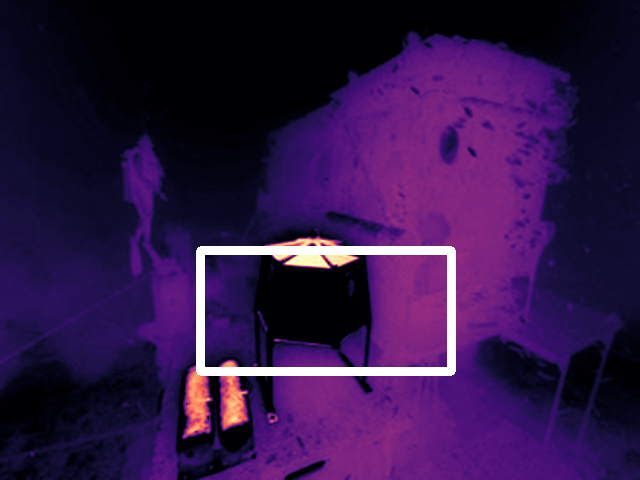} &
            \includegraphics[height=3cm,width=4cm]{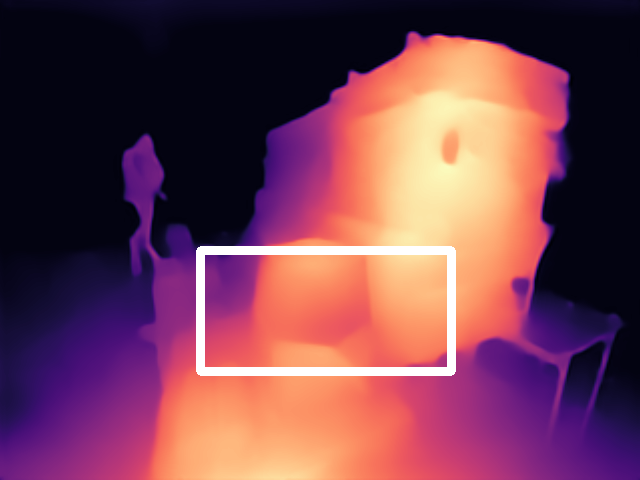} &
            \includegraphics[height=3cm,width=4cm]{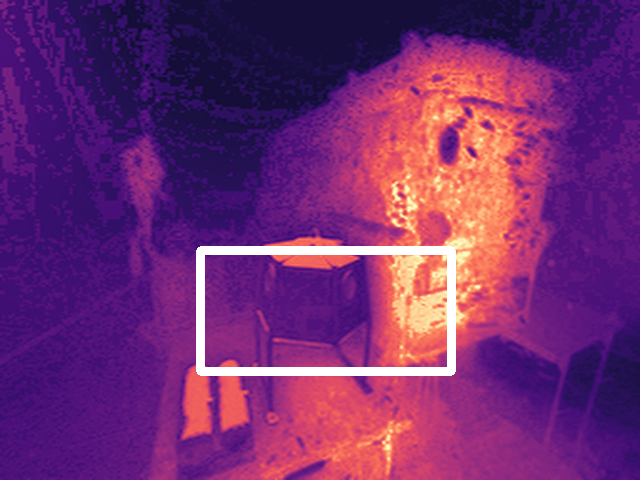} &
            \includegraphics[height=3cm,width=4cm]{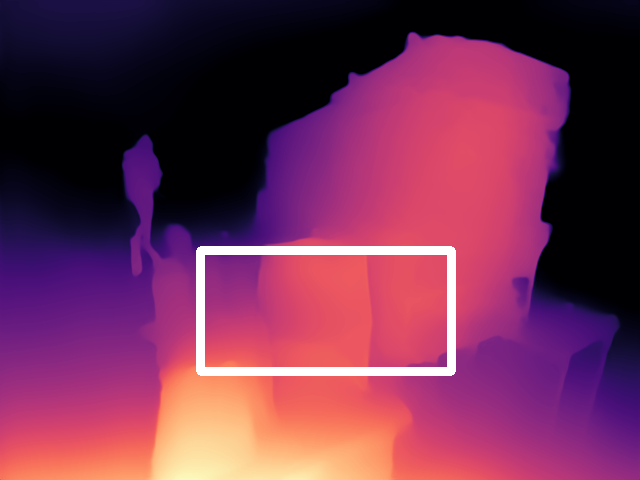} \\
            
            \includegraphics[height=2cm,width=4cm]{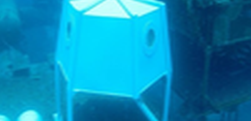} &
            \includegraphics[height=2cm,width=4cm]{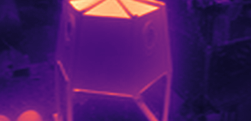} &
            \includegraphics[height=2cm,width=4cm]{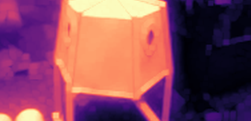} &
            \includegraphics[height=2cm,width=4cm]{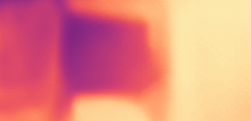} &
            \includegraphics[height=2cm,width=4cm]{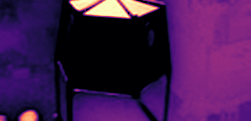} &
            \includegraphics[height=2cm,width=4cm]{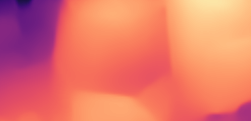} &
            \includegraphics[height=2cm,width=4cm]{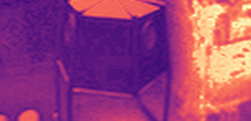} &
            \includegraphics[height=2cm,width=4cm]{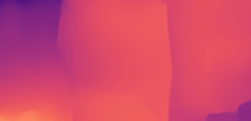} \\

            \includegraphics[height=3cm,width=4cm]{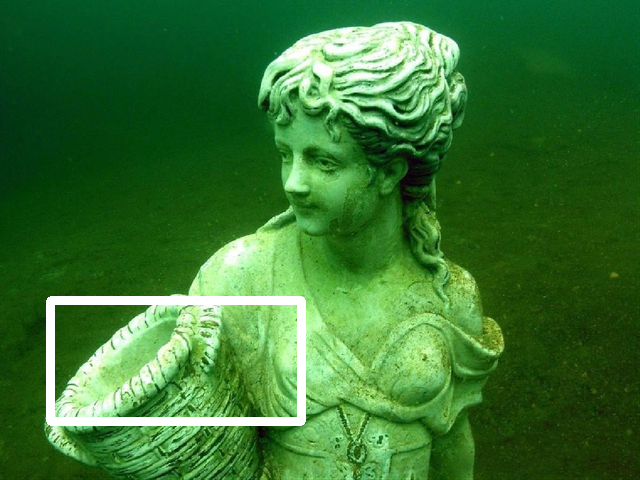} &
            \includegraphics[height=3cm,width=4cm]{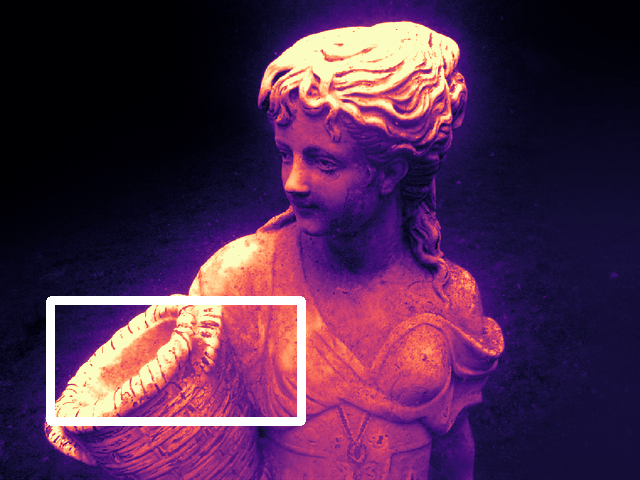} &
            \includegraphics[height=3cm,width=4cm]{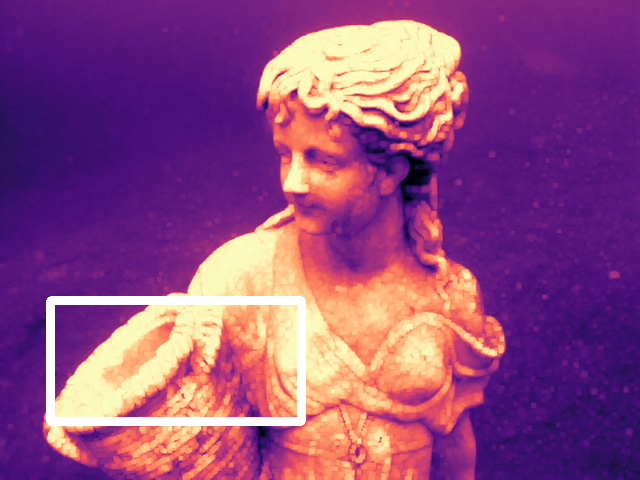} &
            \includegraphics[height=3cm,width=4cm]{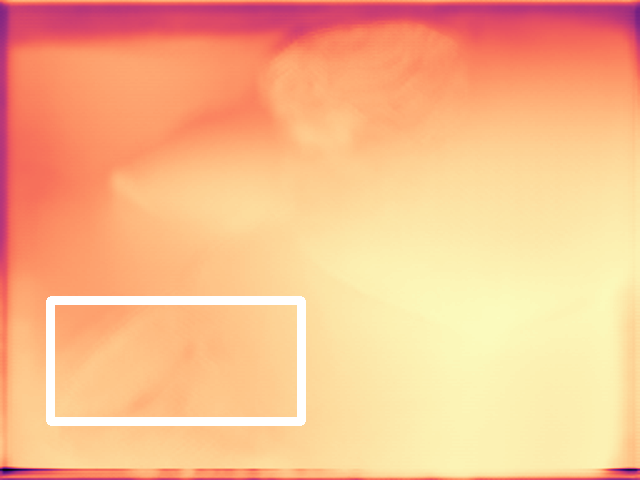} &
            \includegraphics[height=3cm,width=4cm]{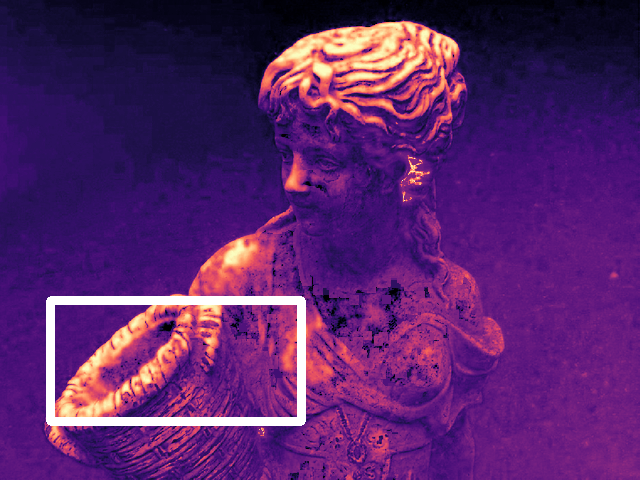} &
            \includegraphics[height=3cm,width=4cm]{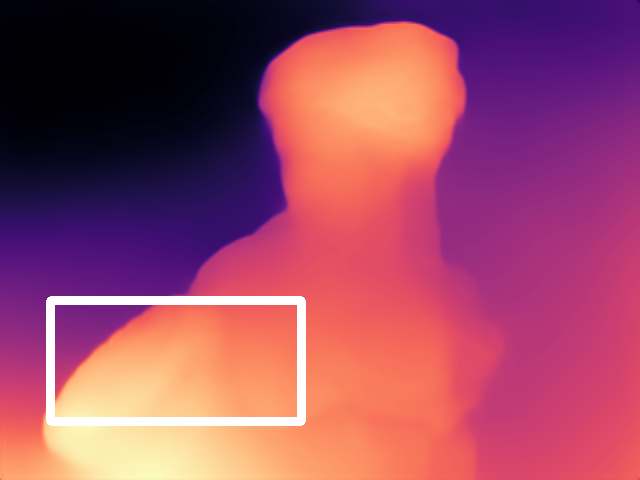} &
            \includegraphics[height=3cm,width=4cm]{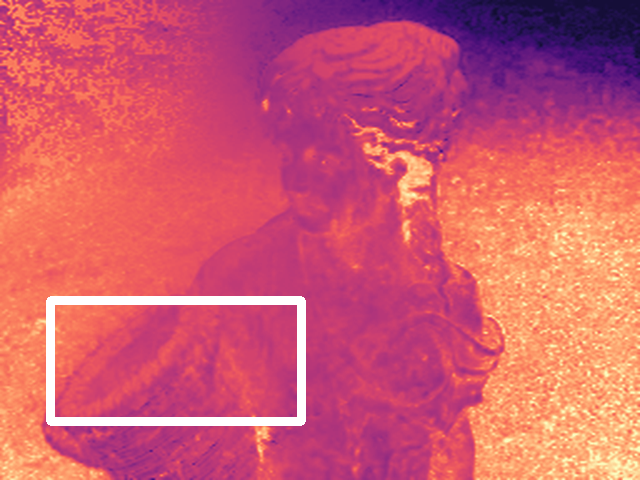} &
            \includegraphics[height=3cm,width=4cm]{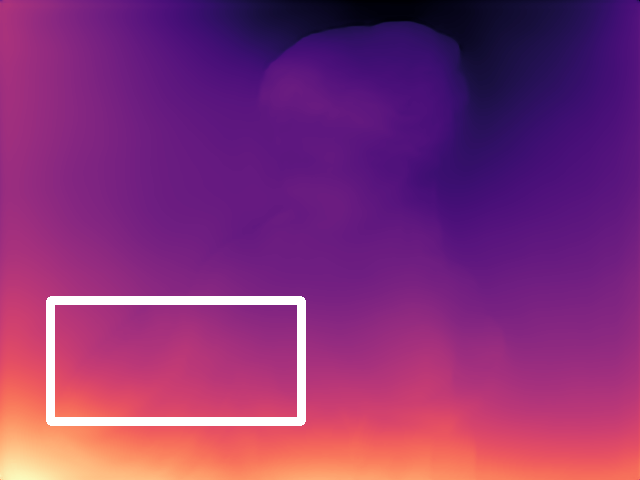} \\

            \includegraphics[height=2cm,width=4cm]{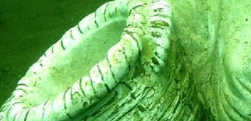} &
            \includegraphics[height=2cm,width=4cm]{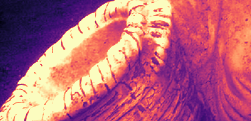} &
            \includegraphics[height=2cm,width=4cm]{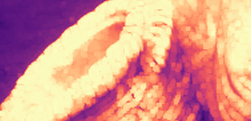} &
            \includegraphics[height=2cm,width=4cm]{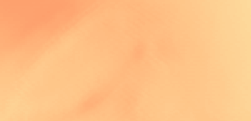} &
            \includegraphics[height=2cm,width=4cm]{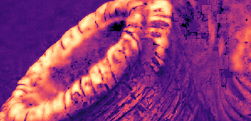} &
            \includegraphics[height=2cm,width=4cm]{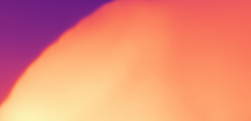} &
            \includegraphics[height=2cm,width=4cm]{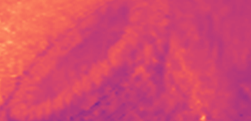} &
            \includegraphics[height=2cm,width=4cm]{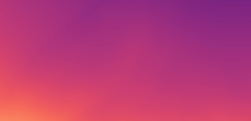} \\
            
            (a) Input & 
            (b) IBLA \cite{IBLA} &
            (c) GDCP \cite{GDCP} & 
            (d) UW-Net \cite{UW-Net} & 
            (e) NUDCP \cite{NUDCP} & 
            (f) UW-GAN \cite{UW-GAN} & 
            (g) HazeLine \cite{HazeLine} & 
            (h) MiDas \cite{MiDas} \\
                
            \includegraphics[height=3cm,width=4cm]{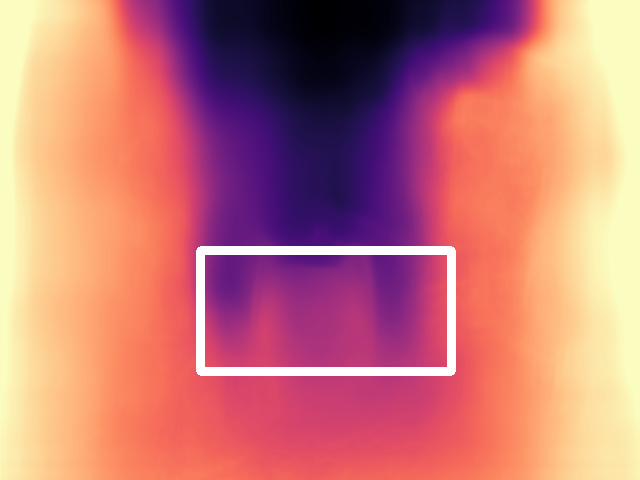} &
            \includegraphics[height=3cm,width=4cm]{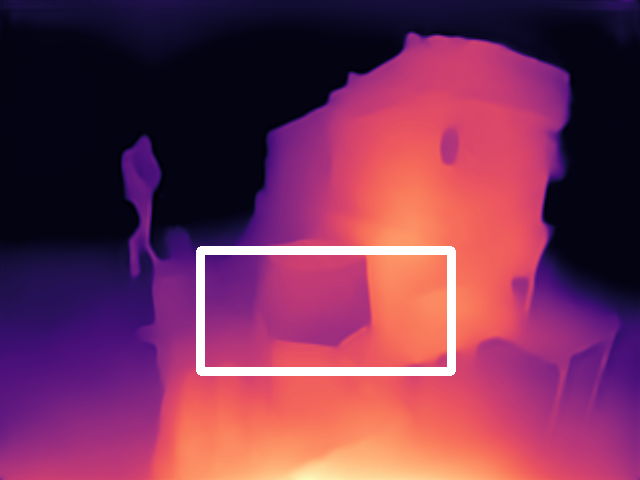} &
            \includegraphics[height=3cm,width=4cm]{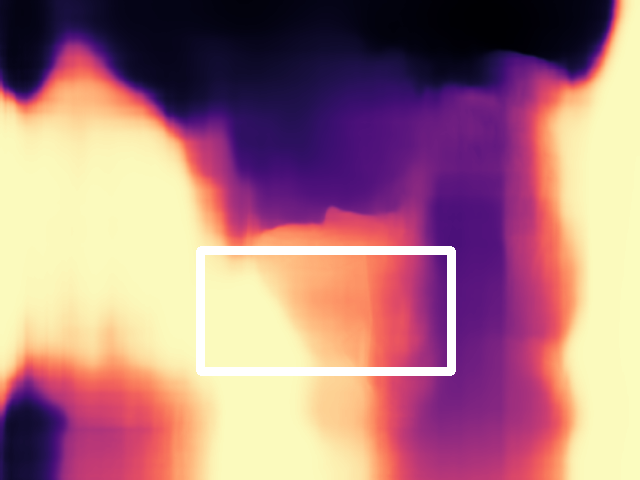} &
            \includegraphics[height=3cm,width=4cm]{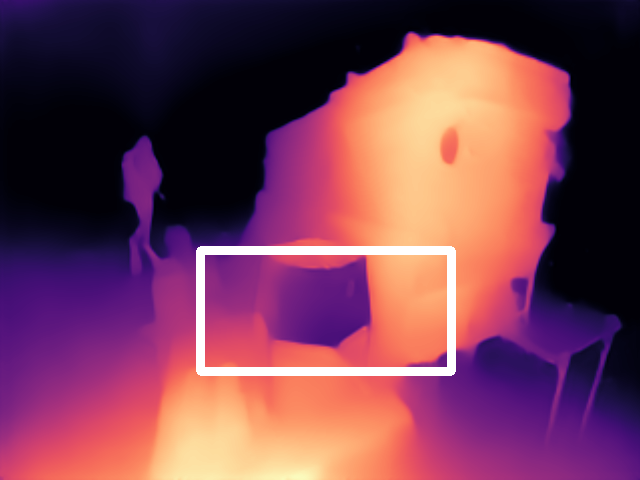} &
            \includegraphics[height=3cm,width=4cm]{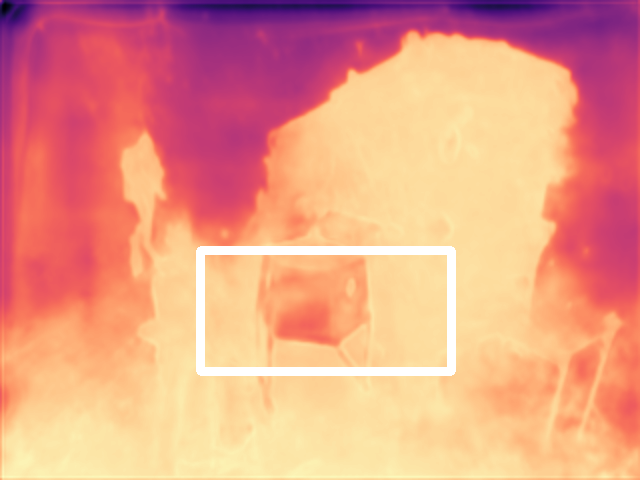} &
            \includegraphics[height=3cm,width=4cm]{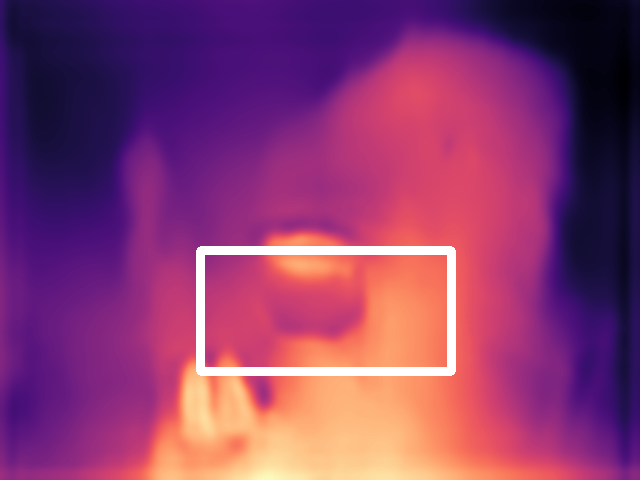} &
            \includegraphics[height=3cm,width=4cm]{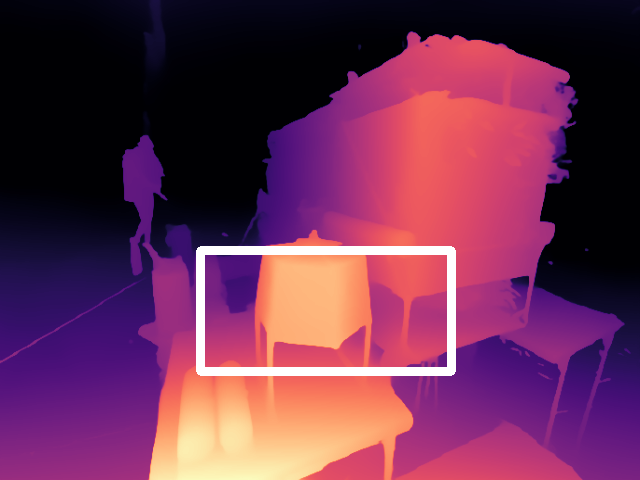} &
            \includegraphics[height=3cm,width=4cm]{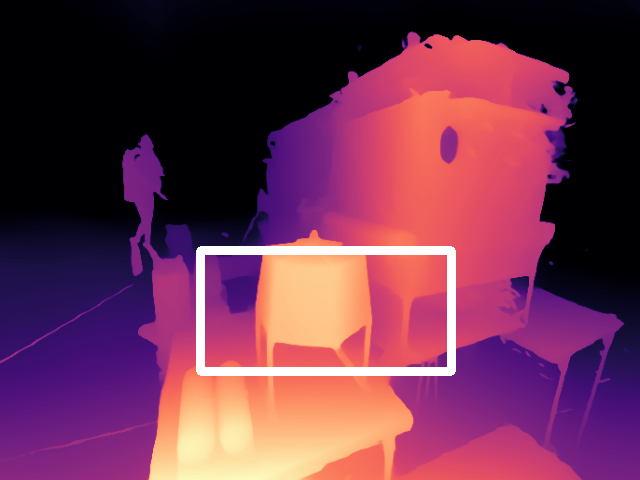} \\	

            \includegraphics[height=2cm,width=4cm]{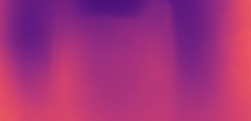} &
            \includegraphics[height=2cm,width=4cm]{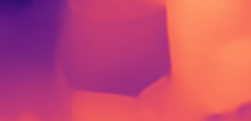} &
            \includegraphics[height=2cm,width=4cm]{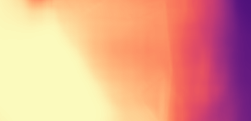} &
            \includegraphics[height=2cm,width=4cm]{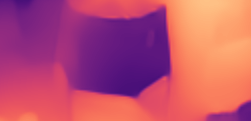} &
            \includegraphics[height=2cm,width=4cm]{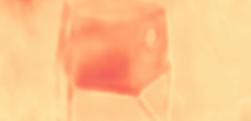} &
            \includegraphics[height=2cm,width=4cm]{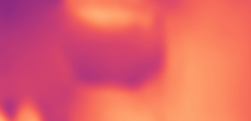} &
            \includegraphics[height=2cm,width=4cm]{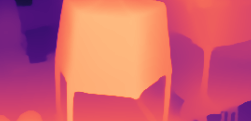} &
            \includegraphics[height=2cm,width=4cm]{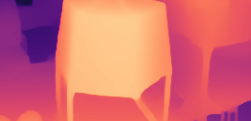} \\

            \includegraphics[height=3cm,width=4cm]{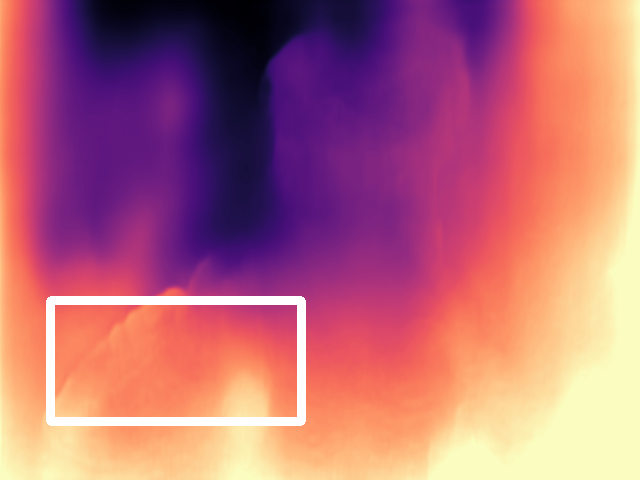} &
            \includegraphics[height=3cm,width=4cm]{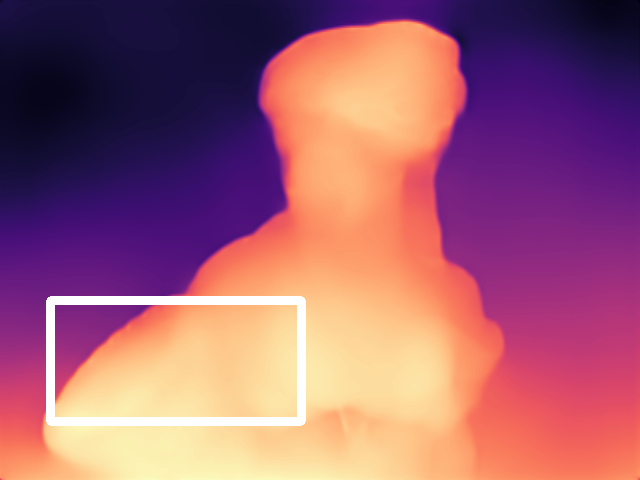} &
            \includegraphics[height=3cm,width=4cm]{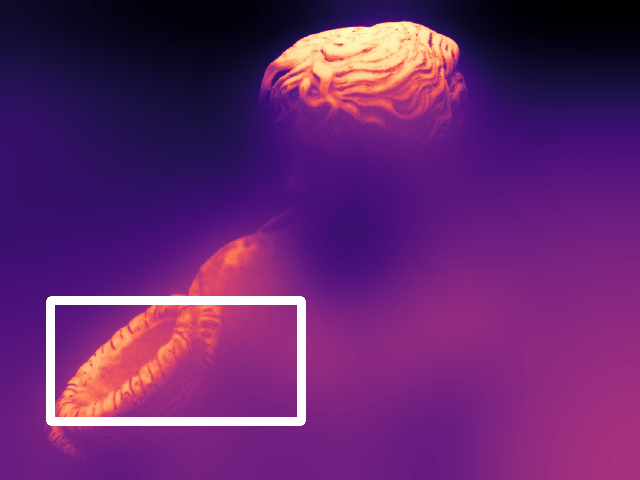} &
            \includegraphics[height=3cm,width=4cm]{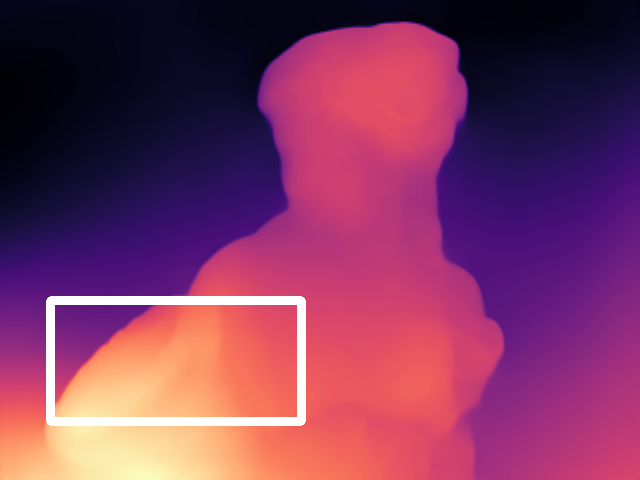} &
            \includegraphics[height=3cm,width=4cm]{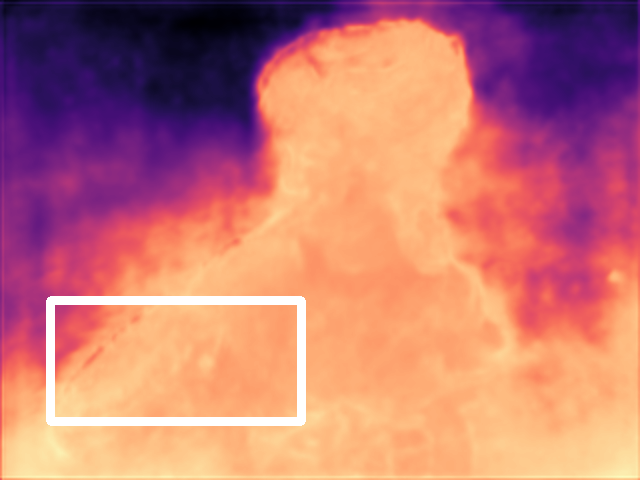} &
            \includegraphics[height=3cm,width=4cm]{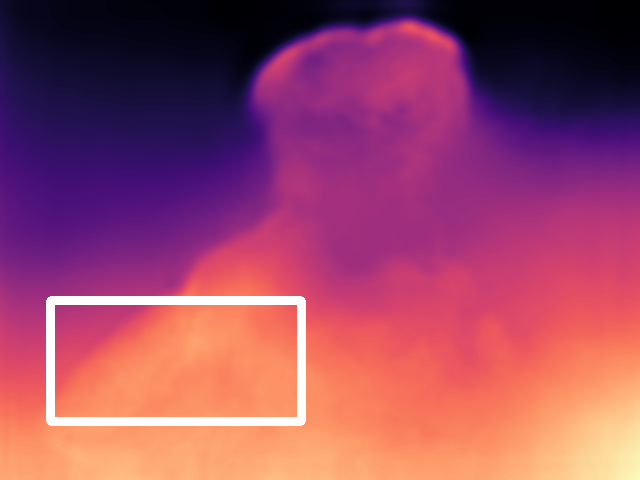} &
            \includegraphics[height=3cm,width=4cm]{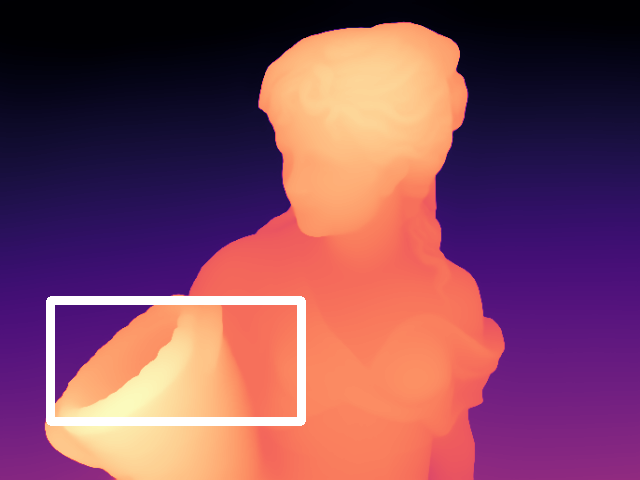} &
            \includegraphics[height=3cm,width=4cm]{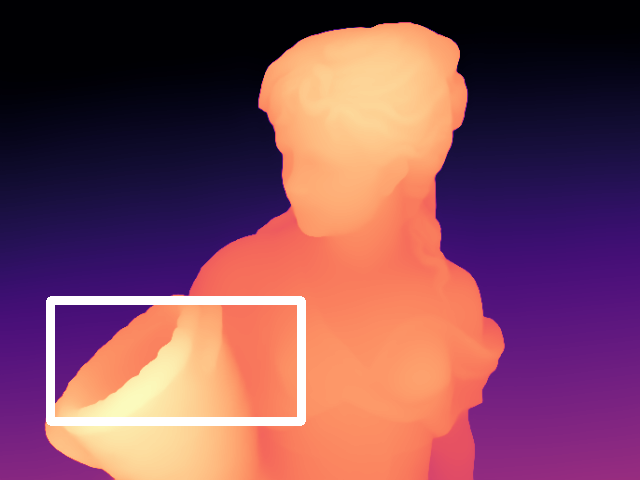} \\	

            \includegraphics[height=2cm,width=4cm]{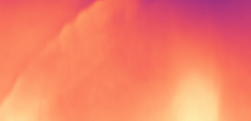} &
            \includegraphics[height=2cm,width=4cm]{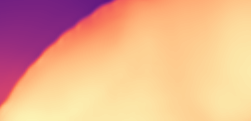} &
            \includegraphics[height=2cm,width=4cm]{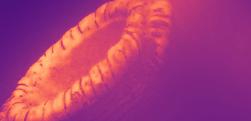} &
            \includegraphics[height=2cm,width=4cm]{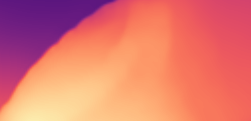} &
            \includegraphics[height=2cm,width=4cm]{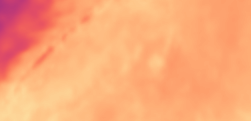} &
            \includegraphics[height=2cm,width=4cm]{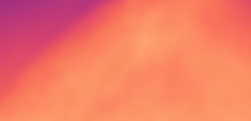} &
            \includegraphics[height=2cm,width=4cm]{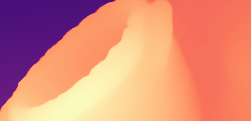} &
            \includegraphics[height=2cm,width=4cm]{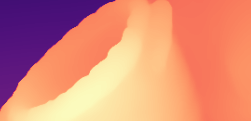} \\
            
            (i) Lite-Mono \cite{Lite-Mono} & 
            (j) UDepth \cite{UDepth} &
            (k) ADPCC \cite{ADPCC} & 
            (l) UW-Depth \cite{UW-Depth} & 
            (m) WsUID-Net \cite{SUIM-SDA} & 
            (n) WaterMono \cite{WaterMono} & 
            (o) Tree-Mamba & 
            (p) Reference \\
        \end{tabular}
    }
    \caption{Visual comparison of different methods on bluish and greenish underwater images from \textbf{Test-FR5691}. Compared with other competitors, our Tree-Mamba method yields better depth results on different degraded underwater images, and our depth maps are closer to the reference images.}
    \label{Qual_R}
\end{figure*}

\noindent
\textbf{Benchmark Datasets.} 
A ratio of 4:1 is used to randomly divide the proposed BlueDepth dataset into the training and test sets \cite{UW-GAN, UPGformer, CD-UDepth, SUIM-SDA}.
The training set contains 30,530 image pairs.
The test set has 7,632 image pairs, including 5,691 real underwater image pairs (\textbf{Test-FR5691}) and 1,941 synthetic underwater image pairs (\textbf{Test-FS1941}).
\textbf{Test-FR5691} and \textbf{Test-FS1941} are used on underwater general and turbid scenes, respectively.
Additionally, we collect a real underwater video set from the UVE38K dataset \cite{UVE38K}.
This video set contains 1,600 no-reference underwater image frames (\textbf{Video-NR1600}), involving 400 bluish, 400 greenish, 400 turbid, and 400 non-uniform light images.
Note that F and N represent full-reference and no-reference, while S and R denote synthetic and real.

\begin{figure*}[!ht]
    \Large
    \centering
    \resizebox{0.84\linewidth}{!}{
        \begin{tabular}{c@{ }c@{ }c@{ }c@{ }c@{ }c@{ }c@{ }c@{ }}
            \includegraphics[height=3cm,width=4cm]{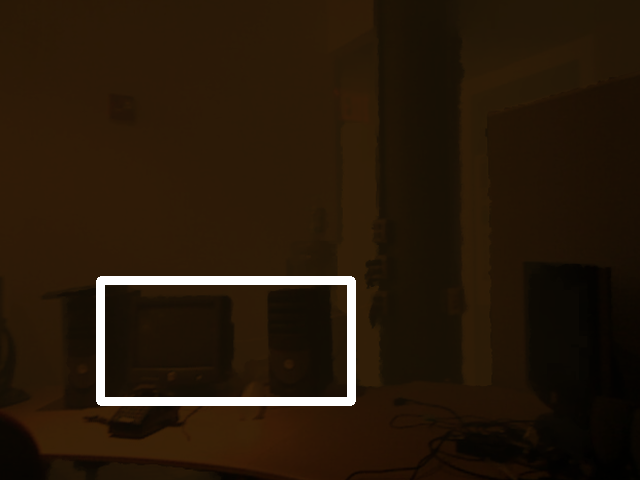} &
            \includegraphics[height=3cm,width=4cm]{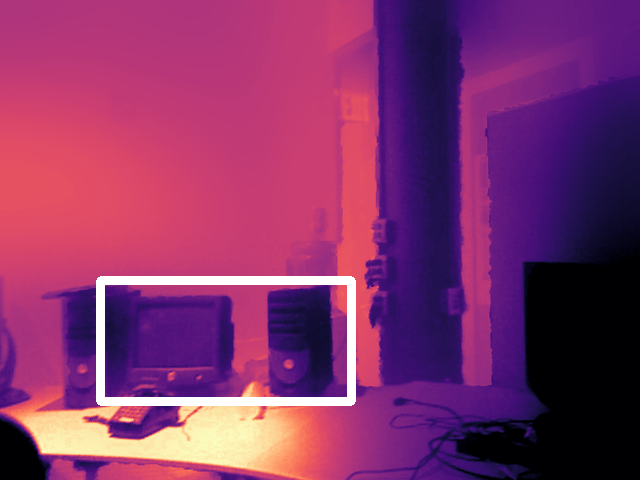} &
            \includegraphics[height=3cm,width=4cm]{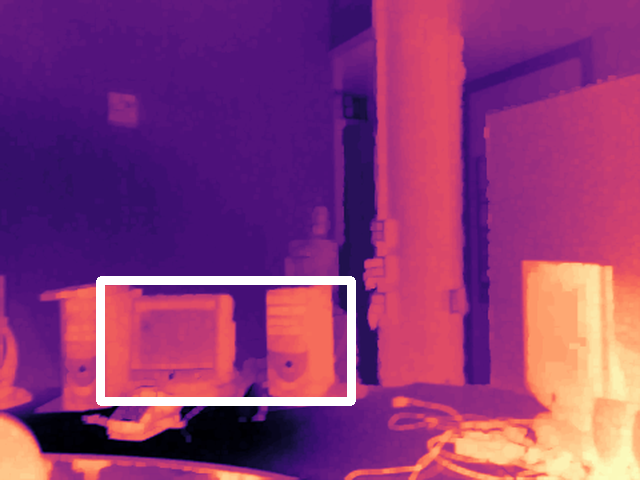} &
            \includegraphics[height=3cm,width=4cm]{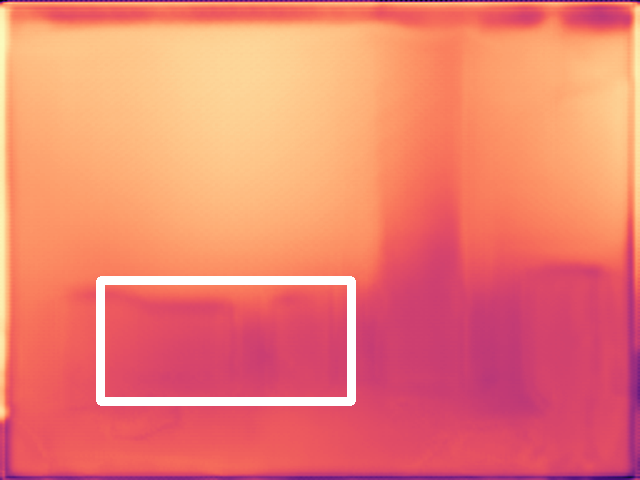} &
            \includegraphics[height=3cm,width=4cm]{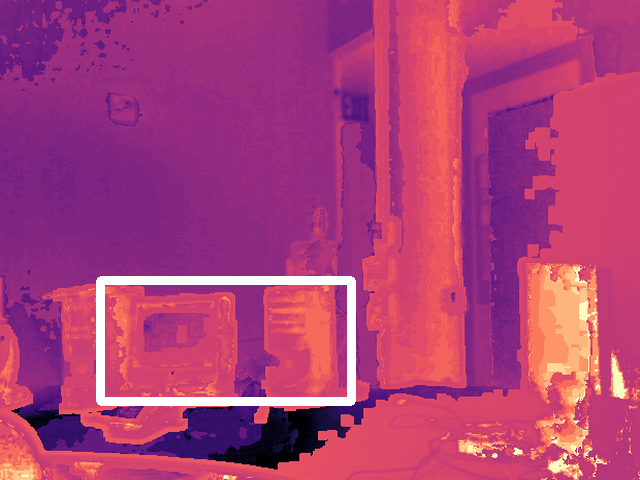} &
            \includegraphics[height=3cm,width=4cm]{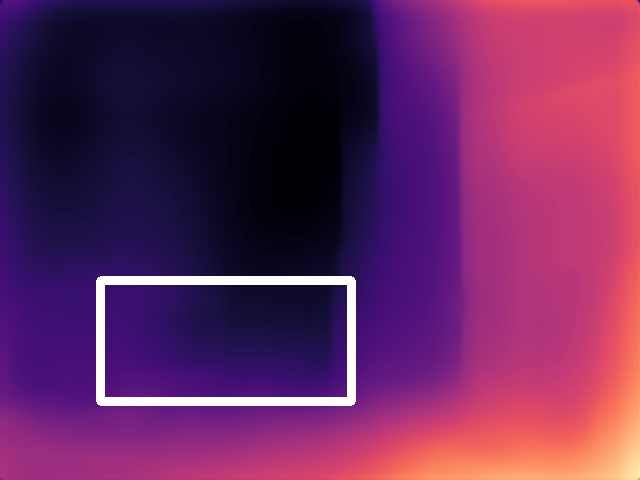} &
            \includegraphics[height=3cm,width=4cm]{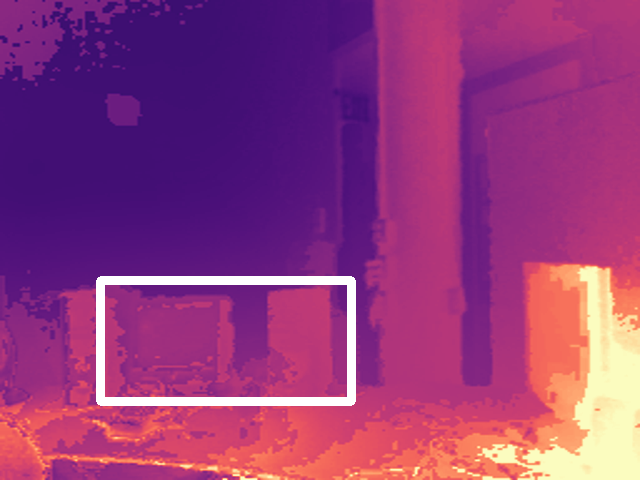} &
            \includegraphics[height=3cm,width=4cm]{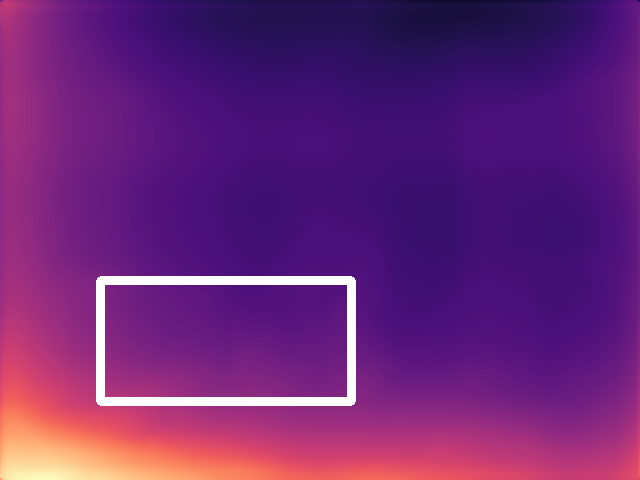} \\
            
            \includegraphics[height=2cm,width=4cm]{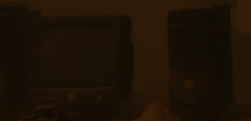} &
            \includegraphics[height=2cm,width=4cm]{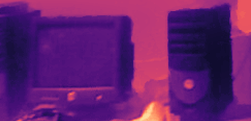} &
            \includegraphics[height=2cm,width=4cm]{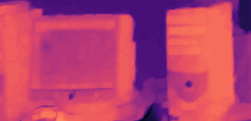} &
            \includegraphics[height=2cm,width=4cm]{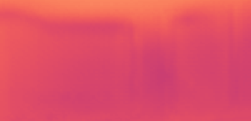} &
            \includegraphics[height=2cm,width=4cm]{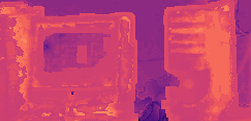} &
            \includegraphics[height=2cm,width=4cm]{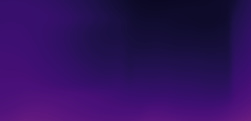} &
            \includegraphics[height=2cm,width=4cm]{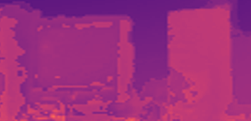} &
            \includegraphics[height=2cm,width=4cm]{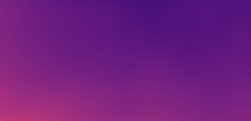} \\

            \includegraphics[height=3cm,width=4cm]{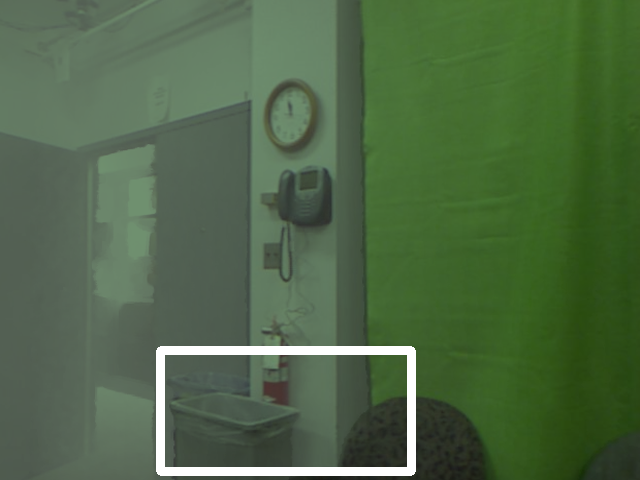} &
            \includegraphics[height=3cm,width=4cm]{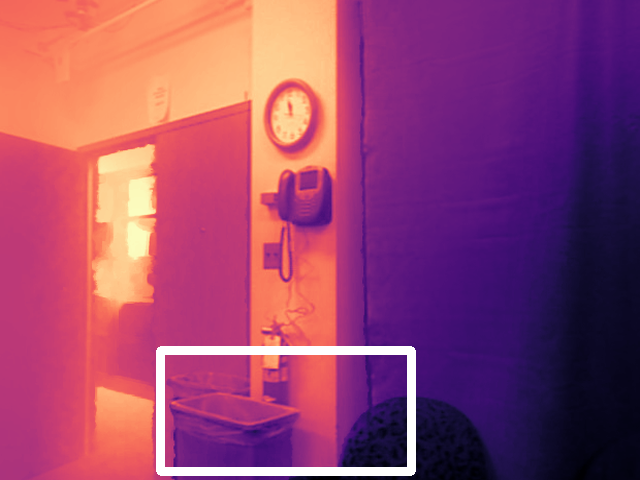} &
            \includegraphics[height=3cm,width=4cm]{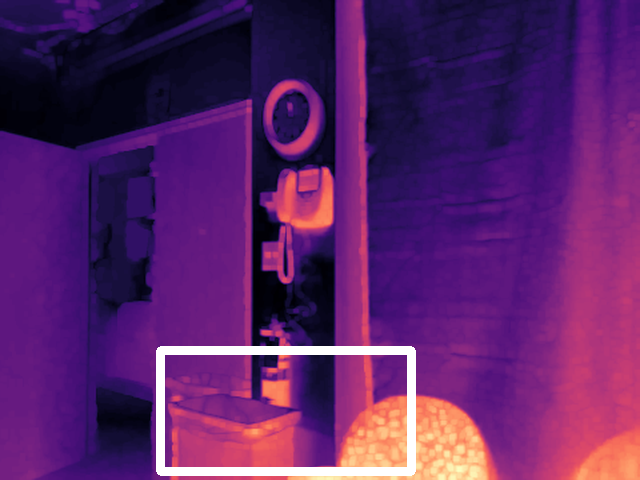} &
            \includegraphics[height=3cm,width=4cm]{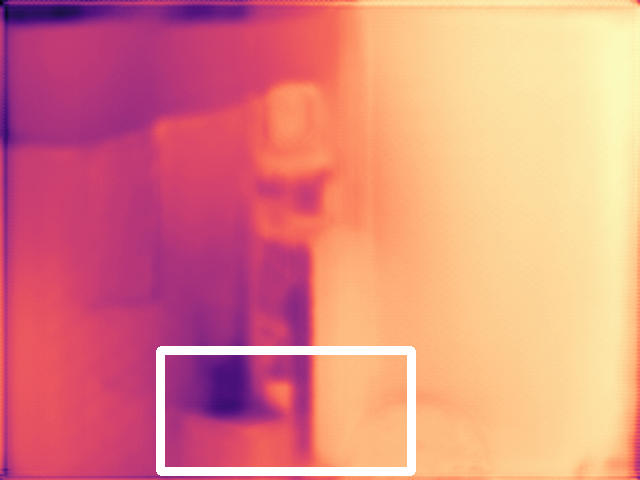} &
            \includegraphics[height=3cm,width=4cm]{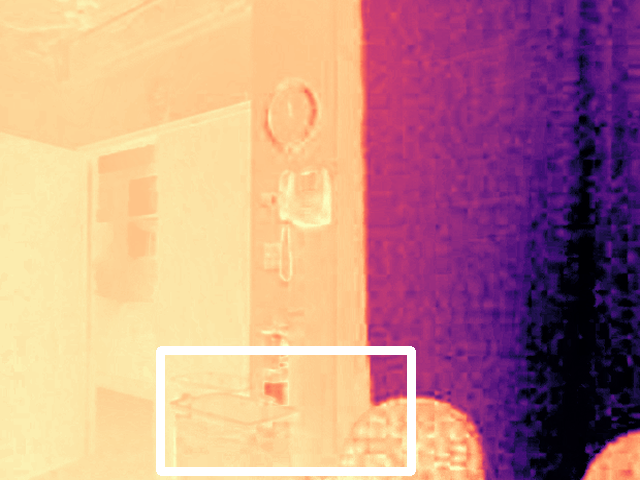} &
            \includegraphics[height=3cm,width=4cm]{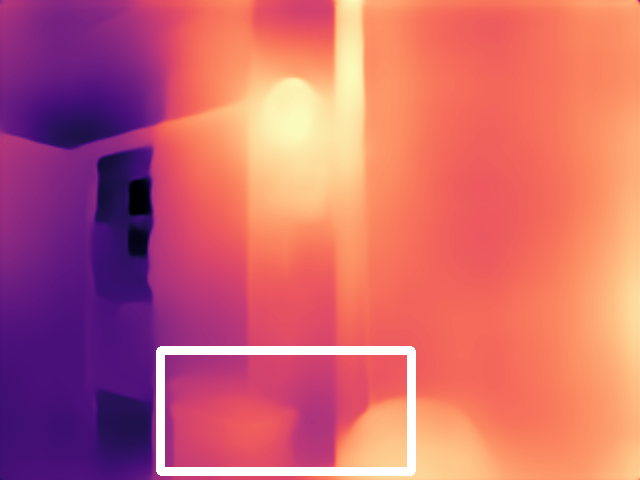} &
            \includegraphics[height=3cm,width=4cm]{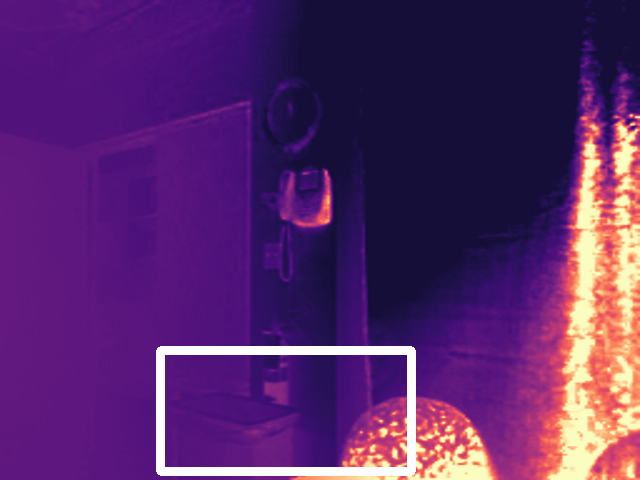} &
            \includegraphics[height=3cm,width=4cm]{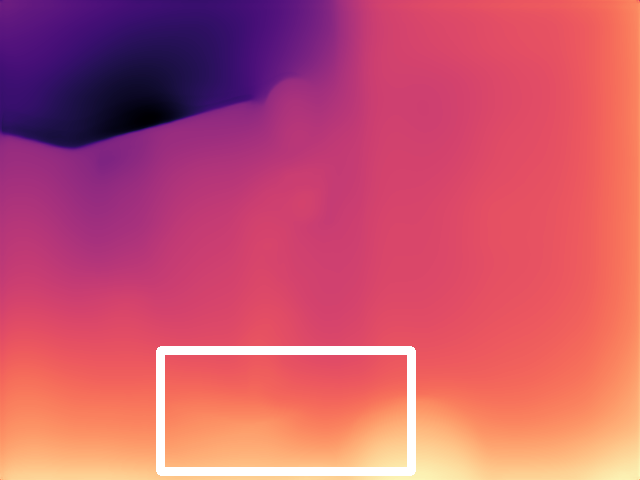} \\

            \includegraphics[height=2cm,width=4cm]{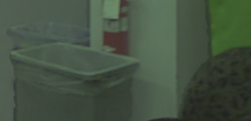} &
            \includegraphics[height=2cm,width=4cm]{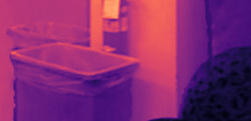} &
            \includegraphics[height=2cm,width=4cm]{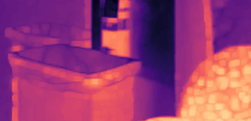} &
            \includegraphics[height=2cm,width=4cm]{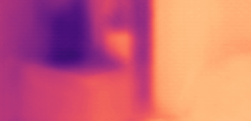} &
            \includegraphics[height=2cm,width=4cm]{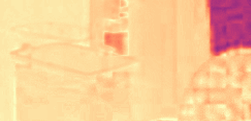} &
            \includegraphics[height=2cm,width=4cm]{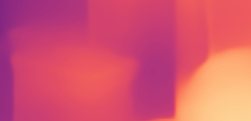} &
            \includegraphics[height=2cm,width=4cm]{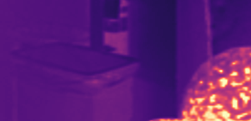} &
            \includegraphics[height=2cm,width=4cm]{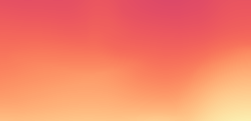} \\
            
            (a) Input & 
            (b) IBLA \cite{IBLA} &
            (c) GDCP \cite{GDCP} & 
            (d) UW-Net \cite{UW-Net} & 
            (e) NUDCP \cite{NUDCP} & 
            (f) UW-GAN \cite{UW-GAN} & 
            (g) HazeLine \cite{HazeLine} & 
            (h) MiDas \cite{MiDas} \\
                
            \includegraphics[height=3cm,width=4cm]{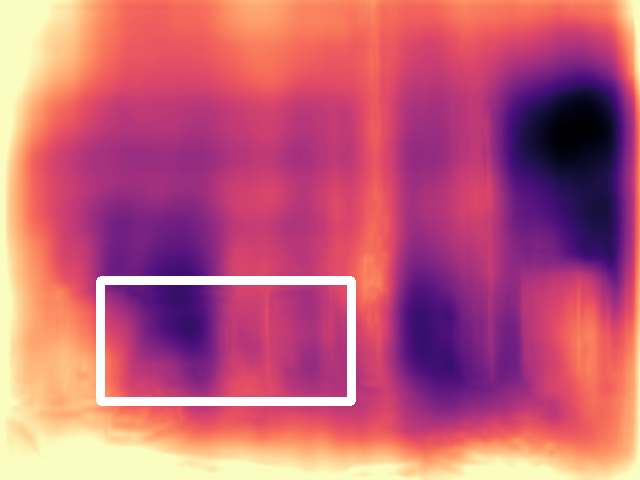} &
            \includegraphics[height=3cm,width=4cm]{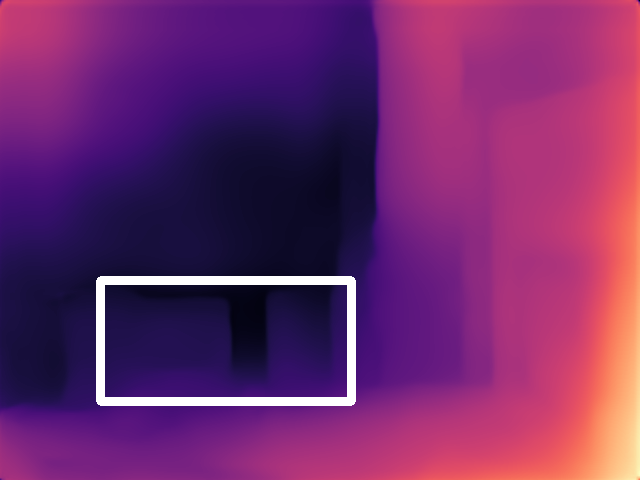} &
            \includegraphics[height=3cm,width=4cm]{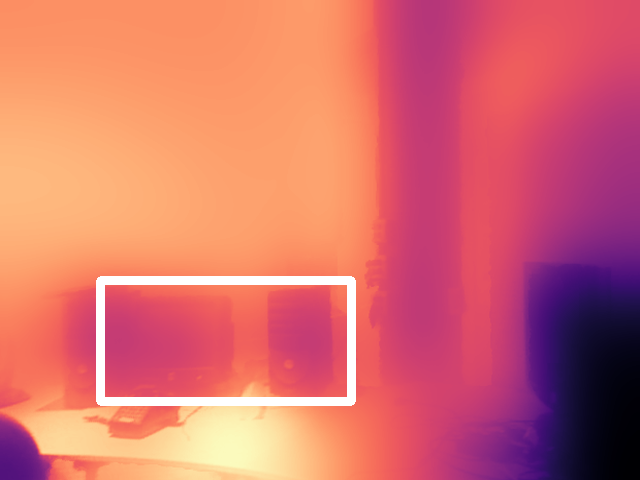} &
            \includegraphics[height=3cm,width=4cm]{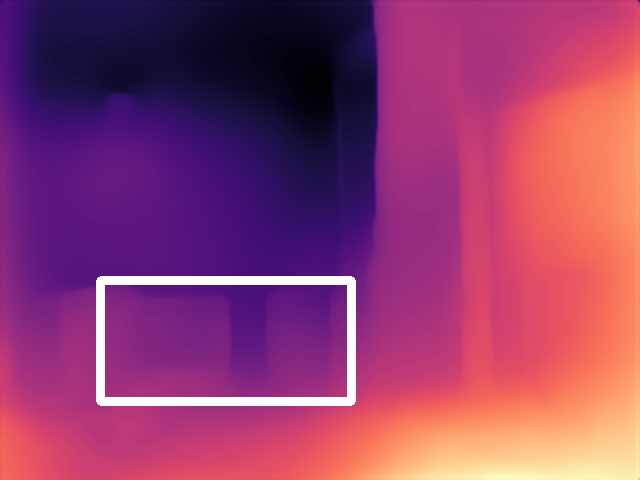} &
            \includegraphics[height=3cm,width=4cm]{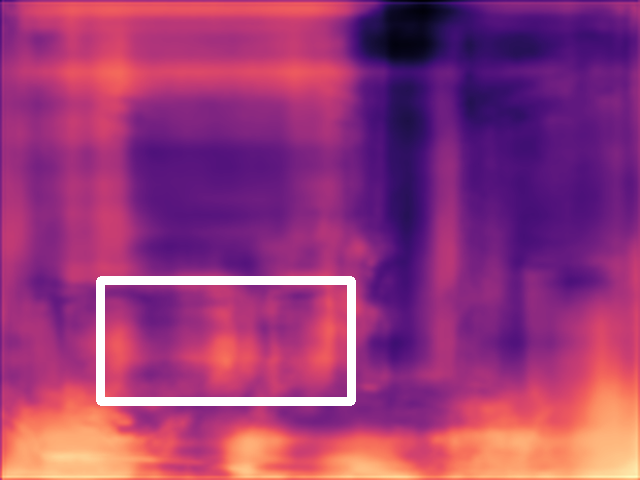} &
            \includegraphics[height=3cm,width=4cm]{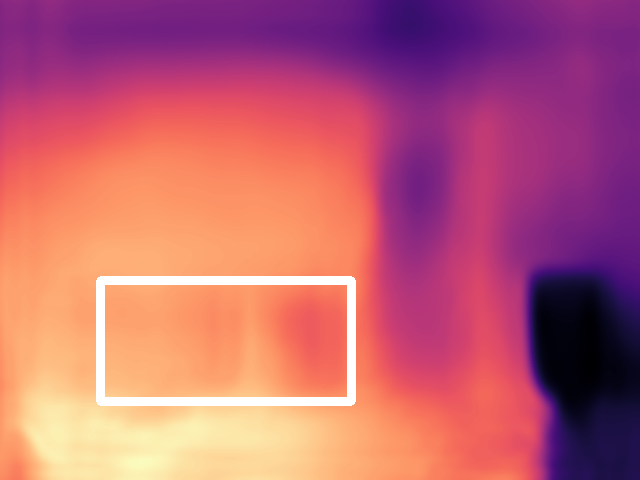} &
            \includegraphics[height=3cm,width=4cm]{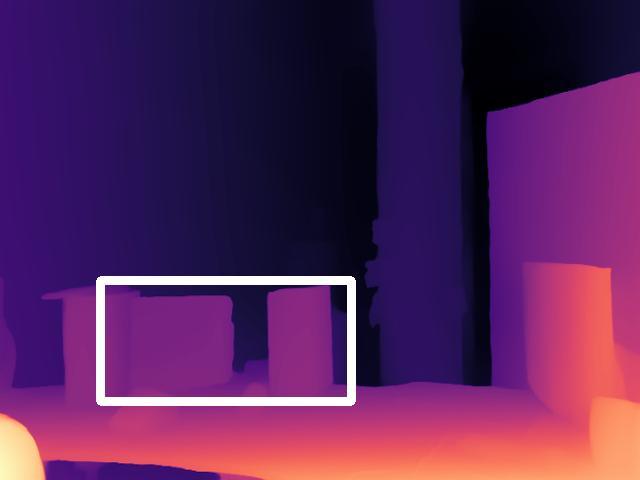} &
            \includegraphics[height=3cm,width=4cm]{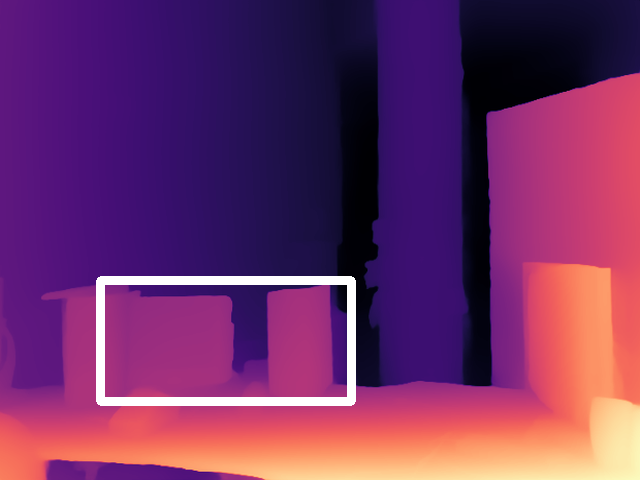} \\	

            \includegraphics[height=2cm,width=4cm]{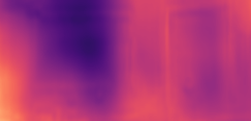} &
            \includegraphics[height=2cm,width=4cm]{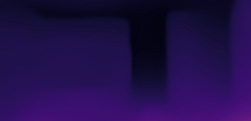} &
            \includegraphics[height=2cm,width=4cm]{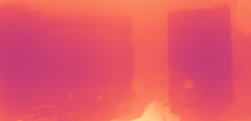} &
            \includegraphics[height=2cm,width=4cm]{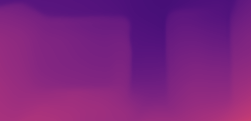} &
            \includegraphics[height=2cm,width=4cm]{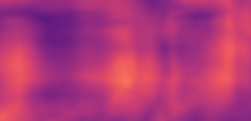} &
            \includegraphics[height=2cm,width=4cm]{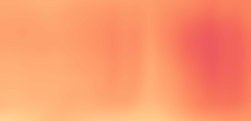} &
            \includegraphics[height=2cm,width=4cm]{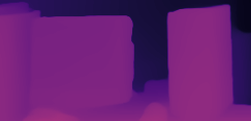} &
            \includegraphics[height=2cm,width=4cm]{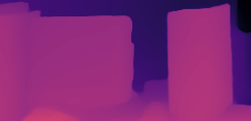} \\

            \includegraphics[height=3cm,width=4cm]{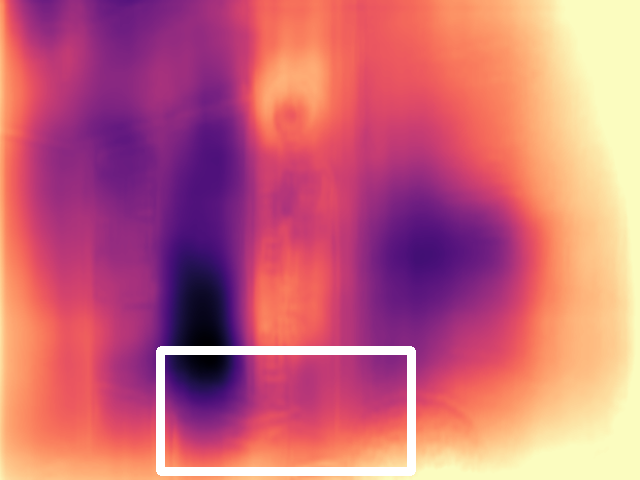} &
            \includegraphics[height=3cm,width=4cm]{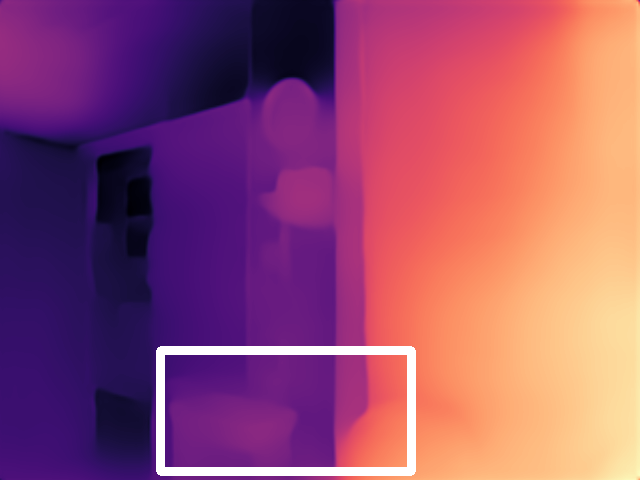} &
            \includegraphics[height=3cm,width=4cm]{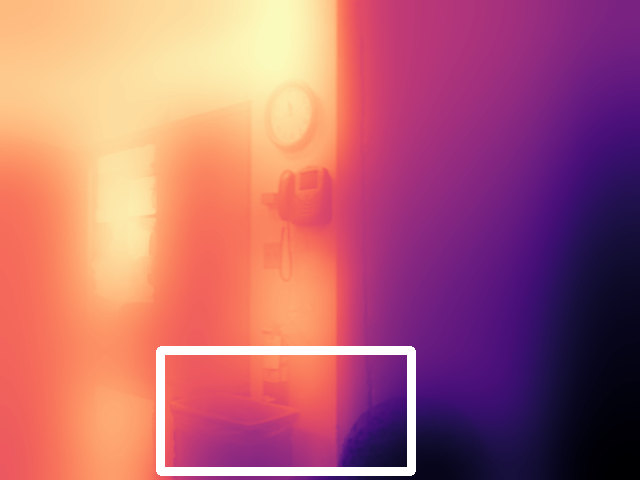} &
            \includegraphics[height=3cm,width=4cm]{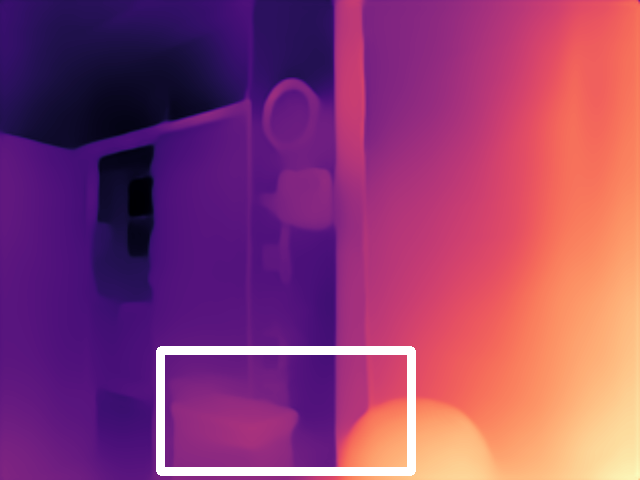} &
            \includegraphics[height=3cm,width=4cm]{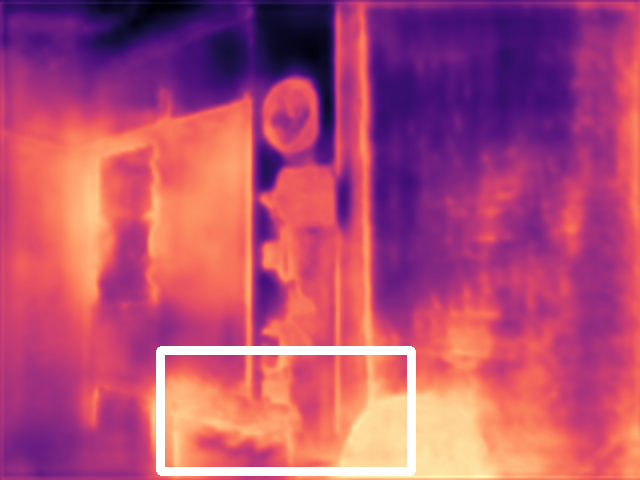} &
            \includegraphics[height=3cm,width=4cm]{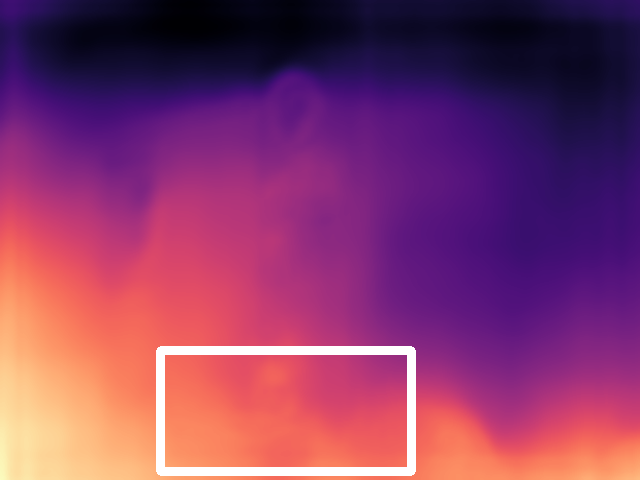} &
            \includegraphics[height=3cm,width=4cm]{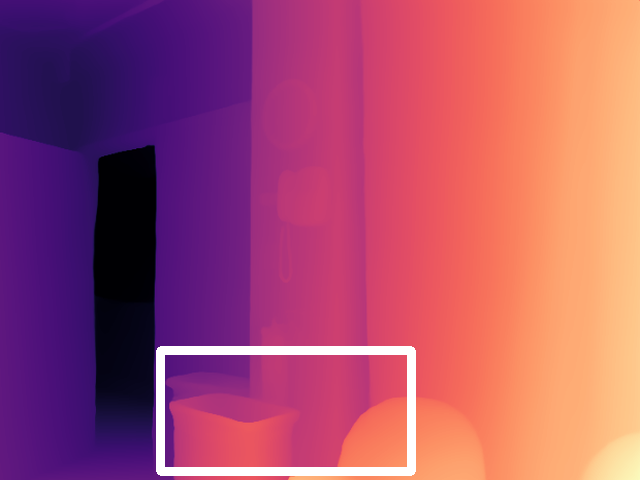} &
            \includegraphics[height=3cm,width=4cm]{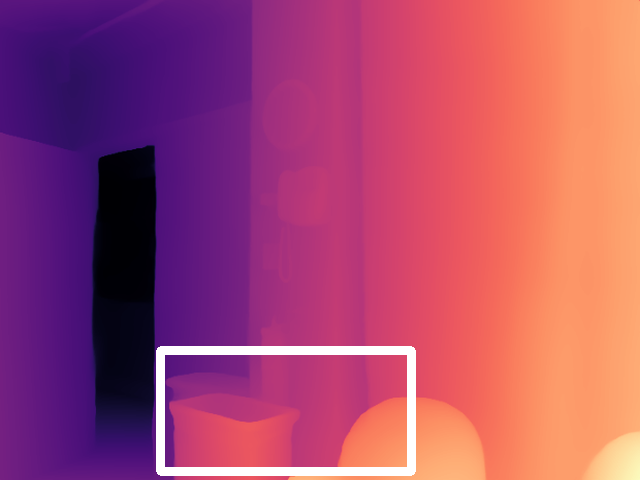} \\	

            \includegraphics[height=2cm,width=4cm]{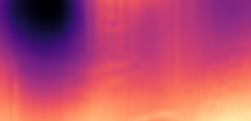} &
            \includegraphics[height=2cm,width=4cm]{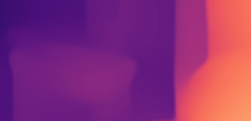} &
            \includegraphics[height=2cm,width=4cm]{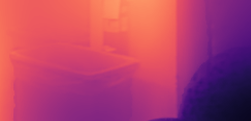} &
            \includegraphics[height=2cm,width=4cm]{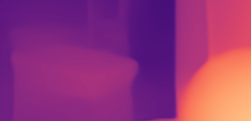} &
            \includegraphics[height=2cm,width=4cm]{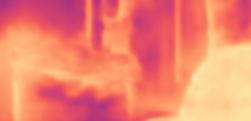} &
            \includegraphics[height=2cm,width=4cm]{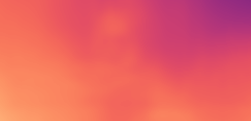} &
            \includegraphics[height=2cm,width=4cm]{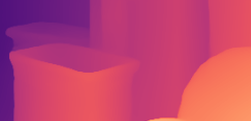} &
            \includegraphics[height=2cm,width=4cm]{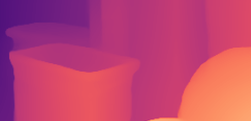} \\
            
            (i) Lite-Mono \cite{Lite-Mono} & 
            (j) UDepth \cite{UDepth} &
            (k) ADPCC \cite{ADPCC} & 
            (l) UW-Depth \cite{UW-Depth} & 
            (m) WsUID-Net \cite{SUIM-SDA} & 
            (n) WaterMono \cite{WaterMono} & 
            (o) Tree-Mamba & 
            (p) Reference \\
        \end{tabular}
    }
    \caption{Visual comparison of different methods on low-light and turbid underwater images from \textbf{Test-FS1941}.  Compared with other competitors, our Tree-Mamba method yields better depth results on different degraded underwater images, and our depth maps are closer to the reference images.}
    \label{Qual_S}
\end{figure*}

\begin{figure*}[!ht]
    \Large
    \centering
    \resizebox{0.84\linewidth}{!}{
        \begin{tabular}{c@{ }c@{ }c@{ }c@{ }c@{ }c@{ }c@{ }c@{ }}
            \includegraphics[height=3cm,width=4cm]{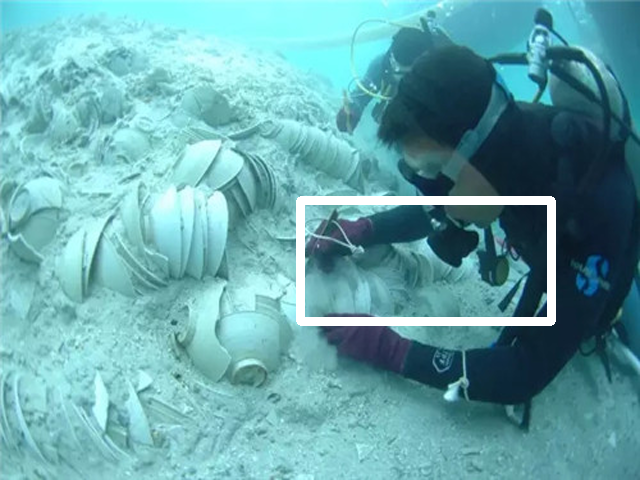} &
            \includegraphics[height=3cm,width=4cm]{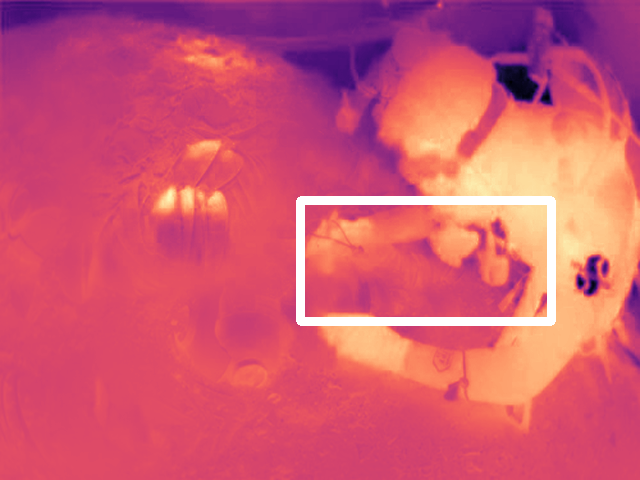} &
            \includegraphics[height=3cm,width=4cm]{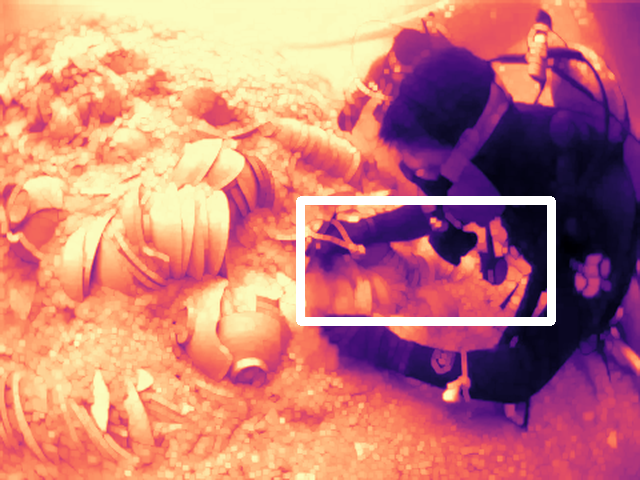} &
            \includegraphics[height=3cm,width=4cm]{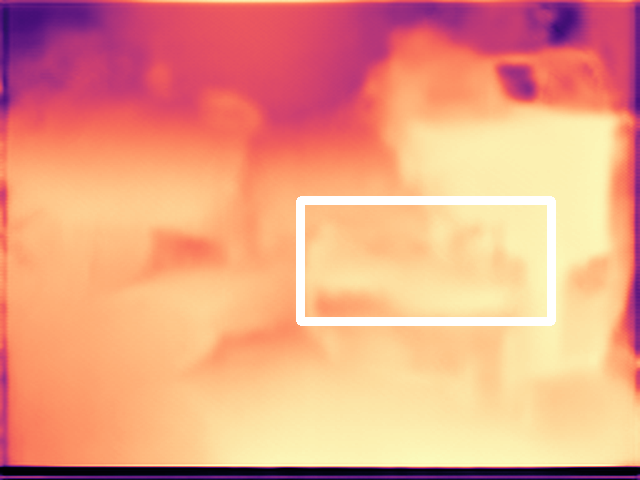} &
            \includegraphics[height=3cm,width=4cm]{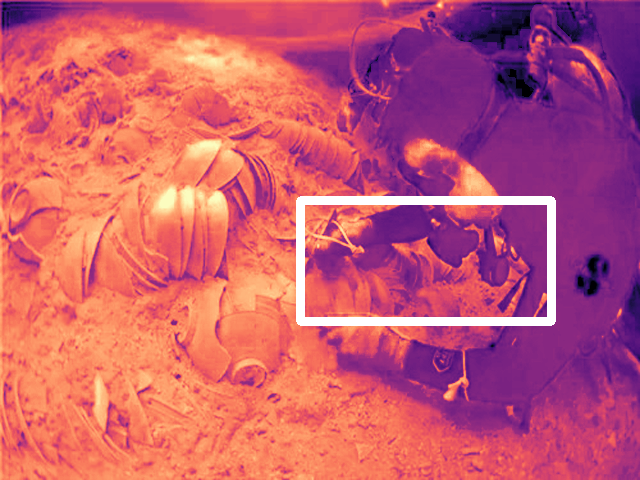} &
            \includegraphics[height=3cm,width=4cm]{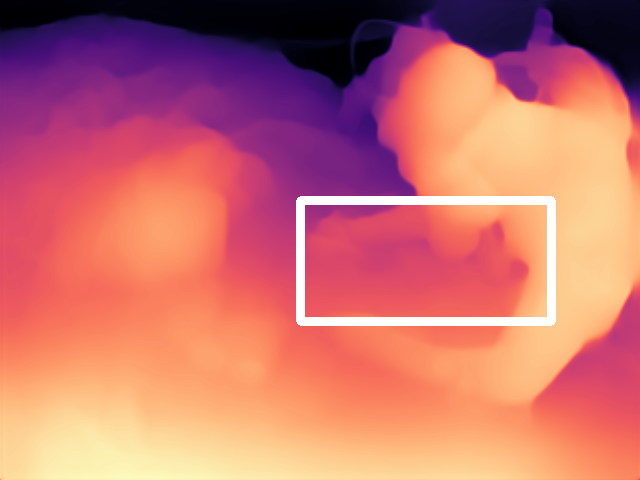} &
            \includegraphics[height=3cm,width=4cm]{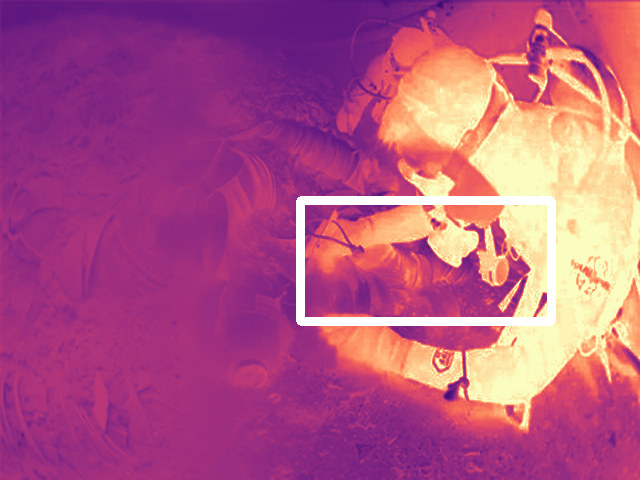} &
            \includegraphics[height=3cm,width=4cm]{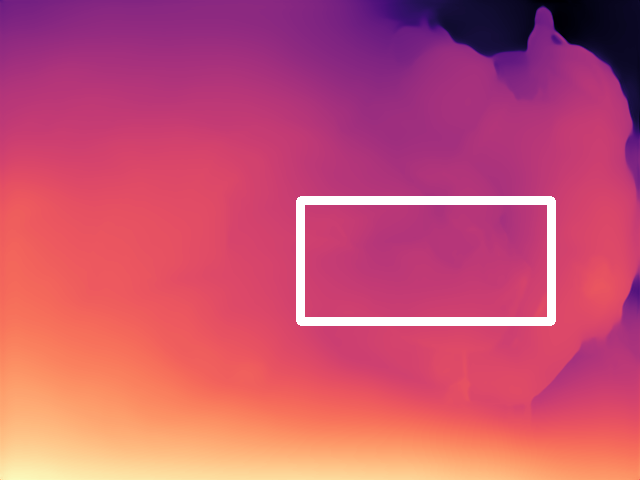} \\
            
            \includegraphics[height=2cm,width=4cm]{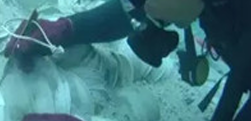} &
            \includegraphics[height=2cm,width=4cm]{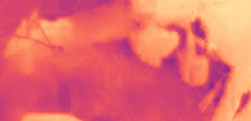} &
            \includegraphics[height=2cm,width=4cm]{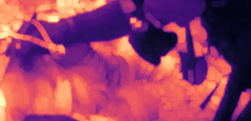} &
            \includegraphics[height=2cm,width=4cm]{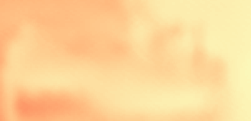} &
            \includegraphics[height=2cm,width=4cm]{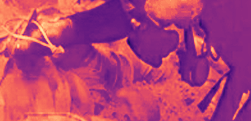} &
            \includegraphics[height=2cm,width=4cm]{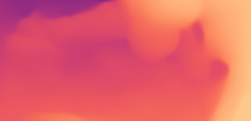} &
            \includegraphics[height=2cm,width=4cm]{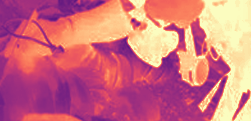} &
            \includegraphics[height=2cm,width=4cm]{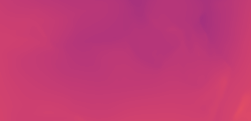} \\

            \includegraphics[height=3cm,width=4cm]{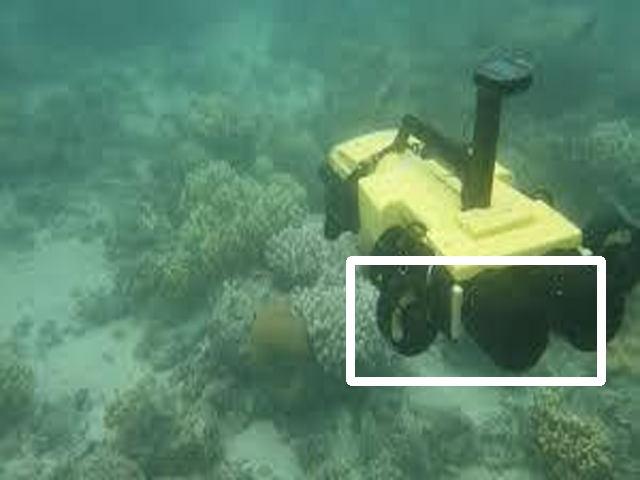} &
            \includegraphics[height=3cm,width=4cm]{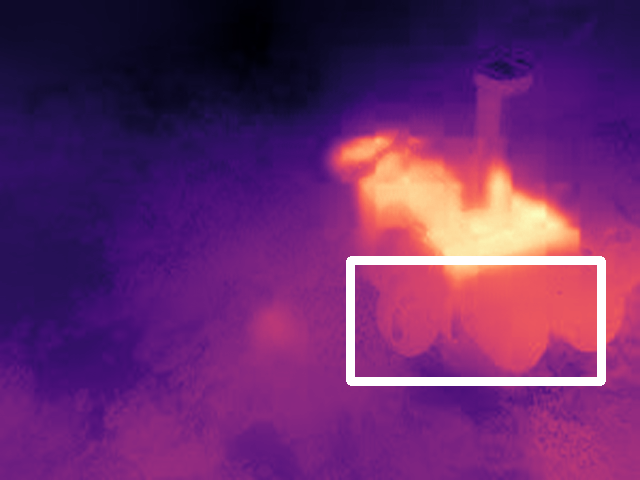} &
            \includegraphics[height=3cm,width=4cm]{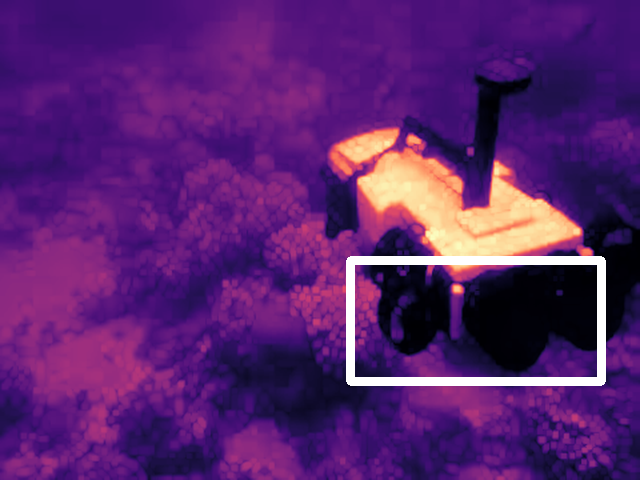} &
            \includegraphics[height=3cm,width=4cm]{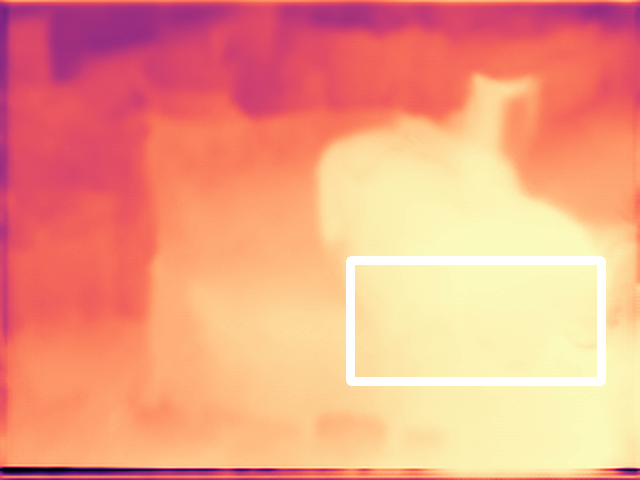} &
            \includegraphics[height=3cm,width=4cm]{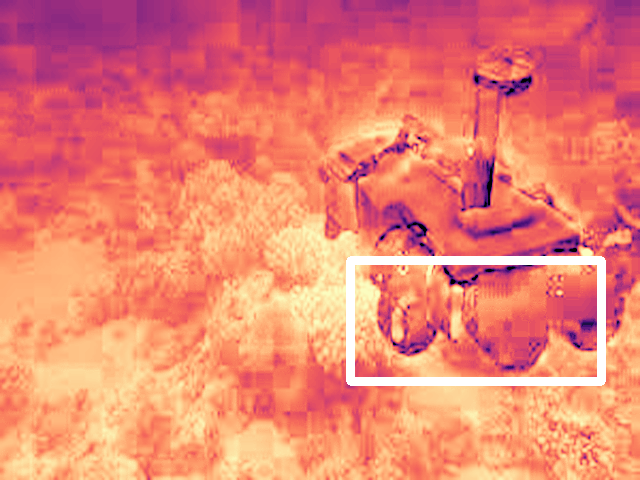} &
            \includegraphics[height=3cm,width=4cm]{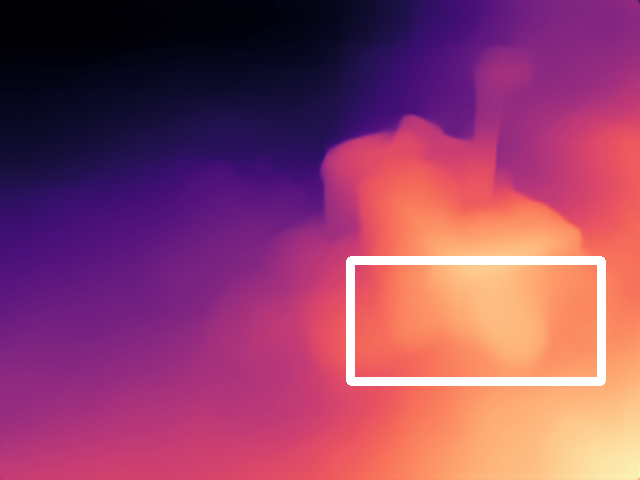} &
            \includegraphics[height=3cm,width=4cm]{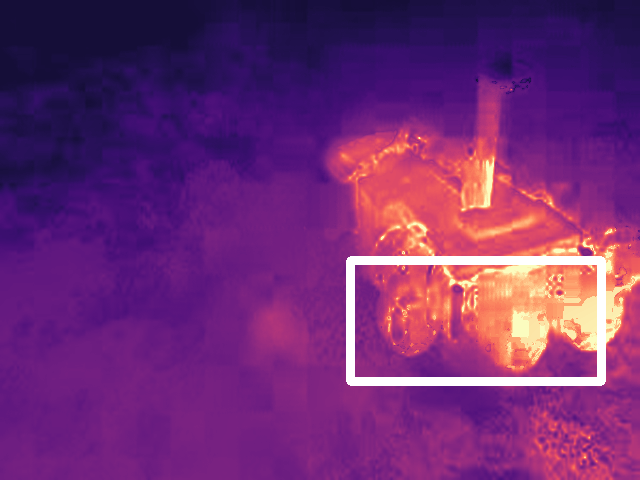} &
            \includegraphics[height=3cm,width=4cm]{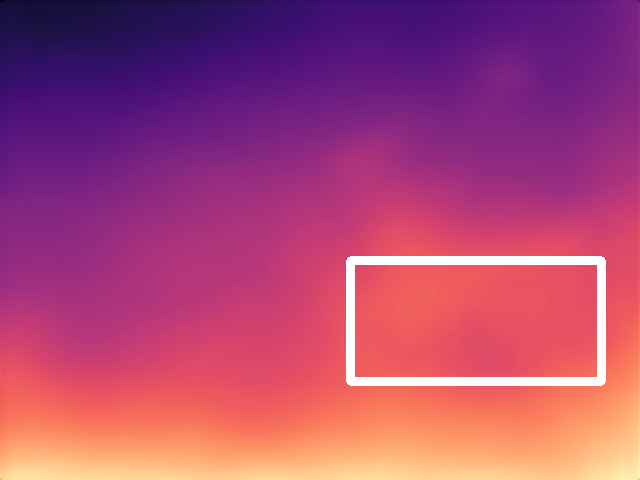} \\

            \includegraphics[height=2cm,width=4cm]{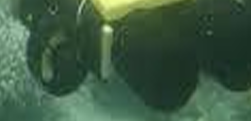} &
            \includegraphics[height=2cm,width=4cm]{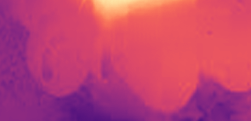} &
            \includegraphics[height=2cm,width=4cm]{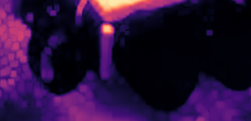} &
            \includegraphics[height=2cm,width=4cm]{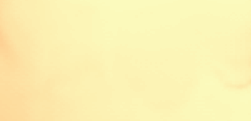} &
            \includegraphics[height=2cm,width=4cm]{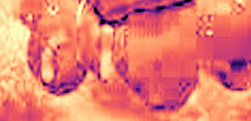} &
            \includegraphics[height=2cm,width=4cm]{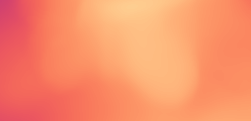} &
            \includegraphics[height=2cm,width=4cm]{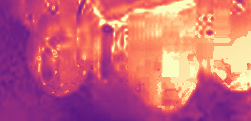} &
            \includegraphics[height=2cm,width=4cm]{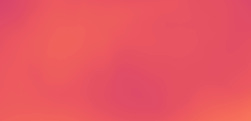} \\
            
            (a) Input & 
            (b) IBLA \cite{IBLA} &
            (c) GDCP \cite{GDCP} & 
            (d) UW-Net \cite{UW-Net} & 
            (e) NUDCP \cite{NUDCP} & 
            (f) UW-GAN \cite{UW-GAN} & 
            (g) HazeLine \cite{HazeLine} & 
            (h) MiDas \cite{MiDas} \\
                
            \includegraphics[height=3cm,width=4cm]{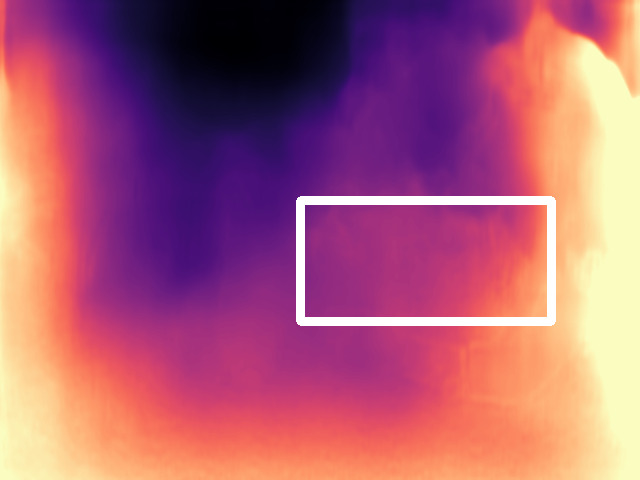} &
            \includegraphics[height=3cm,width=4cm]{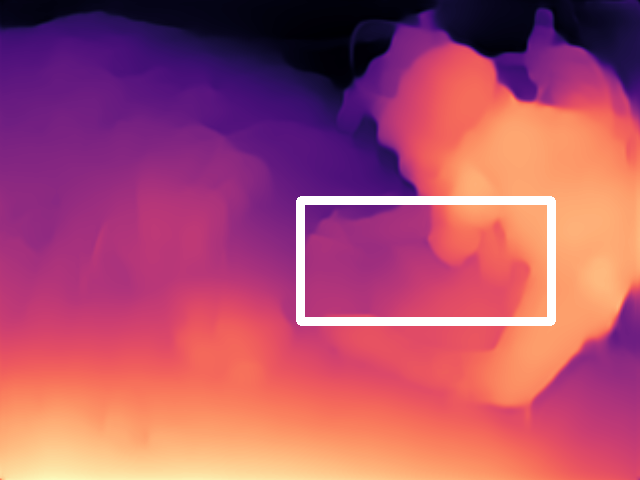} &
            \includegraphics[height=3cm,width=4cm]{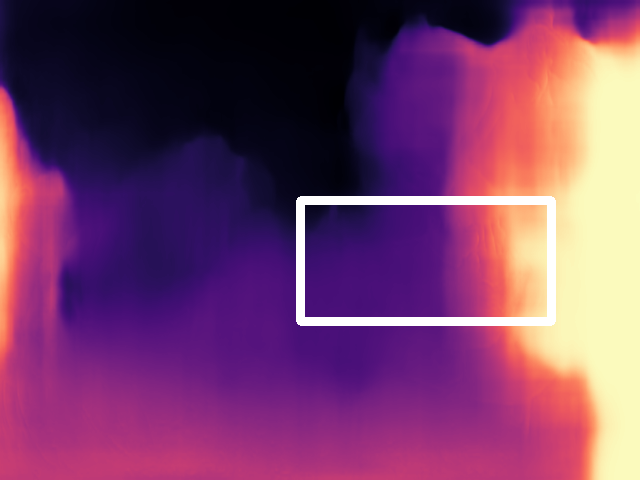} &
            \includegraphics[height=3cm,width=4cm]{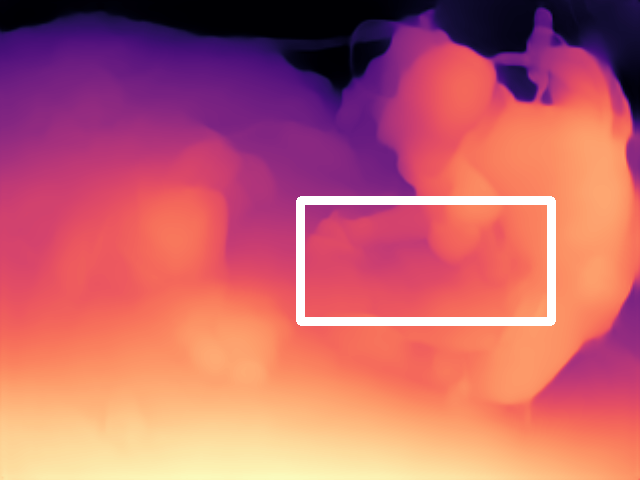} &
            \includegraphics[height=3cm,width=4cm]{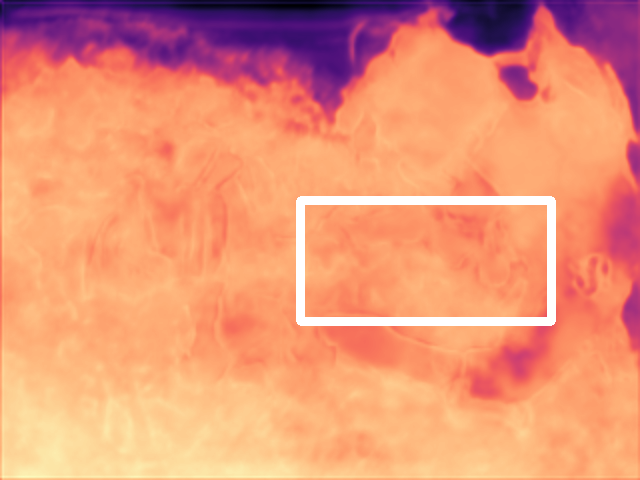} &
            \includegraphics[height=3cm,width=4cm]{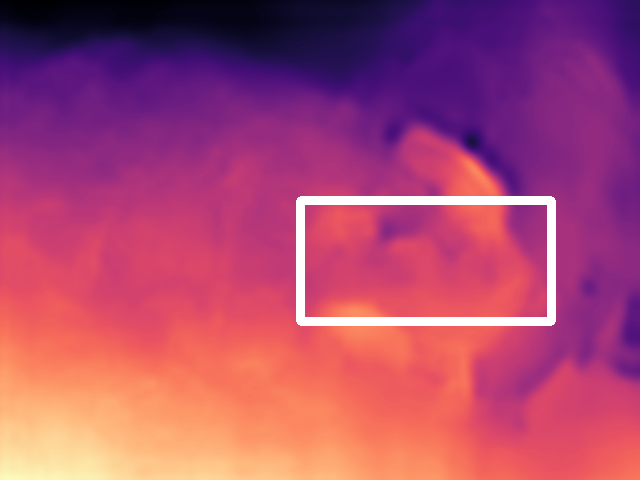} &
            \includegraphics[height=3cm,width=4cm]{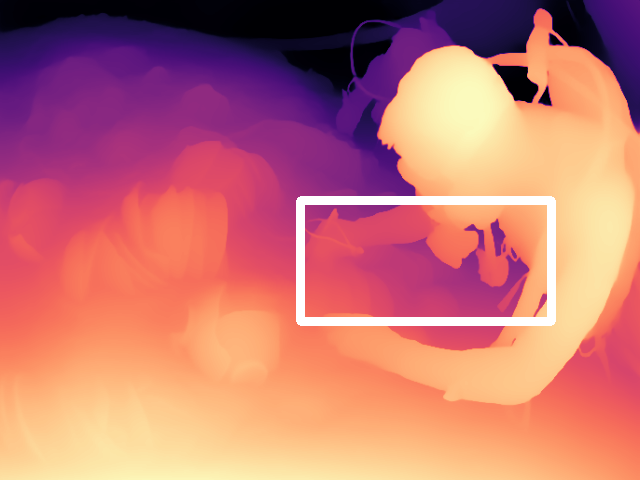} &
            \includegraphics[height=3cm,width=4cm]{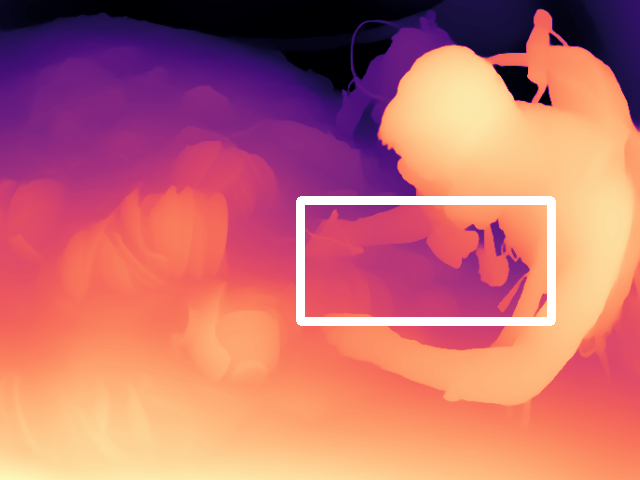} \\	

            \includegraphics[height=2cm,width=4cm]{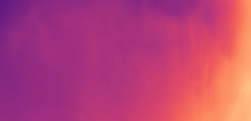} &
            \includegraphics[height=2cm,width=4cm]{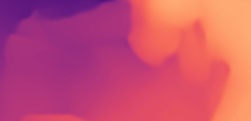} &
            \includegraphics[height=2cm,width=4cm]{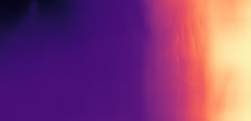} &
            \includegraphics[height=2cm,width=4cm]{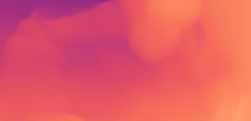} &
            \includegraphics[height=2cm,width=4cm]{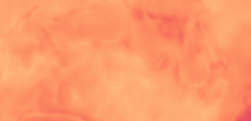} &
            \includegraphics[height=2cm,width=4cm]{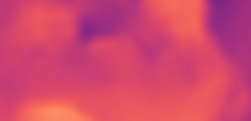} &
            \includegraphics[height=2cm,width=4cm]{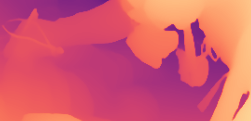} &
            \includegraphics[height=2cm,width=4cm]{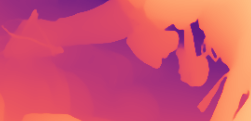} \\

            \includegraphics[height=3cm,width=4cm]{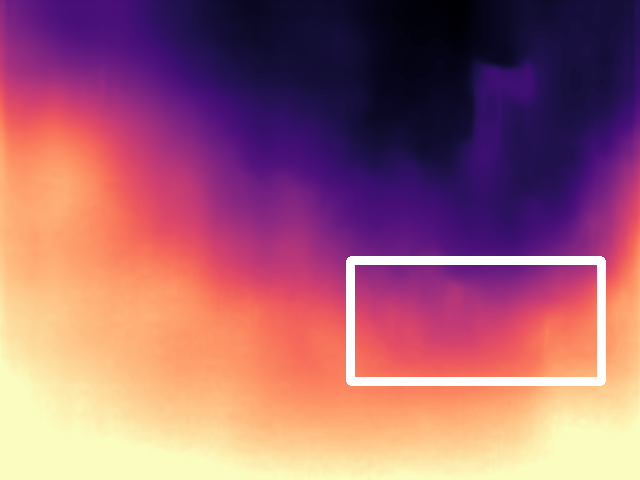} &
            \includegraphics[height=3cm,width=4cm]{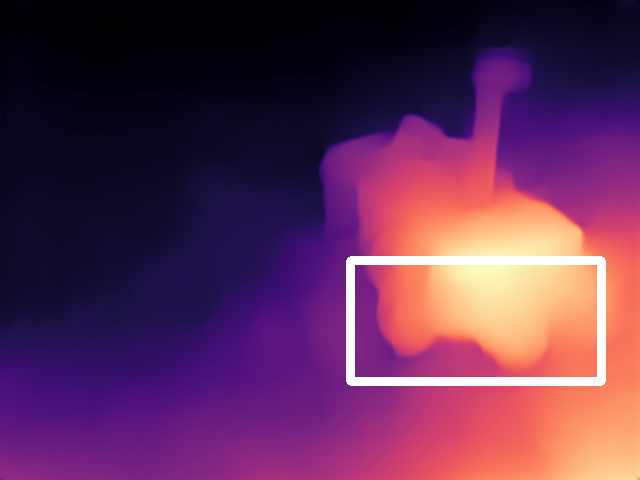} &
            \includegraphics[height=3cm,width=4cm]{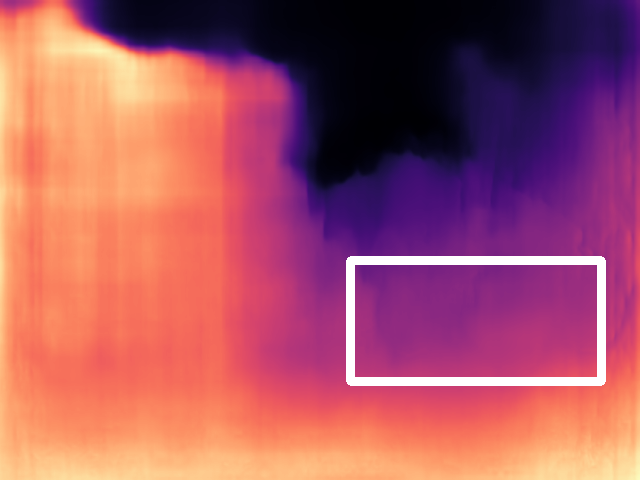} &
            \includegraphics[height=3cm,width=4cm]{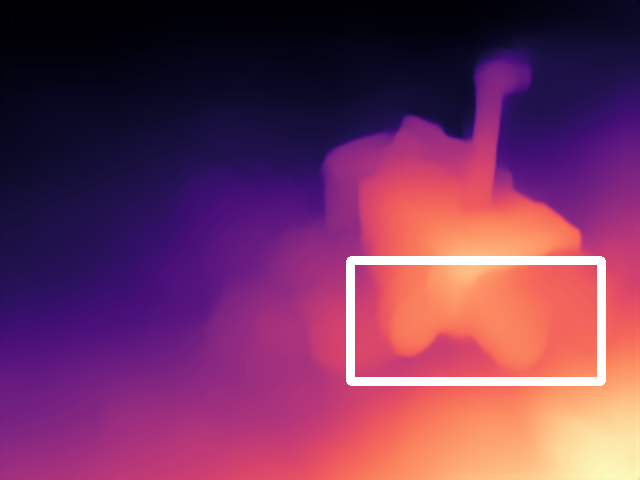} &
            \includegraphics[height=3cm,width=4cm]{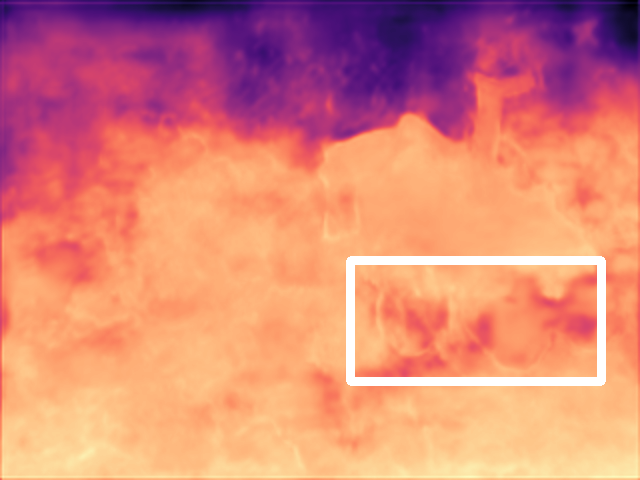} &
            \includegraphics[height=3cm,width=4cm]{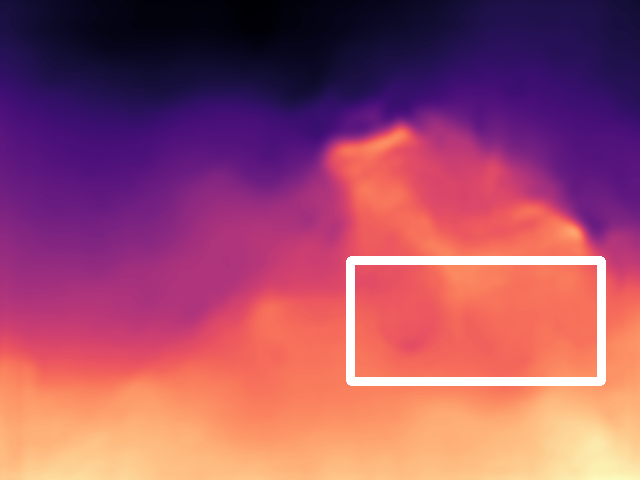} &
            \includegraphics[height=3cm,width=4cm]{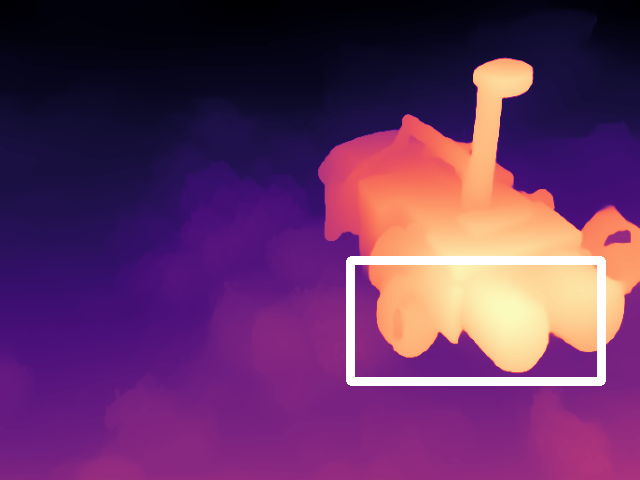} &
            \includegraphics[height=3cm,width=4cm]{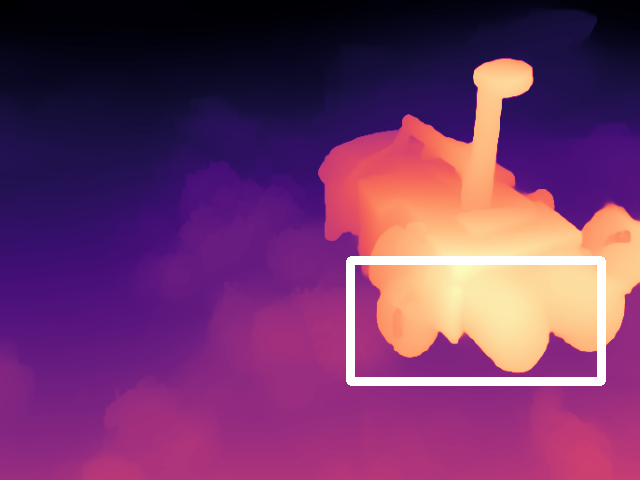} \\	

            \includegraphics[height=2cm,width=4cm]{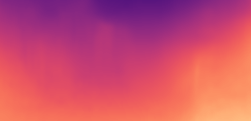} &
            \includegraphics[height=2cm,width=4cm]{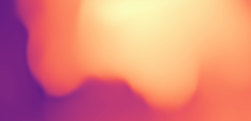} &
            \includegraphics[height=2cm,width=4cm]{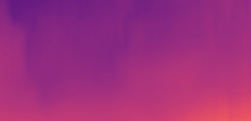} &
            \includegraphics[height=2cm,width=4cm]{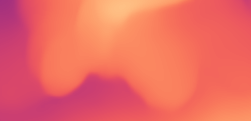} &
            \includegraphics[height=2cm,width=4cm]{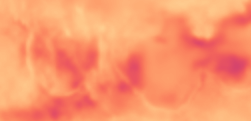} &
            \includegraphics[height=2cm,width=4cm]{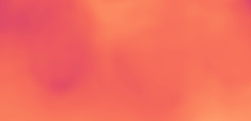} &
            \includegraphics[height=2cm,width=4cm]{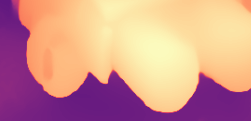} &
            \includegraphics[height=2cm,width=4cm]{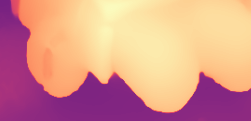} \\
            
            (i) Lite-Mono \cite{Lite-Mono} & 
            (j) UDepth \cite{UDepth} &
            (k) ADPCC \cite{ADPCC} & 
            (l) UW-Depth \cite{UW-Depth} & 
            (m) WsUID-Net \cite{SUIM-SDA} & 
            (n) WaterMono \cite{WaterMono} & 
            (o) Tree-Mamba & 
            (p) Reference \\
        \end{tabular}
    }
    \caption{Visual comparison of different methods on other color degradation underwater images from \textbf{Test-FR5691}.  Compared with other competitors, our Tree-Mamba method yields better depth results on different degraded underwater images, and our depth maps are closer to the reference images.}
    \label{Qual_O}
\end{figure*}

\noindent
\textbf{Evaluation Metrics.} 
We use the standard error and accuracy metrics \cite{SI_Log, UDepth, UW-Depth, WaterMono} to assess the performance of different methods on \textbf{Test-FR5691} and \textbf{Test-FS1941}.
The standard error metrics include root mean square error ($\mathrm{RMSE}$) and its log variant ($\mathrm{RMSE_{log}}$), absolute ($\mathrm{A.Rel}$) and square ($\mathrm{S.Rel}$) mean relative error, and absolute error in log-scale ($\mathrm{log_{10}}$).
Lower scores on these error metrics indicate better prediction performance.
The accuracy metrics contain the percentage of inlier pixels with threshold $(\delta _i < 1.25^i,i=1,2,3)$.
Higher $\delta _i$ scores denote better prediction performance.

\begin{figure*}[!htp]
    \Large
    \centering
    \resizebox{0.84\linewidth}{!}{
        \begin{tabular}{c@{ }c@{ }c@{ }c@{ }c@{ }c@{ }c@{ }c@{ }}
            \includegraphics[height=3cm,width=4cm]{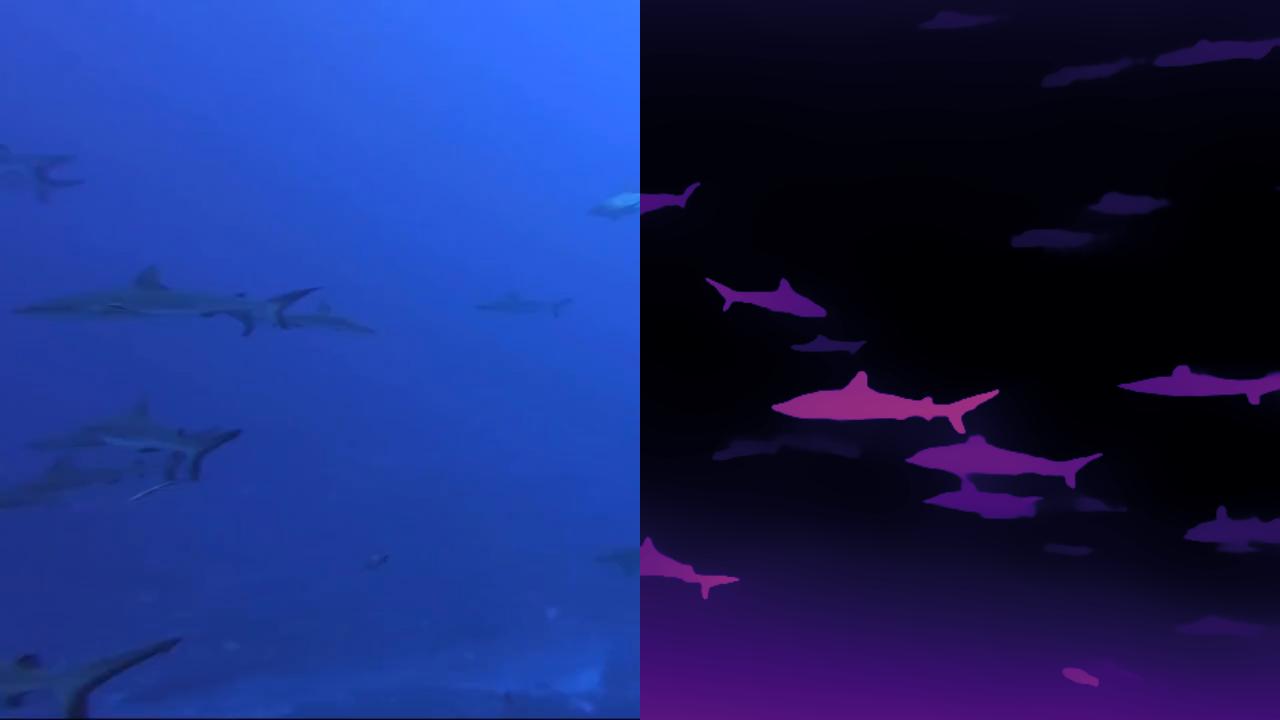} &
            \includegraphics[height=3cm,width=4cm]{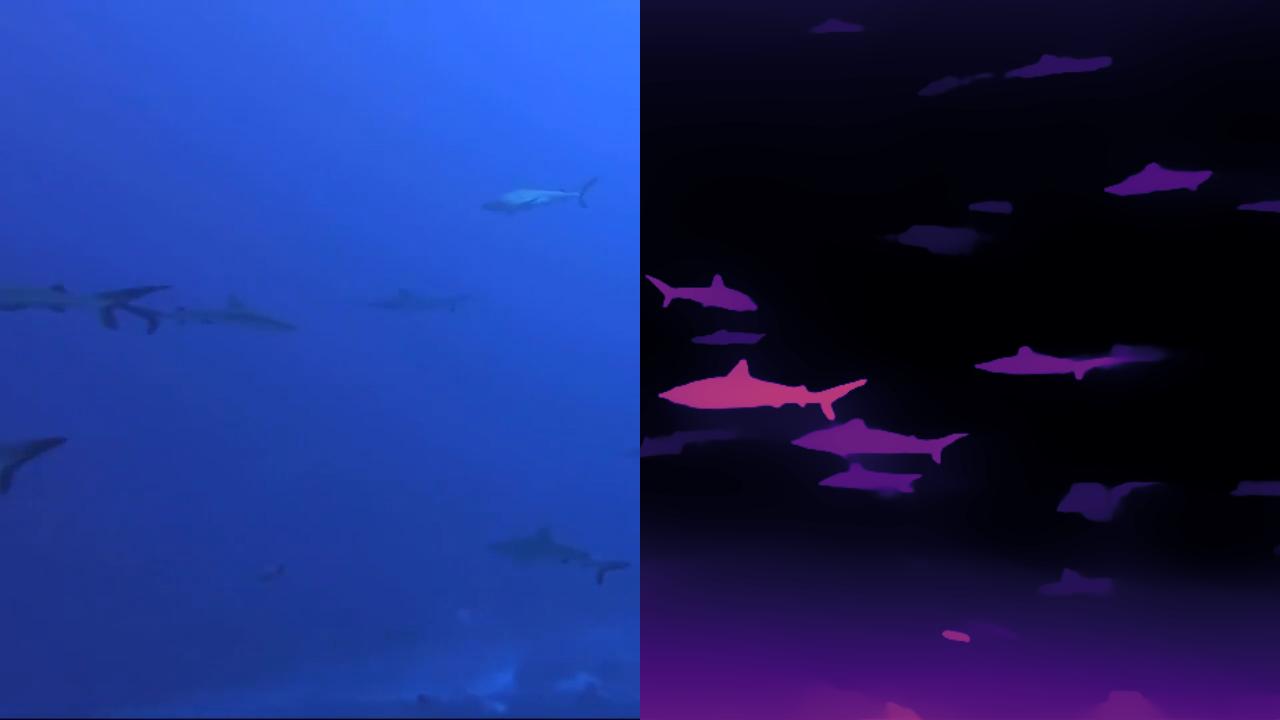} &
            
            \includegraphics[height=3cm,width=4cm]{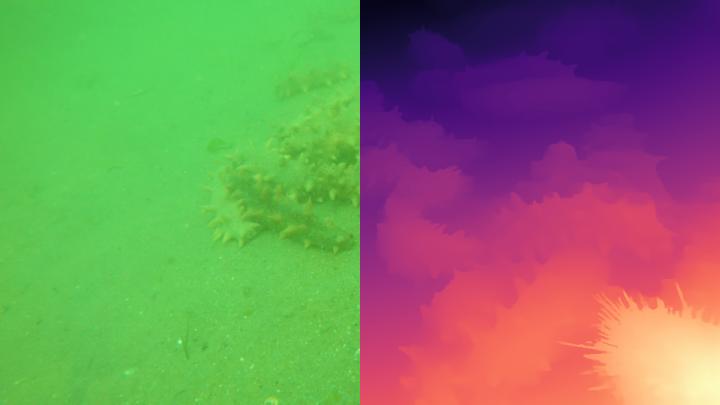} &
            \includegraphics[height=3cm,width=4cm]{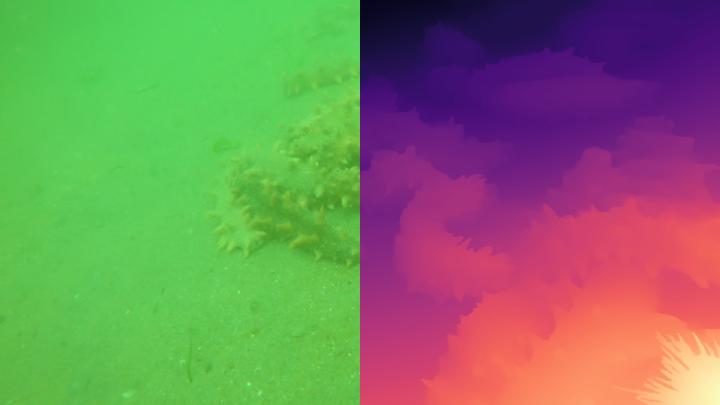} &
            
            \includegraphics[height=3cm,width=4cm]{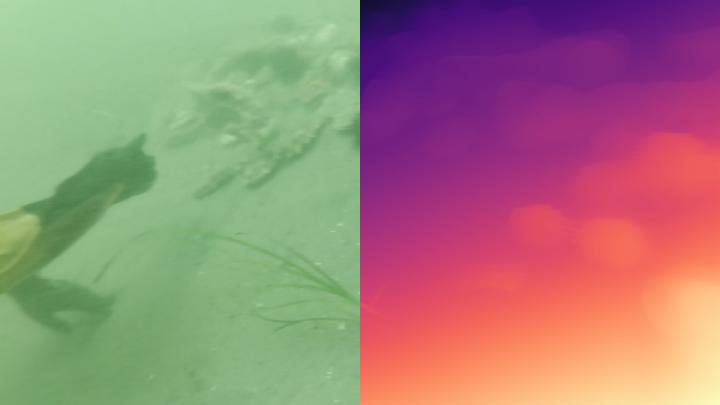} &
            \includegraphics[height=3cm,width=4cm]{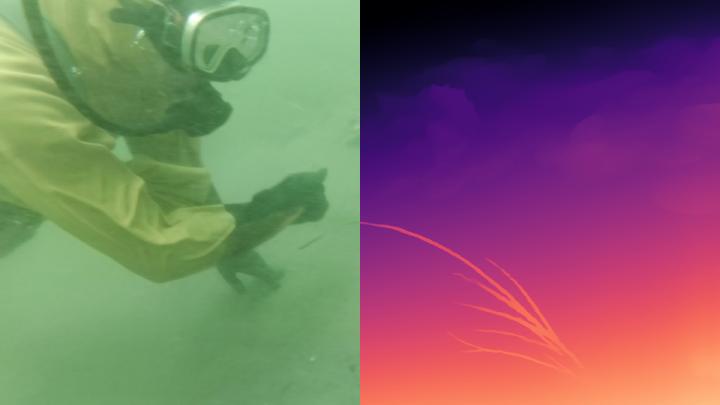} &
            
            \includegraphics[height=3cm,width=4cm]{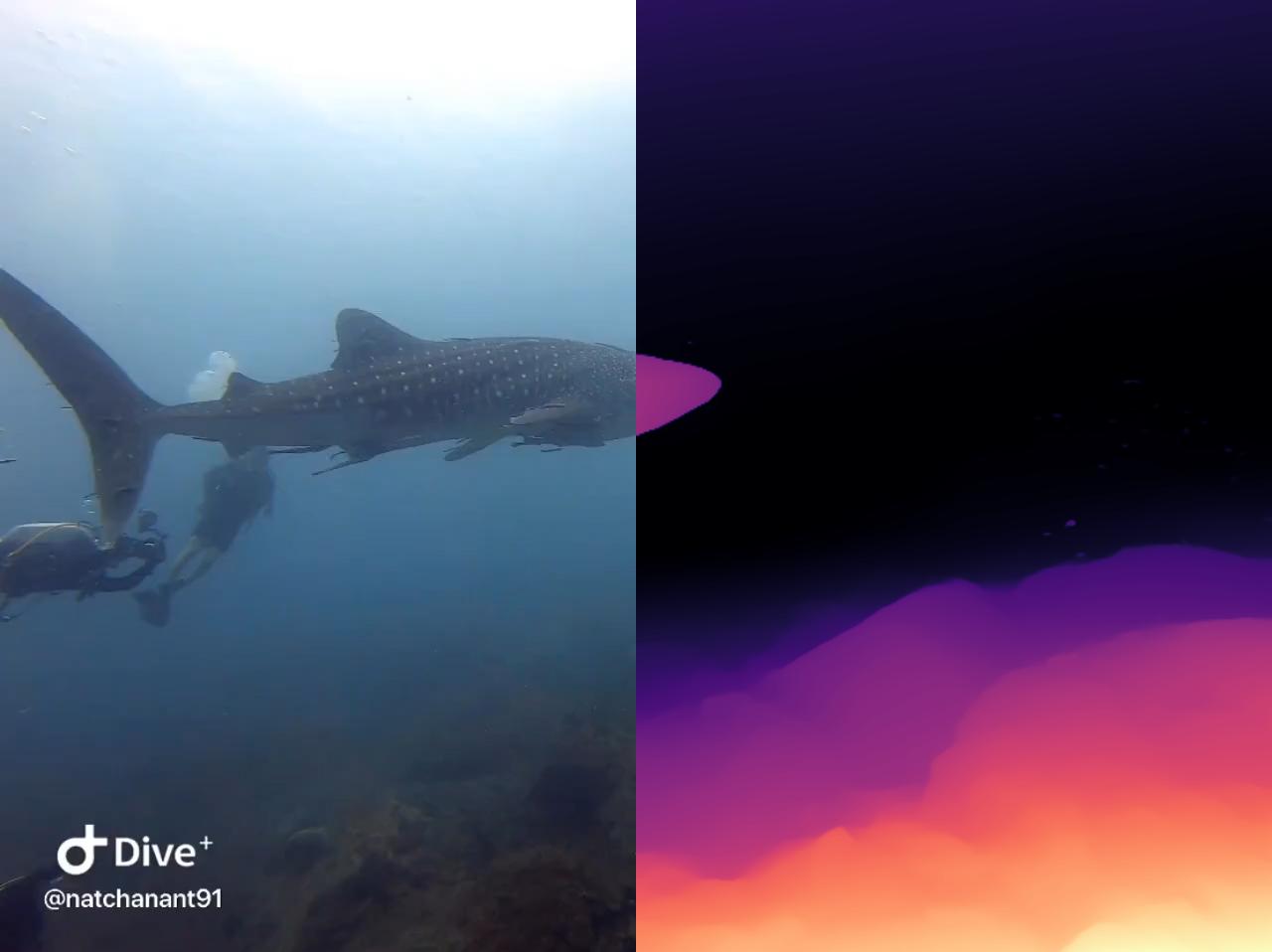} &
            \includegraphics[height=3cm,width=4cm]{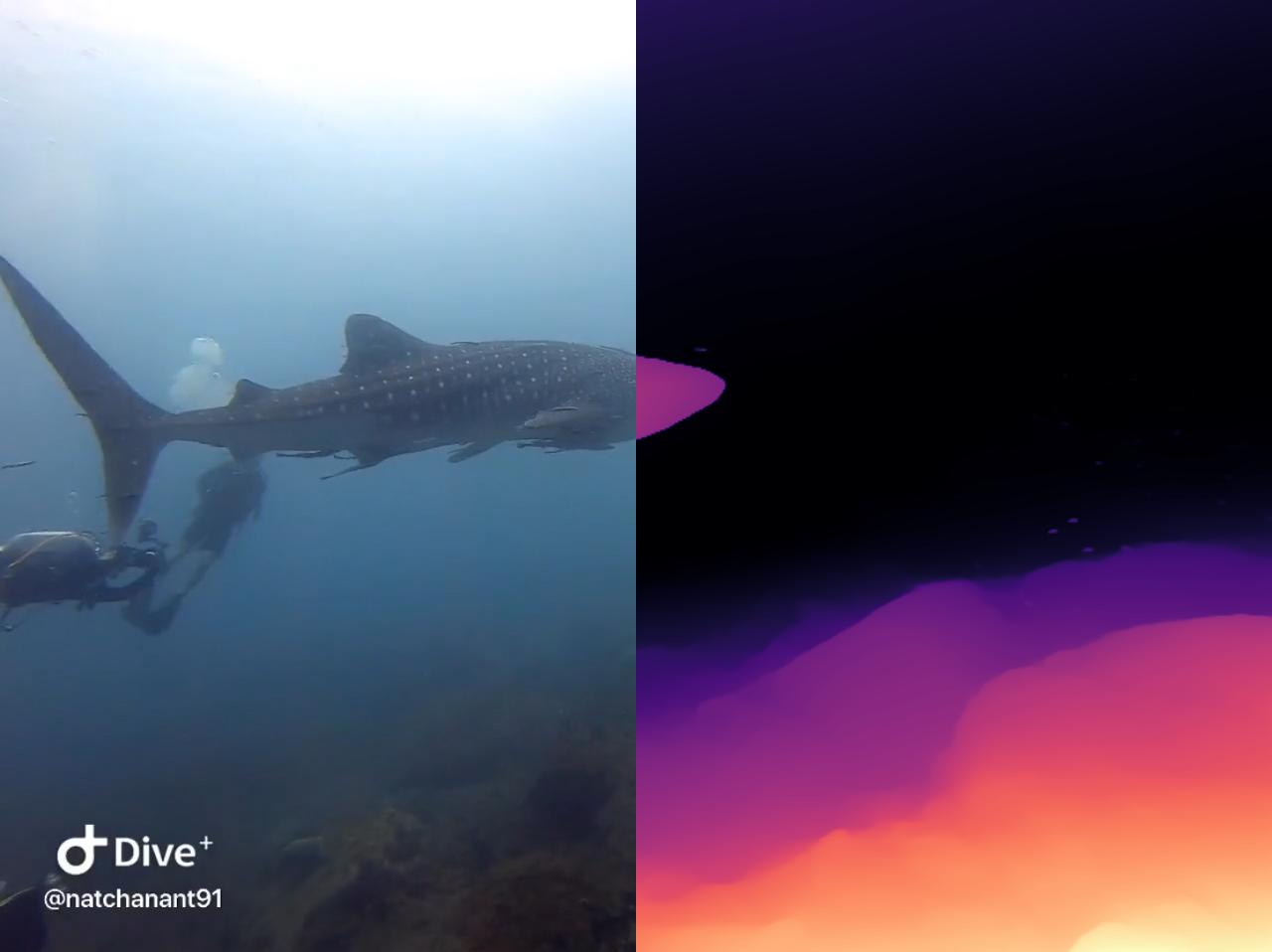} \\

            (a) \#1 & 
            (b) \#11  &
            (c) \#401 & 
            (d) \#411 & 
            (e) \#801 & 
            (f) \#811 & 
            (g) \#1201 & 
            (h) \#1211 \\
           
        \end{tabular}
    }
    \caption{Underwater video frames from \textbf{Video-NR1600} predicted by our Tree-Mamba method. The left half shows raw underwater images, while the right half displays predicted depth maps. The predicted depth maps exhibit noticeable consistency across frames, which highlights the scalability of our Tree-Mamba.}
    \label{Qual_V}
\end{figure*}

\begin{figure*}[!htp]
    \centering
    \resizebox{0.84\linewidth}{!}{
        \begin{tabular}{c@{ }c@{ }c@{ }}
            \includegraphics[height=2cm,width=7cm]{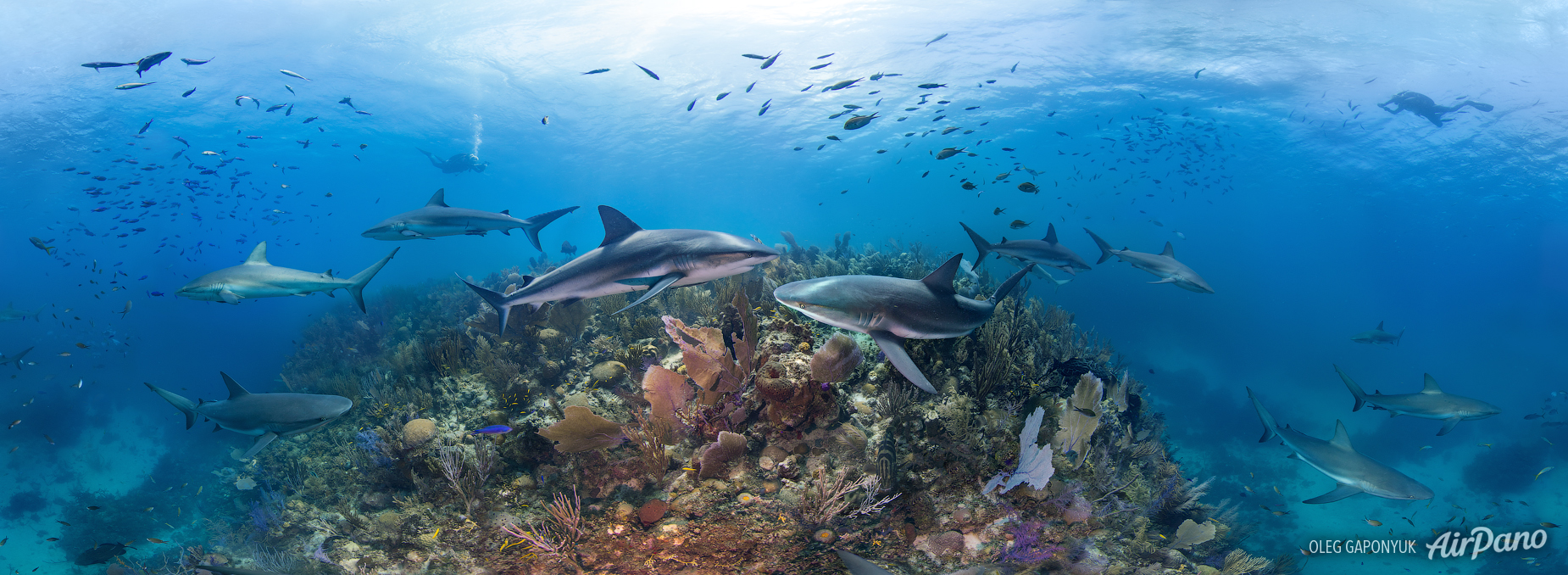} &
            \includegraphics[height=2cm,width=7cm]{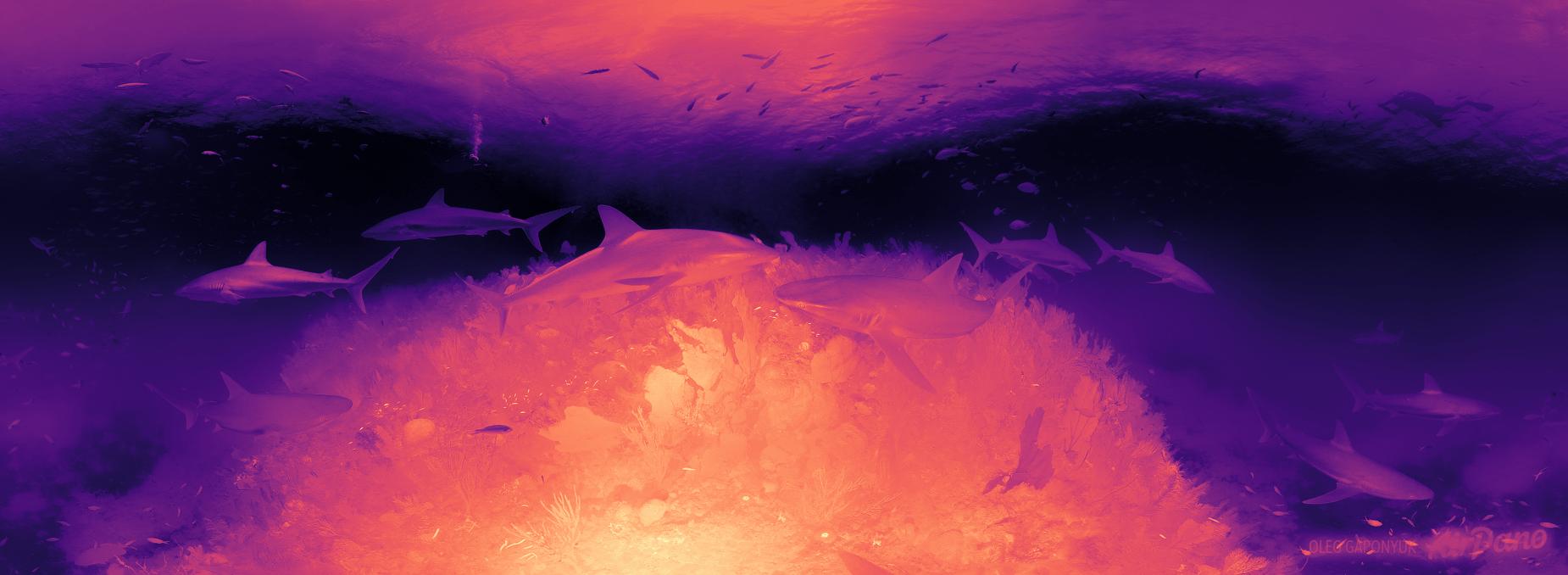} &
            \includegraphics[height=2cm,width=7cm]{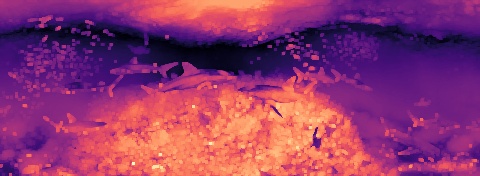} \\
            (a) Input $\mid$ Quality score (↑) & 
            (b) IBLA \cite{IBLA} $\mid$ 41.29 &
            (c) GDCP \cite{GDCP} $\mid$ 35.10 \\
            
            \includegraphics[height=2cm,width=7cm]{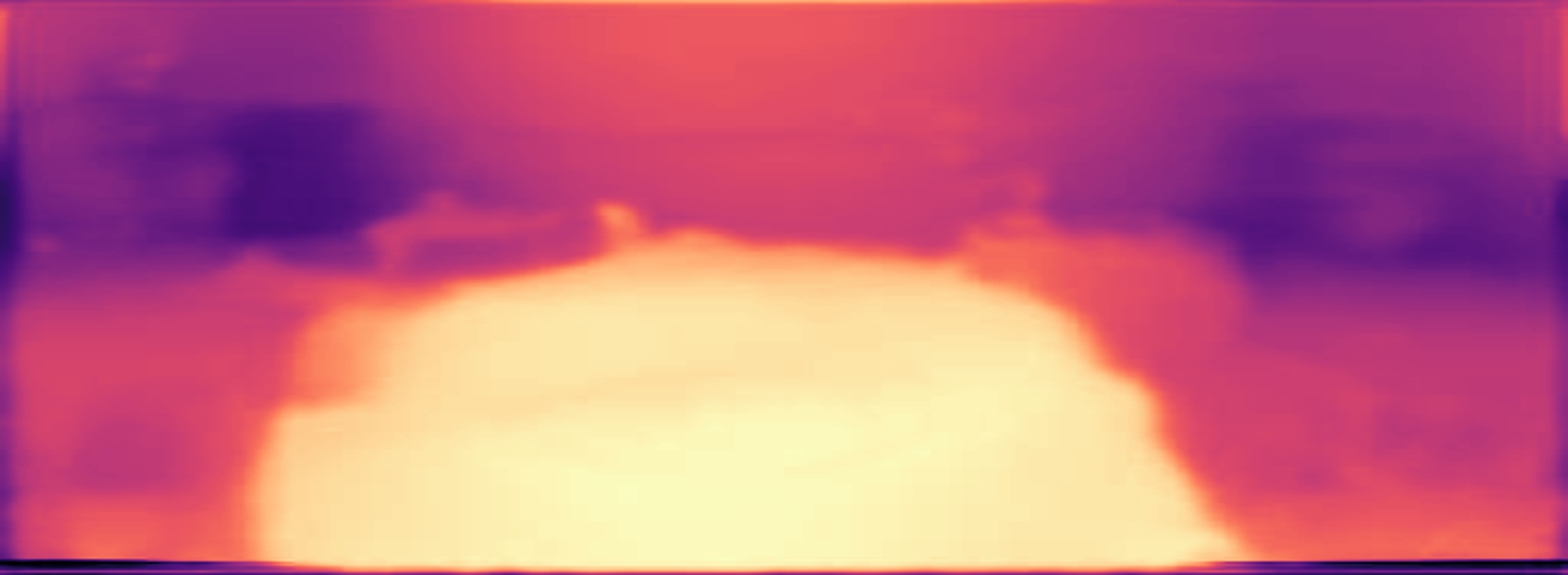} &
            \includegraphics[height=2cm,width=7cm]{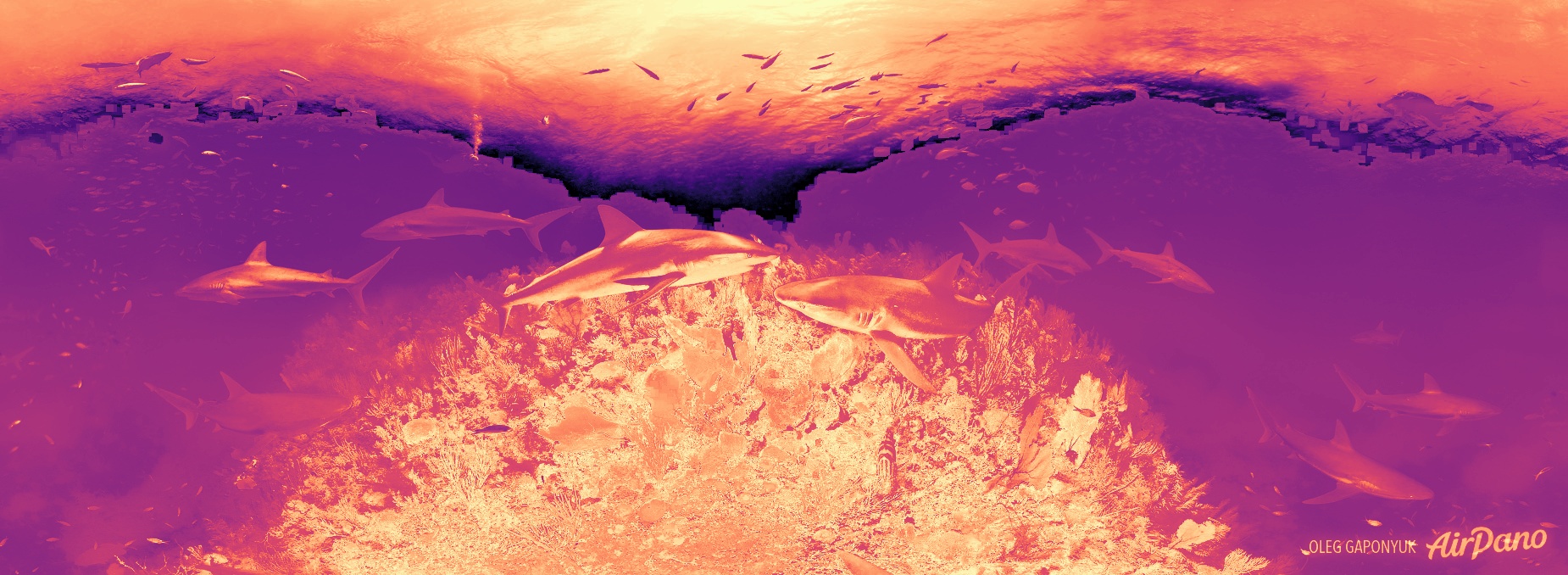} &
            \includegraphics[height=2cm,width=7cm]{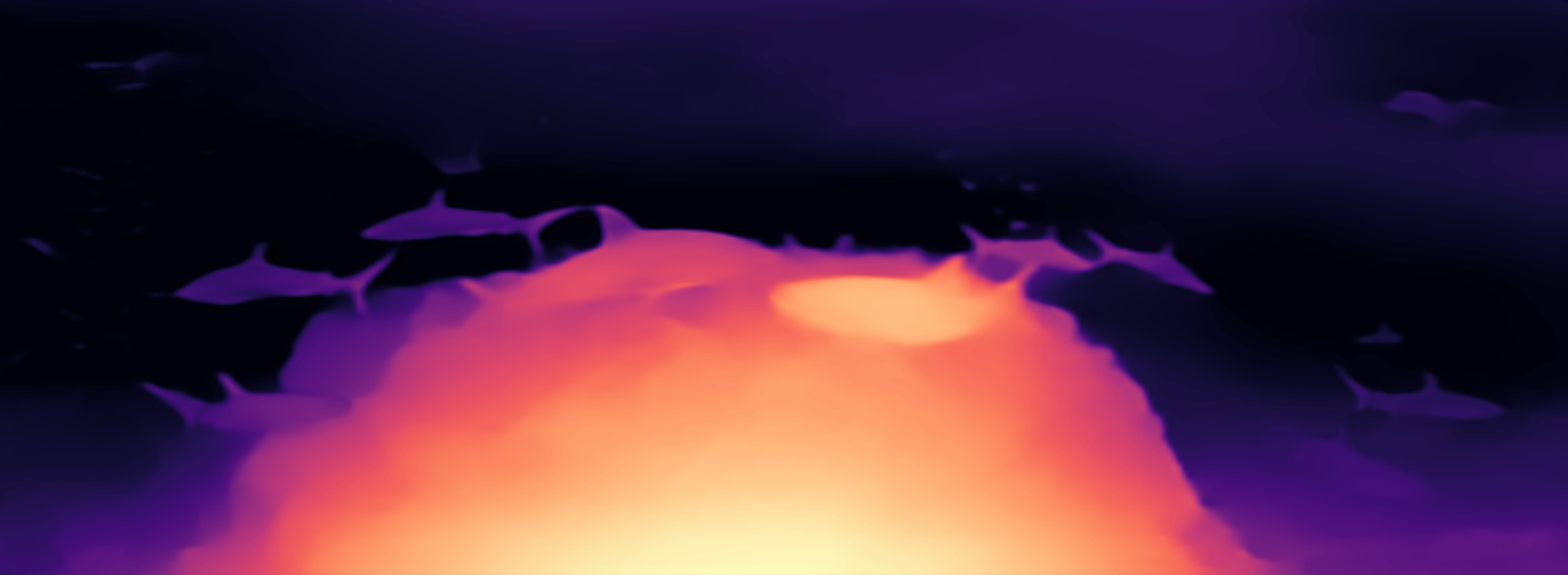} \\
            
            (d) UW-Net \cite{UW-Net} $\mid$ 43.66 & 
            (e) NUDCP \cite{NUDCP} $\mid$ 38.68 & 
            (f) UW-GAN \cite{UW-GAN} $\mid$ 79.19 \\
            
            \includegraphics[height=2cm,width=7cm]{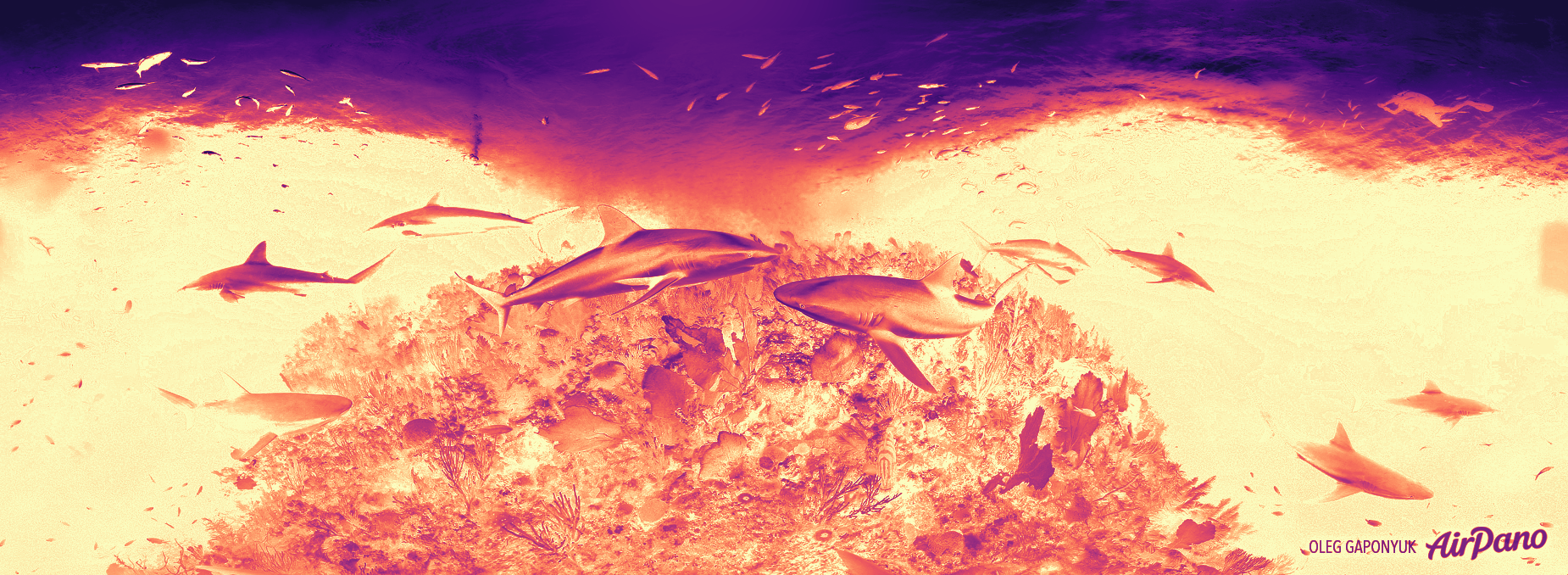} &
            \includegraphics[height=2cm,width=7cm]{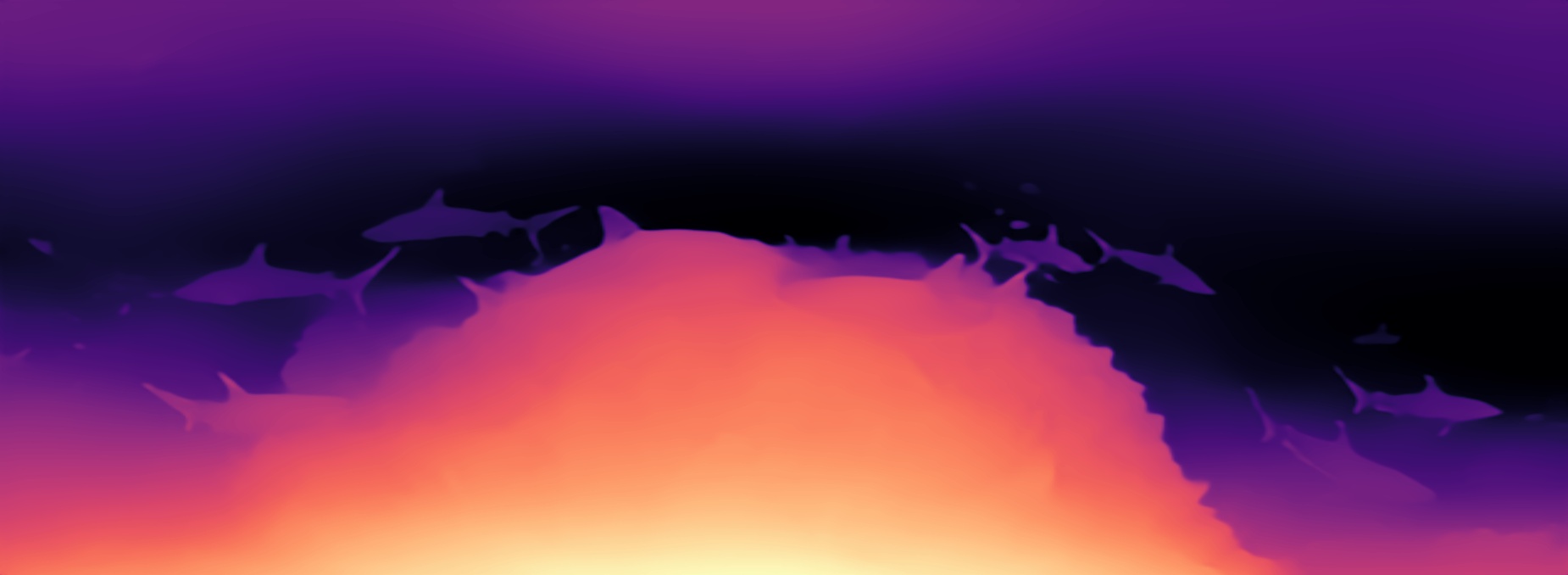} &
            \includegraphics[height=2cm,width=7cm]{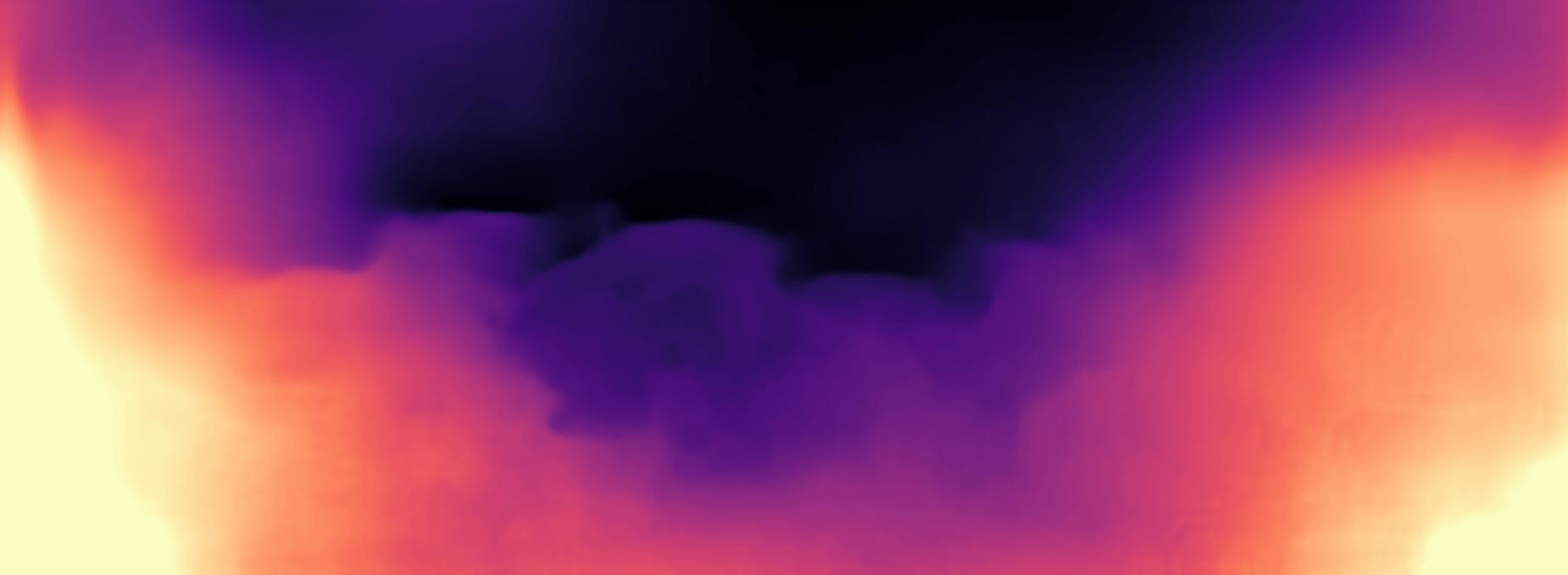} \\
            
            (g) HazeLine \cite{HazeLine} $\mid$ 33.85 & 
            (h) MiDas \cite{MiDas} $\mid$ 67.60 &
            (i) Lite-Mono \cite{Lite-Mono} $\mid$ 23.63 \\

            \includegraphics[height=2cm,width=7cm]{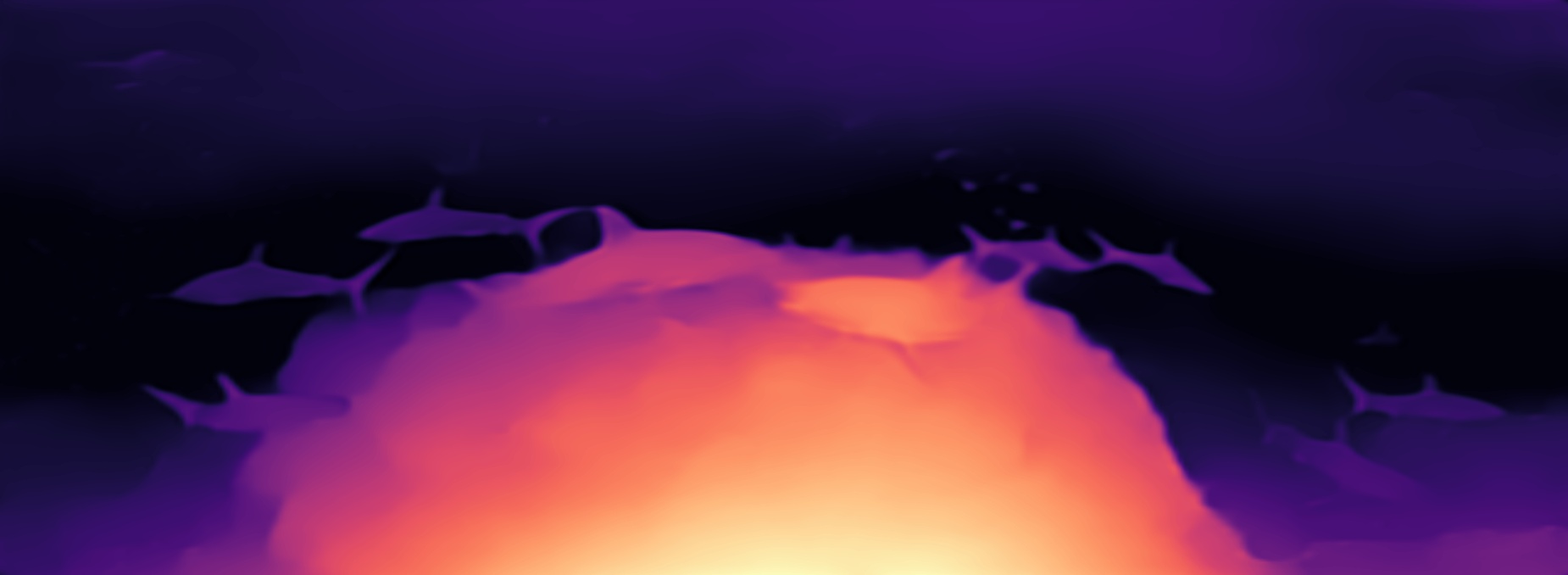} &
            \includegraphics[height=2cm,width=7cm]{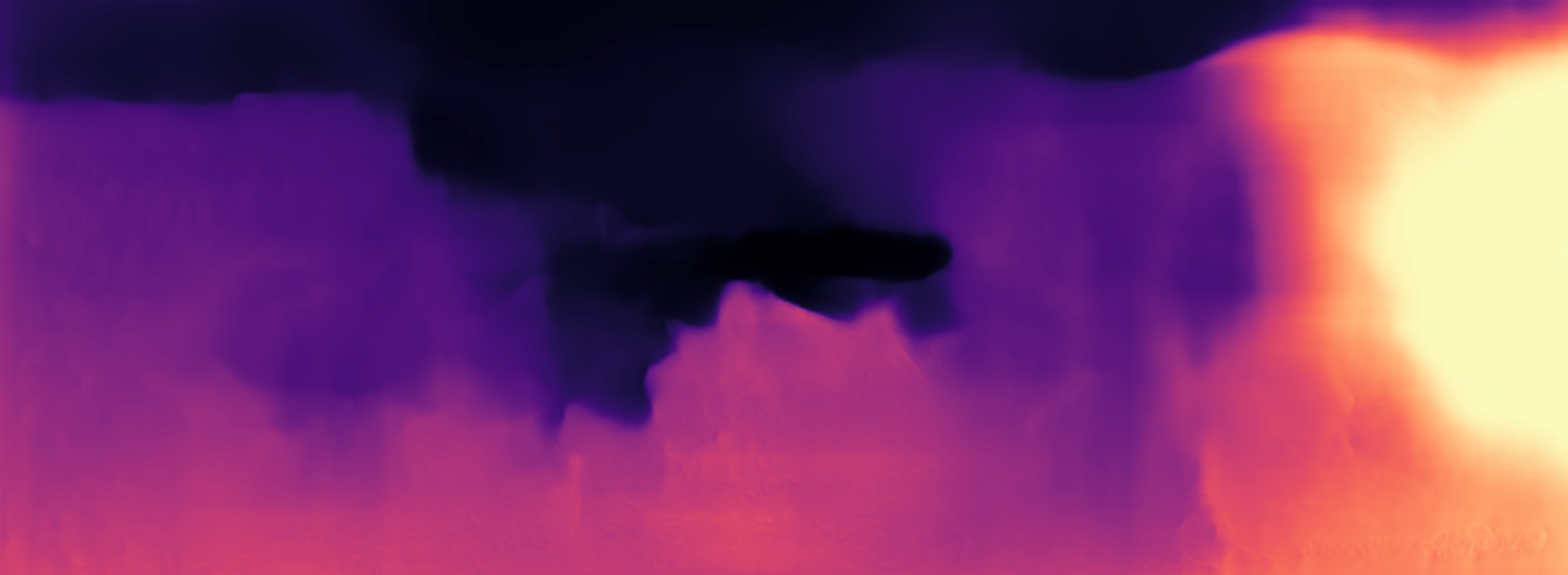} &
            \includegraphics[height=2cm,width=7cm]{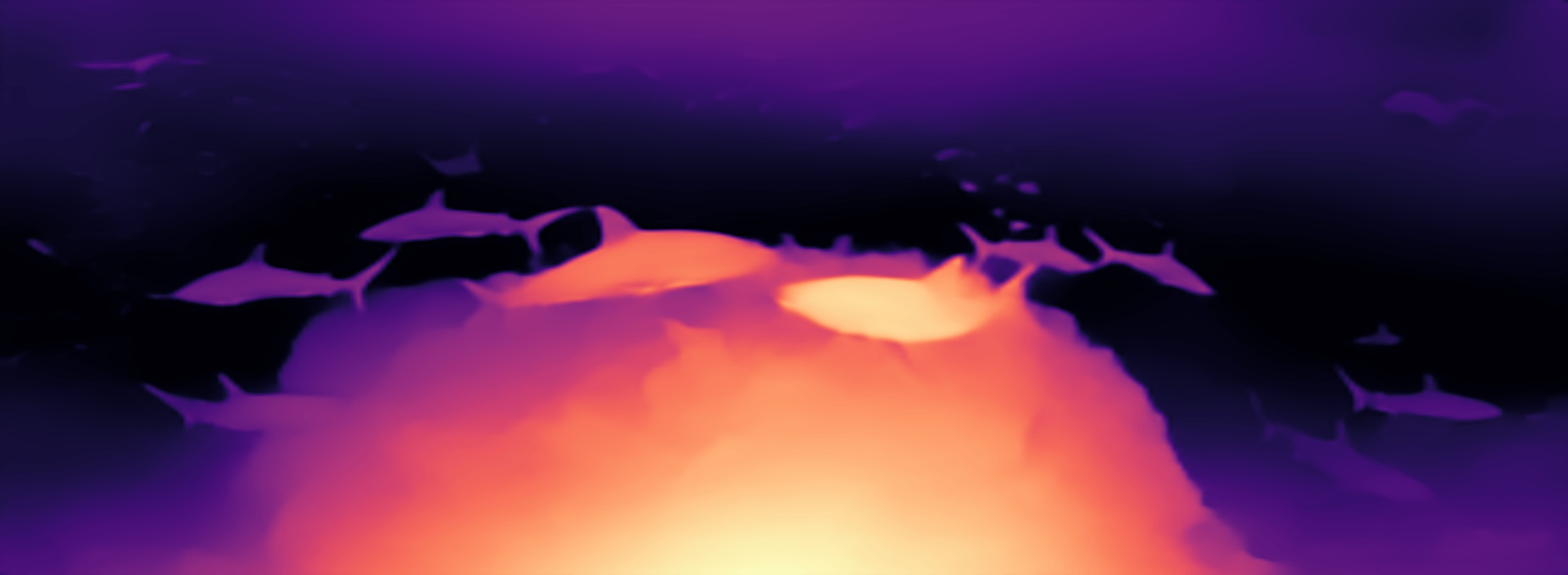} \\

            (j) UDepth \cite{UDepth} $\mid$ 78.79 &
            (k) ADPCC \cite{ADPCC} $\mid$ 22.13 & 
            (l) UW-Depth \cite{UW-Depth} $\mid$ 86.16\\
            
            \includegraphics[height=2cm,width=7cm]{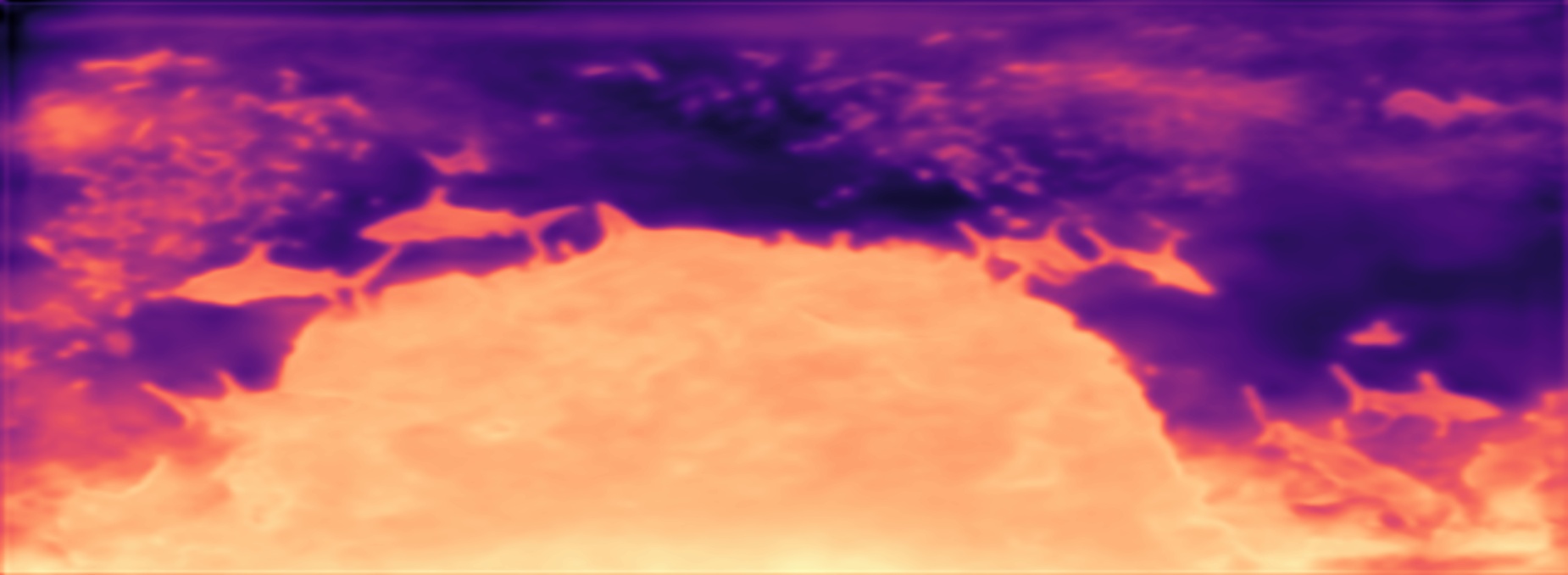} &
            \includegraphics[height=2cm,width=7cm]{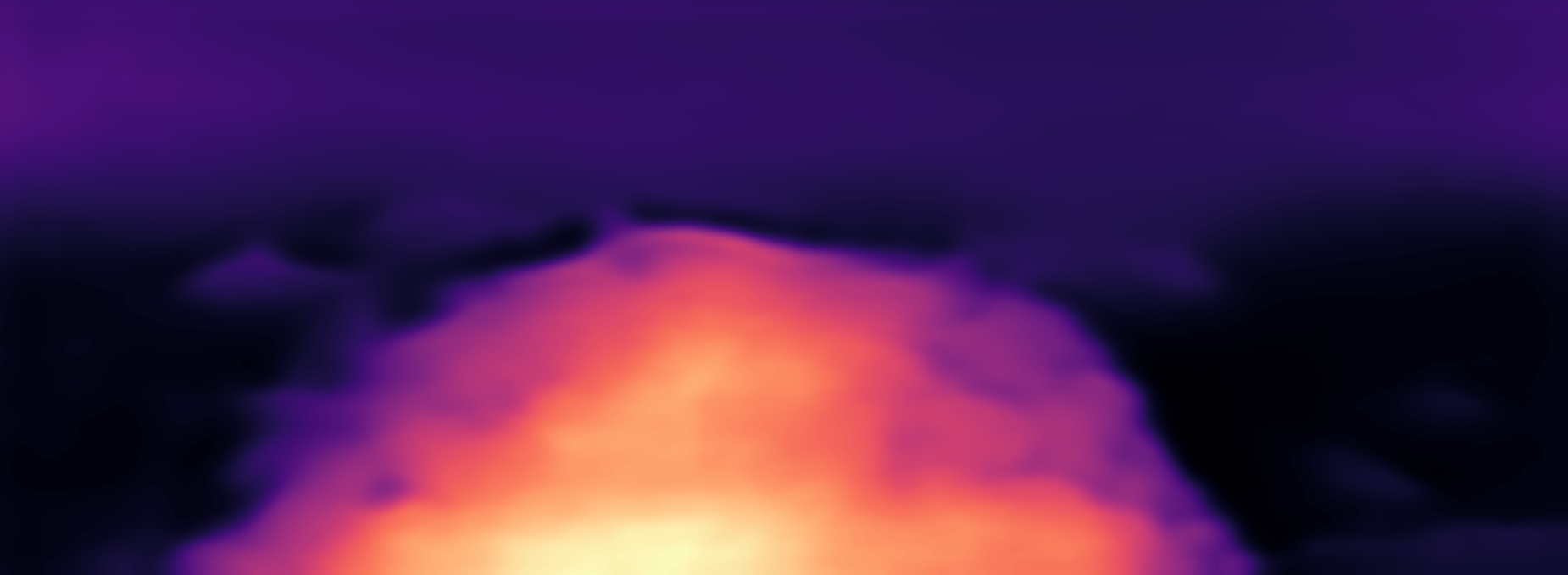} &
            \includegraphics[height=2cm,width=7cm]{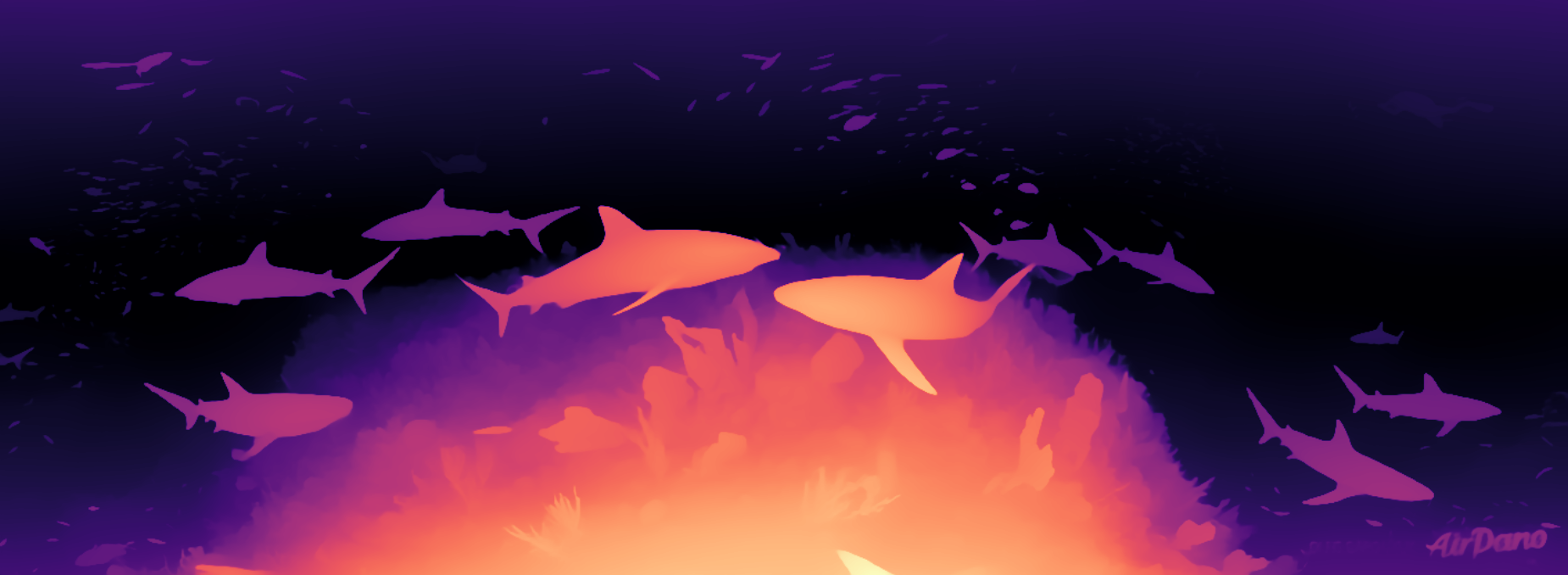} \\
            
            (m) WsUID-Net \cite{SUIM-SDA} $\mid$ 46.56 & 
            (n) WaterMono \cite{WaterMono} $\mid$ 52.76 & 
            (o) Tree-Mamba $\mid$ \textcolor{red}{97.74} \\
            
        \end{tabular}
    }
     \caption{Visual comparison of different methods on an underwater panoramic image. The quality score is evaluated by the fine-tuned LAR-IQA model \cite{LAR-IQA}. The best result is marked in red. Compared with other competitors, our Tree-Mamba method yields better results of both panoramic depth and quality score.}
    \label{Qual_P}
\end{figure*}

\begin{table*}[t]
\centering
\Large
\caption{Quantitative results of different methods on \textbf{Test-FR5691} and \textbf{Test-FS1941}. The \textcolor{red}{best} results are marked in red.}
\label{tab quan}
\resizebox{1\textwidth}{!}{

    \begin{tabular}{c|c|ccccc|ccc|cccccccc}
    \rowcolor[HTML]{FFCCC9} 
    \cellcolor[HTML]{FFCCC9} & \cellcolor[HTML]{FFCCC9} & \multicolumn{8}{c|}{\cellcolor[HTML]{FFCCC9}Test-FR5691} & \multicolumn{8}{c}{\cellcolor[HTML]{FFCCC9}Test-FS1941} \\ \cline{3-18} 
    \rowcolor[HTML]{FFCCC9} 
    \cellcolor[HTML]{FFCCC9} &
      \cellcolor[HTML]{FFCCC9} &
      \multicolumn{5}{c|}{\cellcolor[HTML]{FFCCC9}Depth Error (↓)} &
      \multicolumn{3}{c|}{\cellcolor[HTML]{FFCCC9}Depth Accuracy(↑)} &
      \multicolumn{5}{c|}{\cellcolor[HTML]{FFCCC9}Depth Error (↓)} &
      \multicolumn{3}{c}{\cellcolor[HTML]{FFCCC9}Depth Accuracy(↑)} \\ \cline{3-18} 
    \rowcolor[HTML]{FFCCC9} 
    \multirow{-3}{*}{\cellcolor[HTML]{FFCCC9}Methods} &
    \multirow{-3}{*}{\cellcolor[HTML]{FFCCC9}Publication} &
    $\mathrm{RMSE}$ & $\mathrm{RMSE_{log}}$ & $\mathrm{A.Rel}$ & $\mathrm{S.Rel}$ & $\mathrm{log_{10}}$ & $\delta_1$ & $\delta_2$ & $\delta_3$ &
    $\mathrm{RMSE}$ & $\mathrm{RMSE_{log}}$ & $\mathrm{A.Rel}$ & $\mathrm{S.Rel}$ &
    \multicolumn{1}{c|}{\cellcolor[HTML]{FFCCC9}$\mathrm{log_{10}}$} & $\delta_1$ & $\delta_2$ & $\delta_3$ \\ 
    \hline \hline
    IBLA\cite{IBLA} & TIP   2017 &  0.35 &  1.89 &  2.00 &   0.60 &  0.71 &  0.11 &  0.21 &  0.29 & 0.28 & 1.36 & 4.26 & 1.81 & 0.43 & 0.24 & 0.41 & 0.52 \\ 
    GDCP\cite{GDCP} & TIP   2018 &  0.31 &  1.54 &  6.88 &   3.16 &  0.50 &  0.17 &  0.32 &  0.45& 0.31 & 1.49 & 5.51 & 2.41 & 0.47 & 0.19 & 0.36 & 0.48   \\
    UW-Net\cite{UW-Net} & ICIP  2019 &  0.48 &  1.76 &  13.01 &  8.55 &  0.56 &  0.17 &  0.30 &  0.42 & 0.46 & 1.70 & 10.11 & 6.54 & 0.56 & 0.15 & 0.29 & 0.40  \\
    NUDCP\cite{NUDCP} & TOB   2020 &  0.34 &  1.58 &  8.19 &   4.27 &  0.48 &  0.21 &  0.37 &  0.50 & 0.39 & 1.61 & 9.40 & 5.76 & 0.51 & 0.20 & 0.35 & 0.46   \\
    UW-GAN\cite{UW-GAN} & TIM   2021 &  0.19 &  0.92 &  1.61 &   0.34 &  0.28 &  0.35 &  0.55 &  0.68 & 0.18 & 0.89 & 2.18 & 0.92 & 0.28 & 0.44 & 0.59 & 0.68   \\
    HazeLine\cite{HazeLine} & TPAMI 2021 &  0.33 &  1.45 &  7.16 &   3.52 &  0.47 &  0.17 &  0.32 &  0.46 & 0.29 & 1.39 & 5.96 & 2.51 & 0.43 & 0.22 & 0.39 & 0.52   \\
    MiDas\cite{MiDas} & TPAMI 2022 &  0.24 &  1.21 &  4.77 &   1.52 &  0.37 &  0.25 &  0.45 &  0.60 & 0.20 & 0.98 & 2.56 & 0.84 & 0.30 & 0.35 & 0.54 & 0.66   \\
    Lite-Mono\cite{Lite-Mono}& CVPR  2023 & 0.31 &  1.45 &  7.93 &   4.35 &  0.44 &  0.22 &  0.39 &  0.53 & 0.23 & 1.27 & 4.77 & 2.01 & 0.38 & 0.27 & 0.45 & 0.57   \\
    UDepth\cite{UDepth}     & ICRA  2023 &  0.20 &  0.97 &  1.90 &   0.47 &  0.30 &  0.32 &  0.53 &  0.67 & 0.16 & 0.92 & 1.87 & 0.55 & 0.28 & 0.46 & 0.61 & 0.69   \\
    ADPCC\cite{ADPCC} & IJCV  2023&   0.32 &  1.44 &  5.32 &   2.67 &  0.47 &  0.19 &  0.35 &  0.47 & 0.33 & 1.47 & 4.99 & 2.56 & 0.48 & 0.15 & 0.30 & 0.44   \\
    UW-Depth\cite{UW-Depth} & ICRA  2024 &  0.19 &  0.92 &  1.90 &   0.42 &  0.28 &  0.32 &  0.54 &  0.67 & 0.16 & 0.86 & 1.69 & 0.56 & 0.26 & 0.47 & 0.63 & 0.70   \\
    WsUID-Net\cite{SUIM-SDA}& TGRS  2024 &  0.39 &  1.60 &  12.01 &  7.07 &  0.51 &  0.19 &  0.33 &  0.47 & 0.41 & 1.66 & 9.37 & 5.33 & 0.53 & 0.20 & 0.33 & 0.43   \\
    WaterMono\cite{WaterMono}& TIM   2025 & 0.26 &  1.26 &  4.36 &   1.21 &  0.40 &  0.23 &  0.41 &  0.55 & 0.24 & 1.25 & 3.01 & 1.08 & 0.39 & 0.27 & 0.43 & 0.55  \\
    \hline
    Tree-Mamba &     -      
    &  \textcolor{red}{0.05}
    &  \textcolor{red}{0.32} 
    &  \textcolor{red}{0.29} 
    &  \textcolor{red}{0.04} 
    &  \textcolor{red}{0.08} 
    &  \textcolor{red}{0.77} 
    &  \textcolor{red}{0.89} 
    &  \textcolor{red}{0.93} 
    &   \textcolor{red}{0.03} 
    &  \textcolor{red}{0.25} 
    &  \textcolor{red}{0.16} 
    &  \textcolor{red}{0.01} 
    &  \textcolor{red}{0.06} 
    &  \textcolor{red}{0.83} 
    &  \textcolor{red}{0.93} 
    &  \textcolor{red}{0.96}\\ 

    \hline
    
    \end{tabular}
}
\end{table*}

\vspace{-0.3cm}
\subsection{Qualitative Evaluation}

We conduct the visual comparisons of different methods on \textbf{Test-FR5691} and \textbf{Test-FS1941}.
Visual results of underwater images with different color degradations are shown in Figs. \ref{Qual_R}, \ref{Qual_S}, and \ref{Qual_O}.
As illustrated in Fig.~\ref{Qual_R} (b), (c), (e), (g), and (k), five traditional methods (IBLA~\cite{IBLA}, UDCP~\cite{UDCP}, NUDCP~\cite{NUDCP}, HazeLine~\cite{HazeLine}, and ADPCC~\cite{ADPCC}) rely on delicate prior assumptions and frequently misestimate depth change trends, especially in the zoomed-in areas—where they incorrectly infer wreckage or sculptures as distant objects.
In Figs. \ref{Qual_R}-\ref{Qual_O} (d), UW-Net \cite{UW-Net} performs well in bluish and turbid underwater images, but fails in greenish, low-light, and other color degradations underwater images, since it takes the haze as a depth cue to estimate depth maps. 
In Figs. \ref{Qual_R}-\ref{Qual_O} (f), UW-GAN \cite{UW-GAN} performs well on various color casts, but fails to capture depth geometry because of a simple CNN architecture for feature learning.
MiDas \cite{MiDas} achieves relatively satisfactory results for the bluish underwater image by using the pre-trained encoders with feature extraction capabilities, but fails in other types of degraded underwater images, as shown in Figs. \ref{Qual_R}-\ref{Qual_O} (h).
Lite-Mono \cite{Lite-Mono} estimates the local trend of depth changes but does not generate holistic content in Fig. \ref{Qual_S} (i).
As shown in Figs. \ref{Qual_O} (j) and (l), UDepth \cite{UDepth} and UW-Depth \cite{UW-Depth} achieve clear scene depth but lack fine structures and details, because their AdaBins module struggles to distinguish depth values with subtle differences. 
In Figs. \ref{Qual_R}-\ref {Qual_O} (m), WsUID-Net \cite {SUIM-SDA} yields global depth trends, but introduces blurred boundaries and details, due to multi-task feature fusion to bring substantial depth noise
for depth prediction.
As shown in Figs. \ref{Qual_R}-\ref {Qual_O} (n), WaterMono \cite{WaterMono} infers depth trend but lacks clear object content, since its teacher-guided anomaly masking module removes part content regions to facilitate the prediction of dynamic regions. 
By contrast, our Tree-Mamba method produces depth maps with richer details, clearer boundaries, and sharper geometry on various degraded underwater images, thanks to the effective adoption of the proposed tree-aware Mamba block.

Moreover, we demonstrate the scalability of our proposed Tree-Mamba method in four types of underwater scenes by estimating image depths from a challenging underwater video. 
The enhanced video by our method is available in the supplementary material\textsuperscript{\ref {sm}}, and  partial results are shown in Fig. \ref{Qual_V}. 
As shown, our Tree-Mamba method produces accurate depth maps across various underwater scenes, and the depth results for different frames are consistent and visually compelling.
In addition, we perform a visual comparison of different methods on an underwater panoramic image. This panoramic image provides a wider field of view but suffers from severe geometric distortions at the image edges, posing a significant challenge to the robustness of UMDE methods.
As illustrated in Fig. \ref{Qual_P}, the proposed Tree-Mamba method can estimate depth results of both fine details and clear structures, which suggests that our method is more robust in estimating better depth maps of underwater panoramic images with severe geometric distortions.
This superiority is attributed to our tree-aware Mamba block that uses the tree-aware scanning scheme to model multi-scale structure features.

\vspace{-0.4cm}
\subsection{Quantitative Assessment}

We quantify the performance of different methods in terms of five standard error metrics ($\mathrm{RMSE}$, $\mathrm{RMSE_{log}}$, $\mathrm{A.Rel}$, $\mathrm{S.Rel}$, $\mathrm{log_{10}}$) and three accuracy metrics ($\delta _1$, $\delta _2$, $\delta _3$) on the \textbf{Test-FR5691} and \textbf{Test-FS1941} respectively in Table \ref{tab quan}.
As shown, our Tree-Mamba method achieves eight highest scores, which reflects that our predicted depth maps are more similar to reference depth maps.
Compared with other competitors, the proposed Tree-Mamba yields lower $\mathrm{RMSE}$, which demonstrates that our results are more consistent with reference depth maps in terms of both object distances and spatial structures.
Moreover, our Tree-Mamba yields the lowest $\mathrm{RMSE_{log}}$, with the superior performance in areas of significant depth variations.
Our method produces the lowest scores of both $\mathrm{A.Rel}$ and $\mathrm{S.Rel}$, suggesting higher depth fidelity and better structural consistency of our results.
In the $\mathrm{log_{10}}$ metric, the results of our Tree-Mamba are ranked first, which indicates the lowest logarithmic error of our depth maps.
For the $\delta_1$, $\delta_2$, and $\delta_3$ metrics, our method achieves the highest scores and yields superior depth results in terms of finer details and sharper object geometry, which demonstrates the superiority of the proposed tree-aware scanning strategy on underwater depth estimation.

\begin{table}[t]
\centering

\caption{The average quality scores (↑) for depth estimation of different methods on \textbf{Video-NR1600}. The \textcolor{red}{best} results are marked in red.}
\label{tab_NR1600}
    \addvspace{-6pt}  
\resizebox{1\linewidth}{!}{
\begin{tabular}{c|c|c|c|c|c}
\hline
\rowcolor[HTML]{FFCCC9} 
Methods      &Publication& Bluish & Greenish & Turbid & \makecell{Non-uniform\\ light} \\ 
\hline \hline

IBLA\cite{IBLA}              &      TIP   2017 & 54.06 & 55.77 & 46.28 & 46.93      \\
GDCP\cite{GDCP}              &      TIP   2018 & 50.45 & 54.14 & 42.03 & 48.88      \\
UW-Net\cite{UW-Net}          &      ICIP  2019 & 67.99 & 72.41 & 60.71 & 71.61      \\ 
NUDCP\cite{NUDCP}            &      TOB   2020 & 51.50 & 53.90 & 44.95 & 43.43      \\ 
UW-GAN\cite{UW-GAN}          &      TIM   2021 & 85.58 & 89.66 & 86.84 & 85.90      \\ 
HazeLine\cite{HazeLine}      &      TPAMI 2021 & 53.51 & 53.99 & 48.67 & 48.57      \\ 
MiDas\cite{MiDas}            &      TPAMI 2022 & 80.00 & 86.49 & 86.16 & 79.57      \\ 
Lite-Mono\cite{Lite-Mono}    &      CVPR  2023 & 63.78 & 64.24 & 66.20 & 75.20      \\
UDepth\cite{UDepth}          &      ICRA  2023 & 84.09 & 88.95 & 90.76 & 89.70      \\ 
ADPCC\cite{ADPCC}            &      IJCV  2023 & 57.85 & 58.09 & 58.40 & 62.40      \\
UW-Depth\cite{UW-Depth}      &      ICRA  2024 & 88.46 & 91.71 & 92.24 & 95.53      \\ 
WsUID-Net\cite{SUIM-SDA}     &      TGRS  2024 & 61.83 & 68.69 & 64.85 & 71.82      \\ 
WaterMono\cite{WaterMono}    &      TIM   2025 & 68.45 & 74.98 & 71.70 & 69.59      \\

\hline
Tree-Mamba                   &      -           
& \textcolor{red}{95.97} 
& \textcolor{red}{96.37}   
& \textcolor{red}{97.57} 
& \textcolor{red}{98.43}      \\   
\hline
\end{tabular}
}
\end{table}

\begin{figure}[t]
    \centering
    \includegraphics[width=0.85\linewidth]{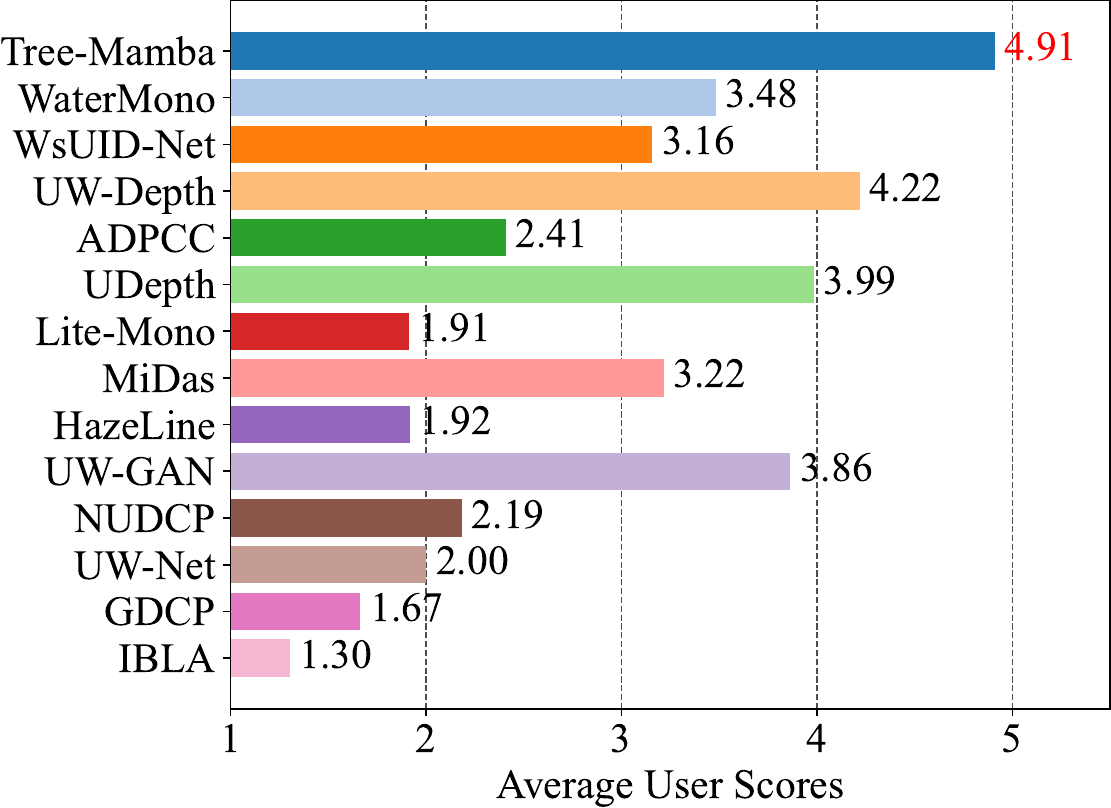}

    \addvspace{-6pt}  

    \caption{User study for different methods on depth results in Figs. \ref{Qual_R}-\ref{Qual_P}. Compared with other methods, our Tree-Mamba achieves the highest score, which suggests that our method produces better depth results in a subjective study.}
    \label{Qual_U}
\end{figure}

Moreover, we evaluate the average quality scores for depth estimation by different methods on the \textbf{Video-NR1600}.
In Table \ref{tab_NR1600}, the fine-tuned LAR-IQA model \cite{LAR-IQA} is used to assess the quality scores of different methods, and underwater image frames in the \textbf{Video-NR1600} are categorized into four types of color degradations: bluish, greenish, turbid, and non-uniform light.
As shown, the proposed Tree-Mamba gains the highest quality scores for depth estimation across four types of color degradations, which suggests the superiority of our method in yielding better depth results, and reflects the consistency of the Tree-Mamba across different frames at various underwater scenes.
Moreover, we measure the quality scores of different methods for estimating depth results of underwater panoramic images.
As shown in Fig. \ref{Qual_P}, the depth maps by our Tree-Mamba approach obtain the highest score, which suggests that our method can estimate better depth results with fine-grained details and overcome relatively serious geometric distortions of underwater panoramic images.
In addition, we conducted a user study to assess the visual depth results of different methods in Figs. \ref{Qual_R}-\ref{Qual_P}. 
We invite 100 volunteers to score the perceptual quality of underwater depth results estimated by different methods, in which the scoring relies on the consistency of content and depth between underwater images and depth results. The scores are rated on a five-point scale: 5 (Excellent), 4 (Good), 3 (Fair), 2 (Poor), and 1 (Very Poor).
Fig. \ref{Qual_U} shows the average scores of the results by different methods, as compared, our method gains the highest average score, which indicates that our method produces better depth results in a subjective study.

\begin{figure*}[!htp]
    \Large
    \centering
    \resizebox{0.85\linewidth}{!}{
        \begin{tabular}{c@{ }c@{ }c@{ }c@{ }c@{ }c@{ }c@{ }c@{ }}
            \includegraphics[height=3cm,width=4cm]{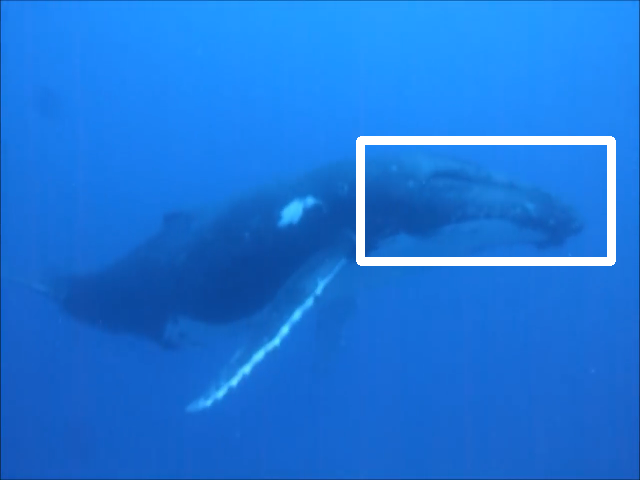} &
            \includegraphics[height=3cm,width=4cm]{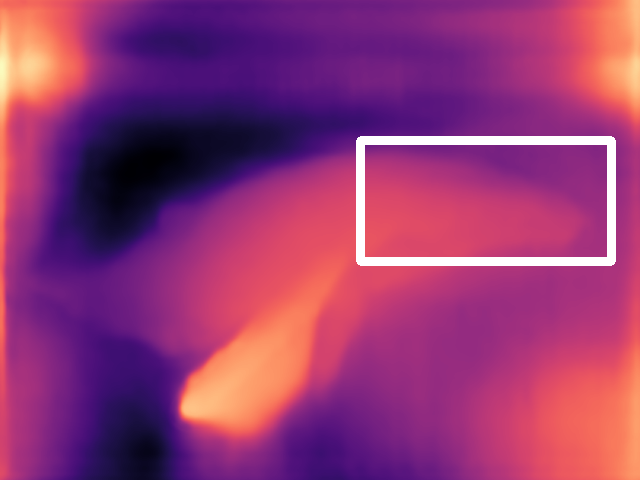} &
            \includegraphics[height=3cm,width=4cm]{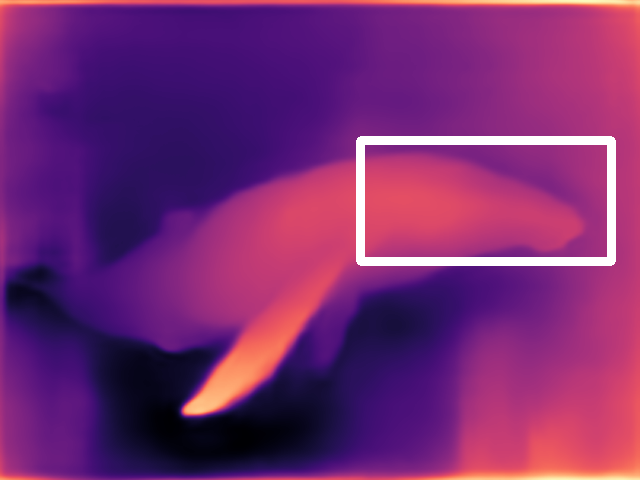} &
            \includegraphics[height=3cm,width=4cm]{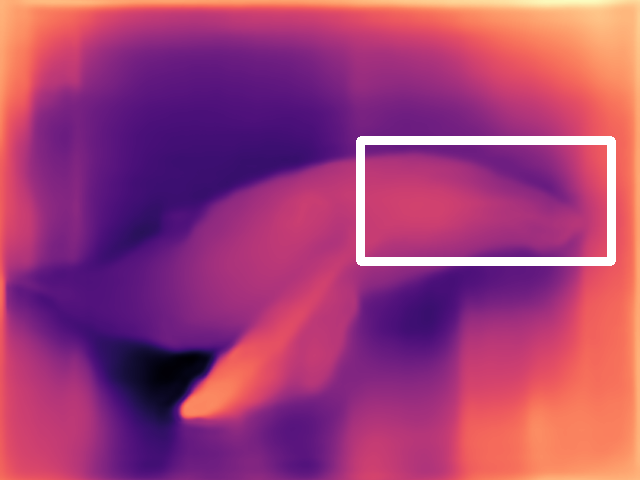} &
            \includegraphics[height=3cm,width=4cm]{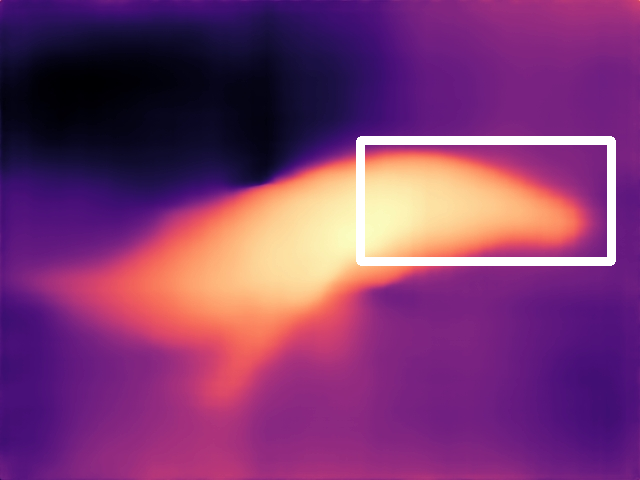} &
            \includegraphics[height=3cm,width=4cm]{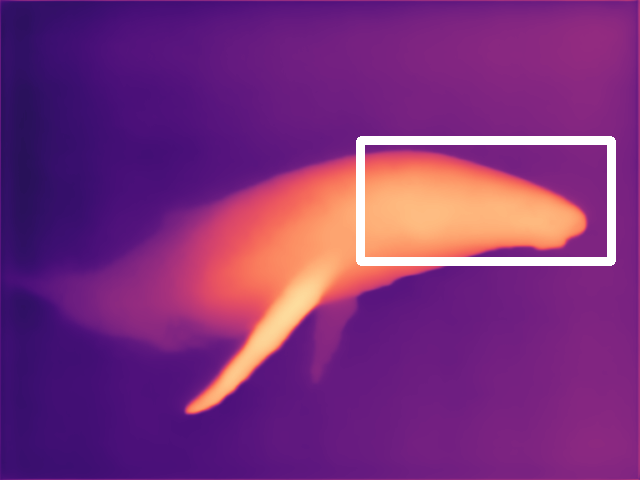} &
            \includegraphics[height=3cm,width=4cm]{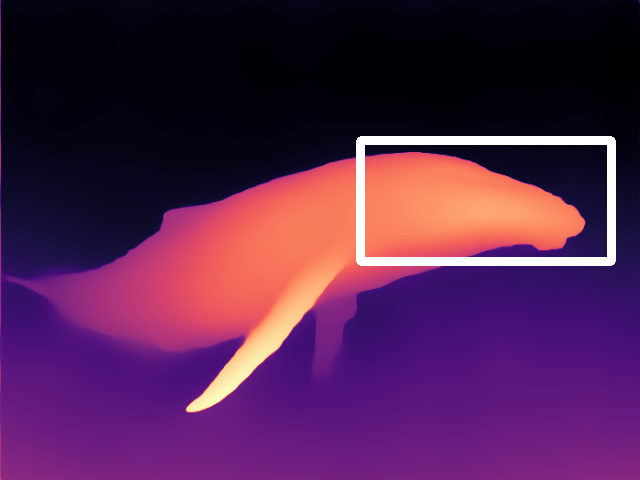} &
            \includegraphics[height=3cm,width=4cm]{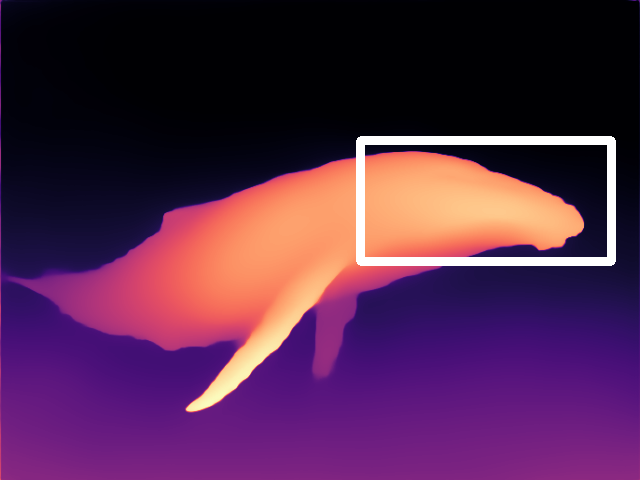} \\
            
            \includegraphics[height=2cm,width=4cm]{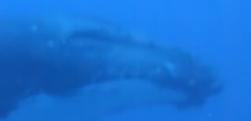} &
            \includegraphics[height=2cm,width=4cm]{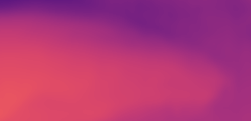} &
            \includegraphics[height=2cm,width=4cm]{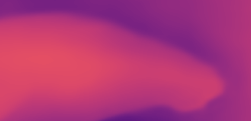} &
            \includegraphics[height=2cm,width=4cm]{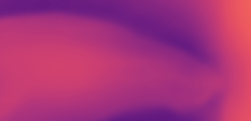} &
            \includegraphics[height=2cm,width=4cm]{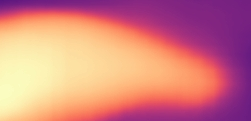} &
            \includegraphics[height=2cm,width=4cm]{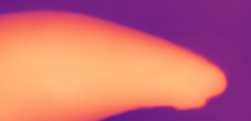} &
            \includegraphics[height=2cm,width=4cm]{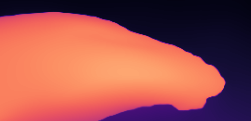} &
            \includegraphics[height=2cm,width=4cm]{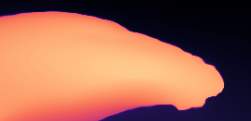} \\

            \includegraphics[height=3cm,width=4cm]{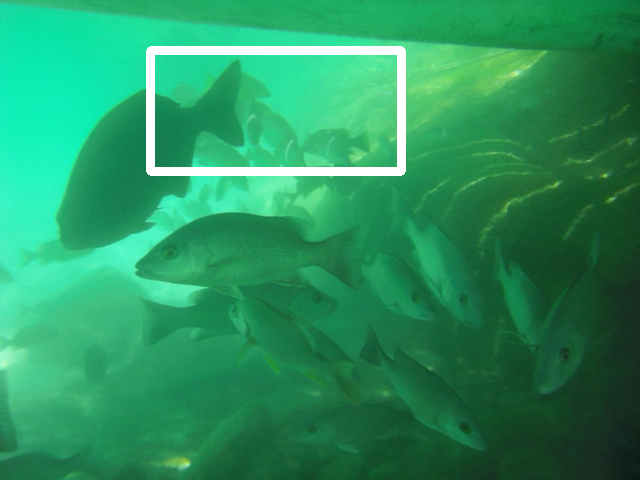} &
            \includegraphics[height=3cm,width=4cm]{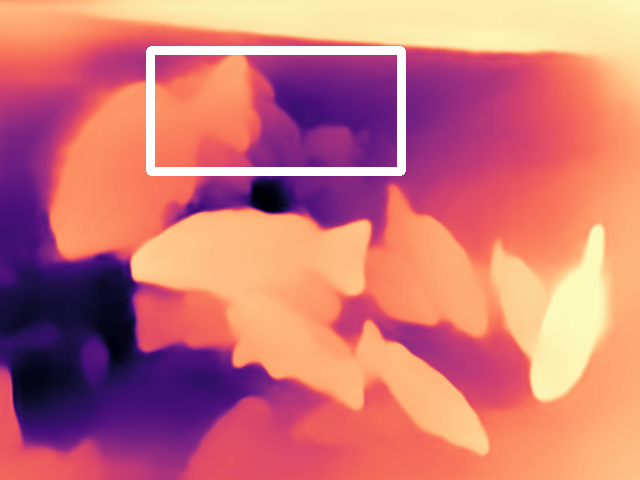} &
            \includegraphics[height=3cm,width=4cm]{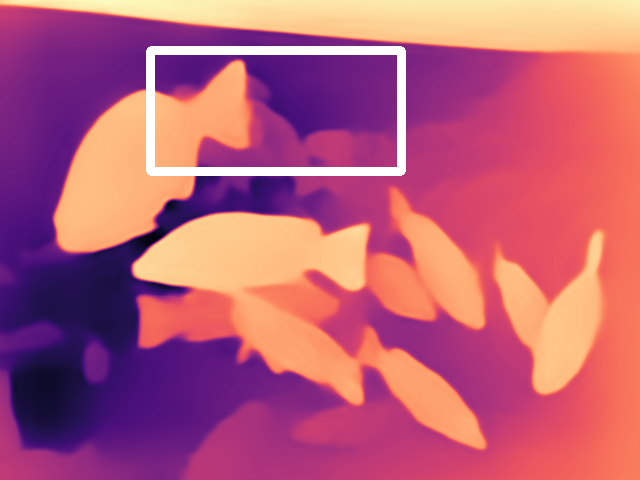} &
            \includegraphics[height=3cm,width=4cm]{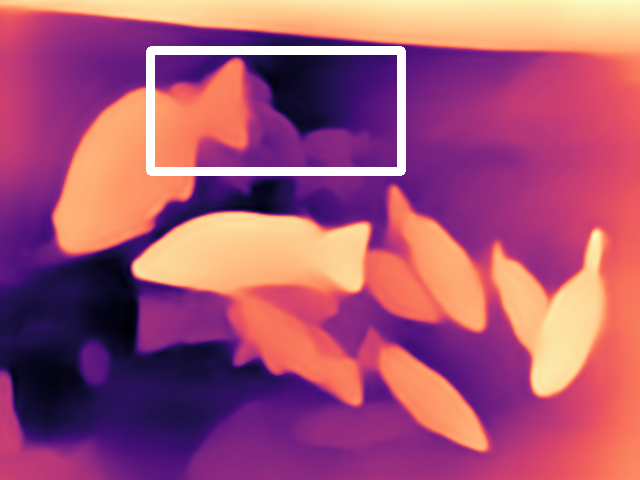} &
            \includegraphics[height=3cm,width=4cm]{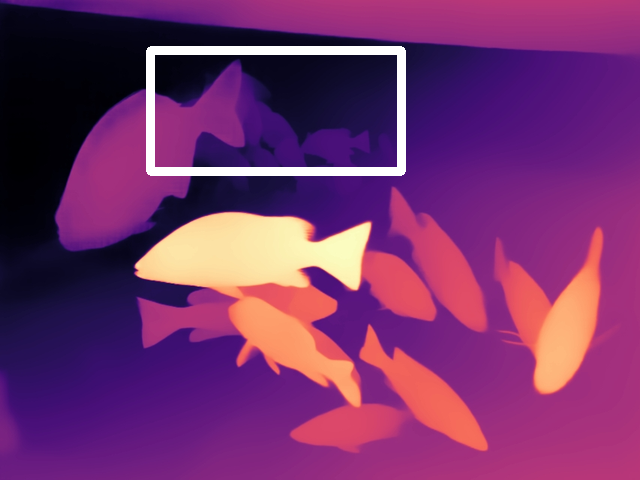} &
            \includegraphics[height=3cm,width=4cm]{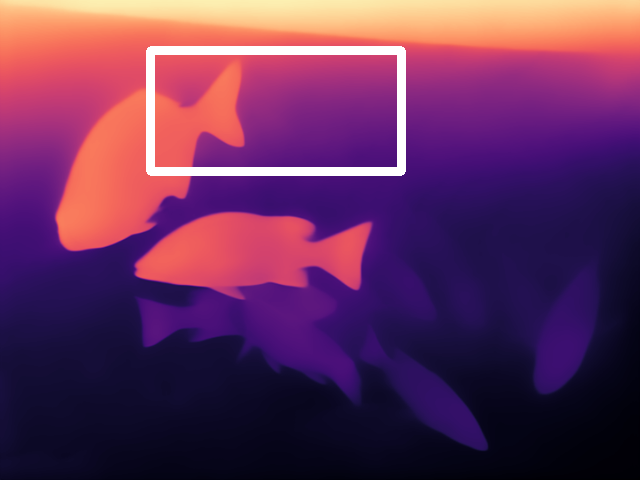} &
            \includegraphics[height=3cm,width=4cm]{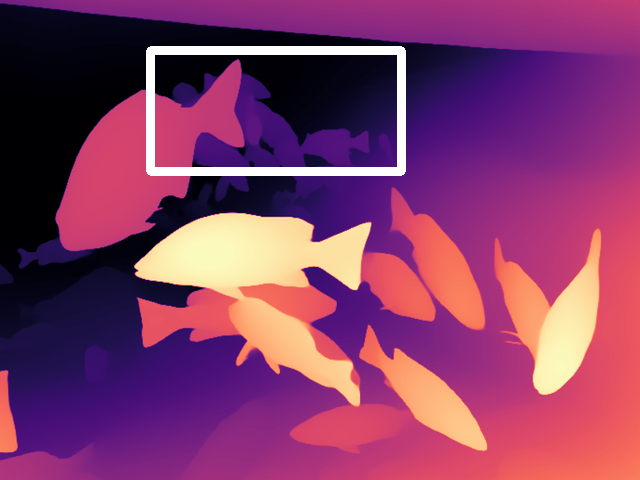} &
            \includegraphics[height=3cm,width=4cm]{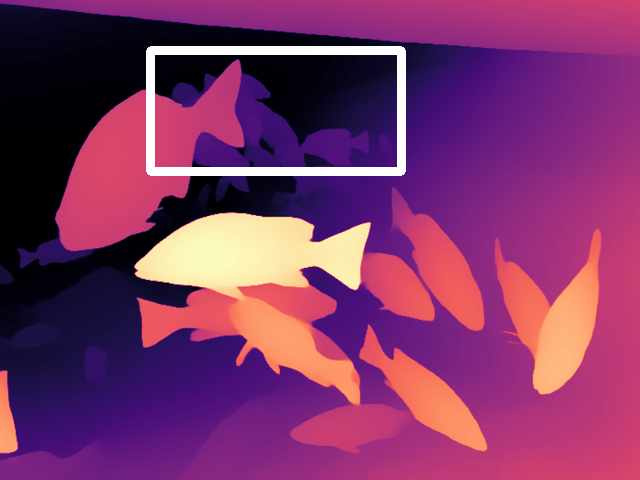} \\
            
            \includegraphics[height=2cm,width=4cm]{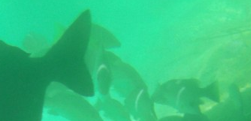} &
            \includegraphics[height=2cm,width=4cm]{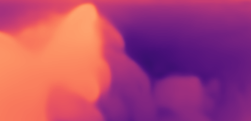} &
            \includegraphics[height=2cm,width=4cm]{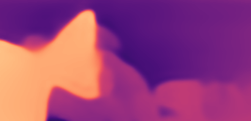} &
            \includegraphics[height=2cm,width=4cm]{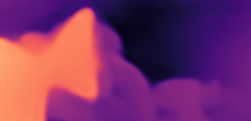} &
            \includegraphics[height=2cm,width=4cm]{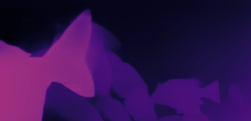} &
            \includegraphics[height=2cm,width=4cm]{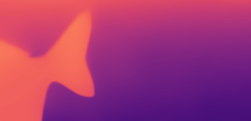} &
            \includegraphics[height=2cm,width=4cm]{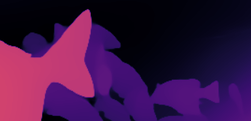} &
            \includegraphics[height=2cm,width=4cm]{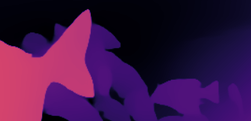} \\

            (a) Input & 
            (b) \textit{w/ RS} &
            (c) \textit{w/ CS} & 
            (d) \textit{w/ DS} & 
            (e) \textit{w/ NSS} & 
            (f) \textit{w/ Manhattan} & 
            (g) \textit{full model} & 
            (h) Reference \\
             
        \end{tabular}
    }
    \caption{Visualization results of ablation study on the tree-aware Mamba block. In the zoomed-in areas, our \textit{full model} yields depth maps of both richer details and sharper edges.}
    \label{Ablation_SSM}
\end{figure*}

\begin{figure*}[!htp]
    \Large
    \centering
    \resizebox{0.85\linewidth}{!}{
        \begin{tabular}{c@{ }c@{ }c@{ }c@{ }c@{ }c@{ }c@{ }c@{ }}
            \includegraphics[height=3cm,width=4cm]{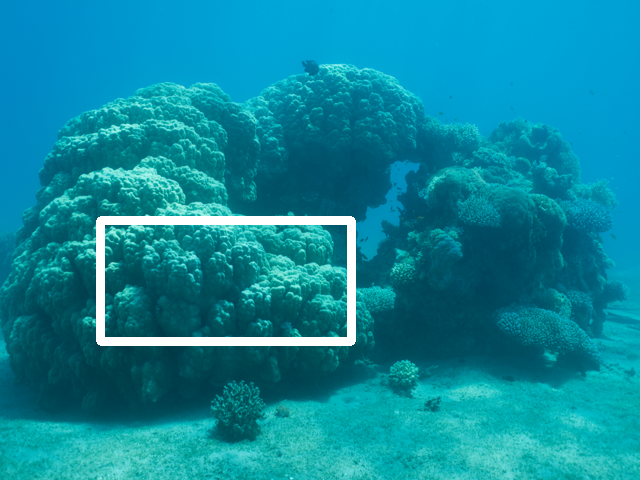} &
            \includegraphics[height=3cm,width=4cm]{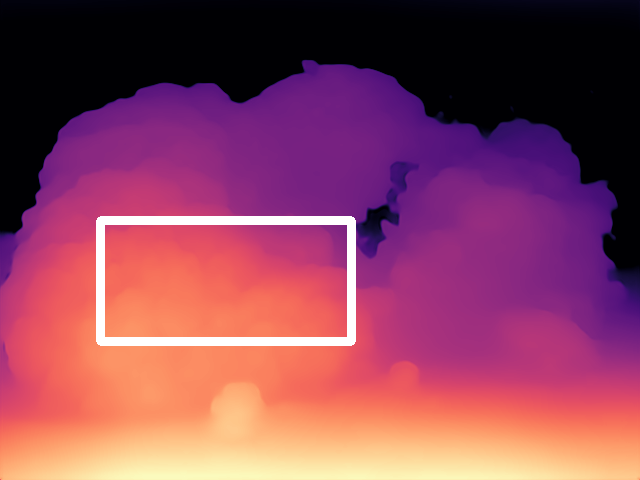} &
            \includegraphics[height=3cm,width=4cm]{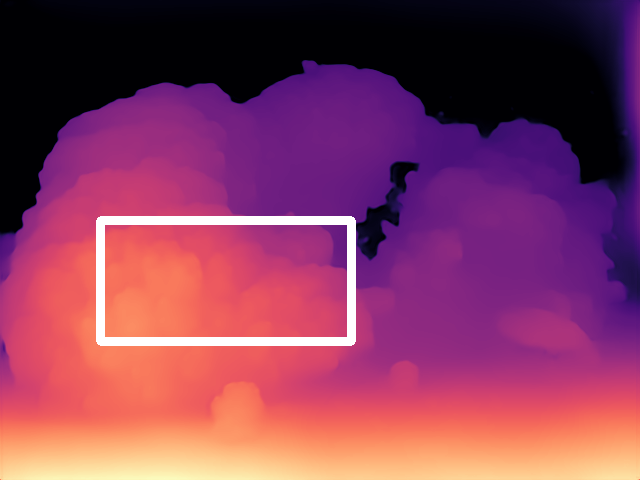} &
            \includegraphics[height=3cm,width=4cm]{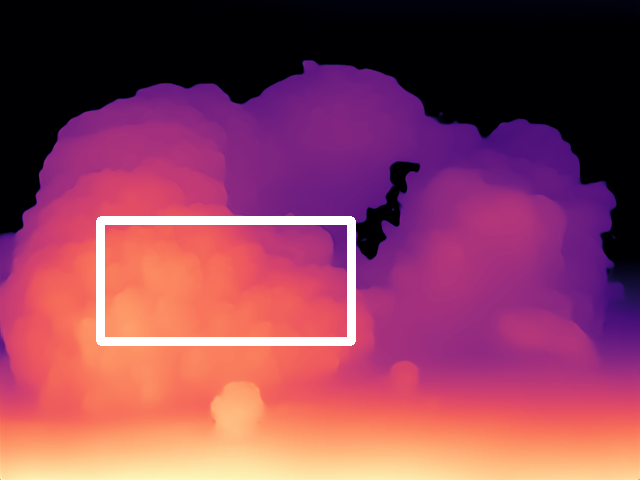} &
            \includegraphics[height=3cm,width=4cm]{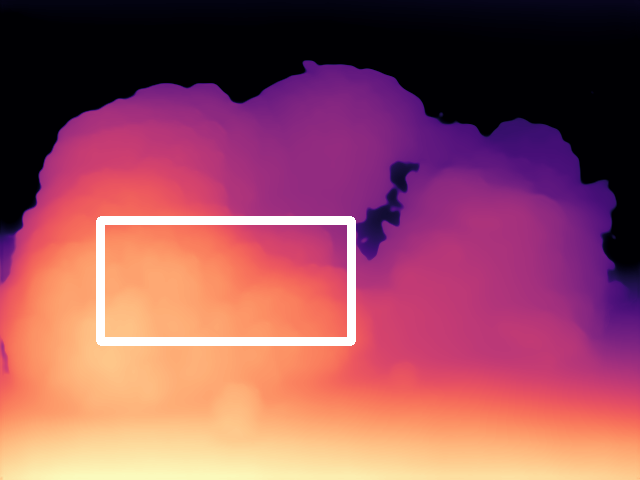} &
            \includegraphics[height=3cm,width=4cm]{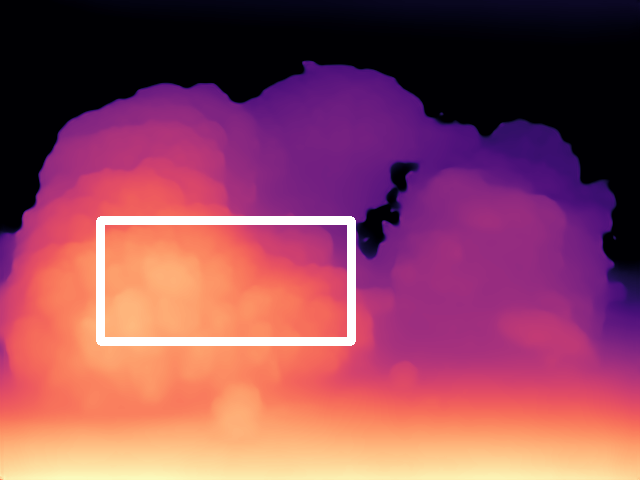} &
            \includegraphics[height=3cm,width=4cm]{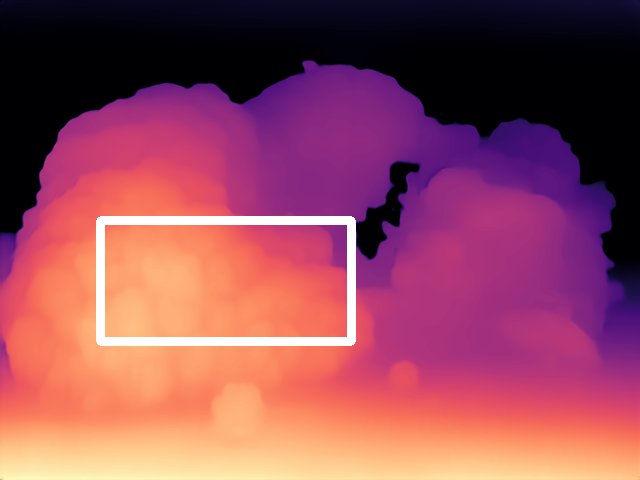} &
            \includegraphics[height=3cm,width=4cm]{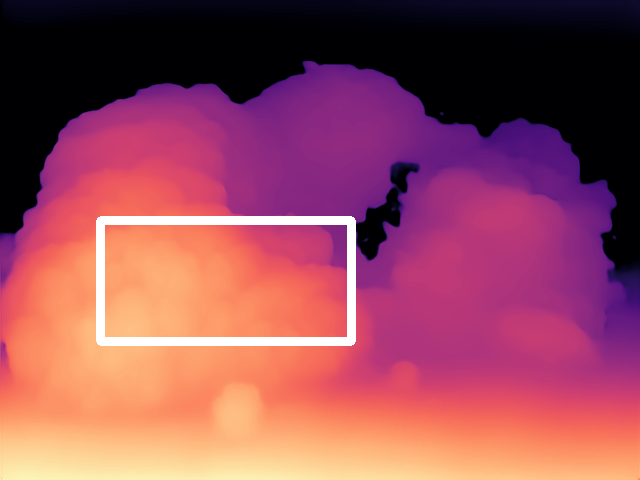} \\
            
            \includegraphics[height=2cm,width=4cm]{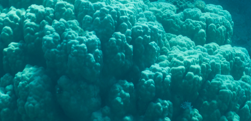} &
            \includegraphics[height=2cm,width=4cm]{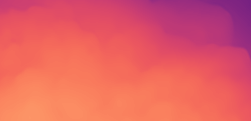} &
            \includegraphics[height=2cm,width=4cm]{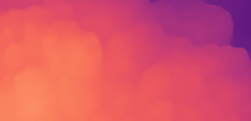} &
            \includegraphics[height=2cm,width=4cm]{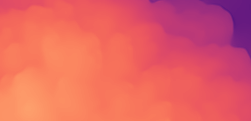} &
            \includegraphics[height=2cm,width=4cm]{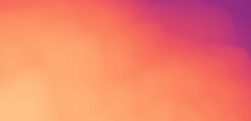} &
            \includegraphics[height=2cm,width=4cm]{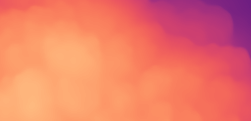} &
            \includegraphics[height=2cm,width=4cm]{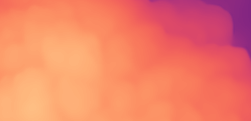} &
            \includegraphics[height=2cm,width=4cm]{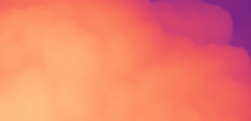} \\

            \includegraphics[height=3cm,width=4cm]{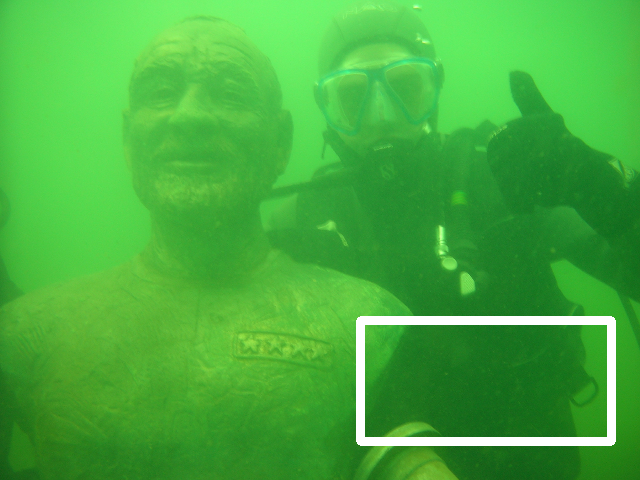} &
            \includegraphics[height=3cm,width=4cm]{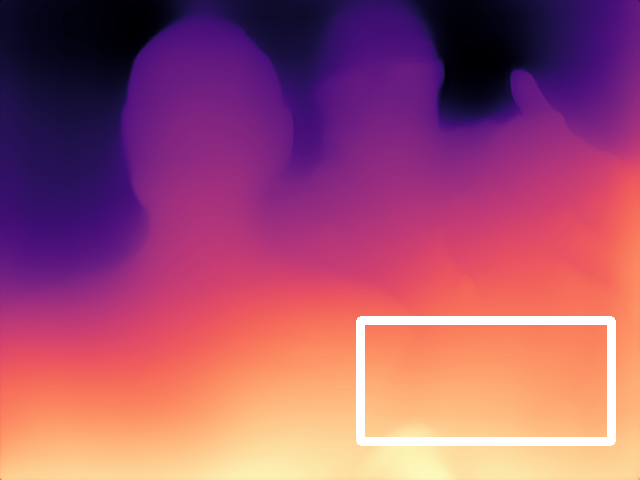} &
            \includegraphics[height=3cm,width=4cm]{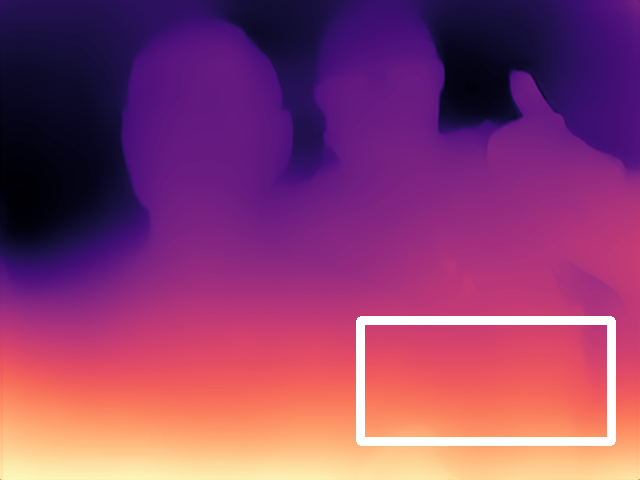} &
            \includegraphics[height=3cm,width=4cm]{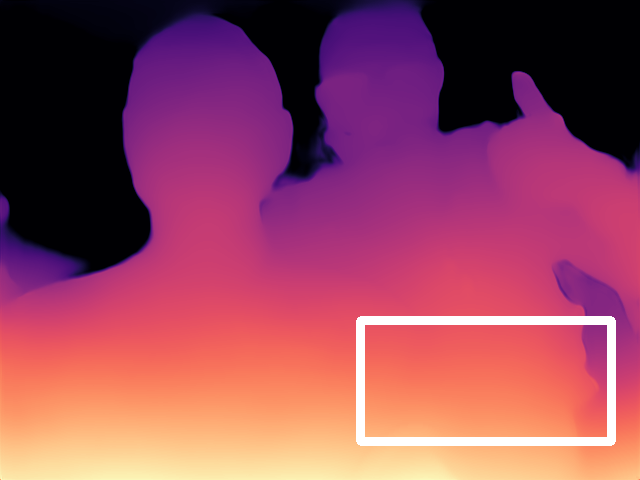} &
            \includegraphics[height=3cm,width=4cm]{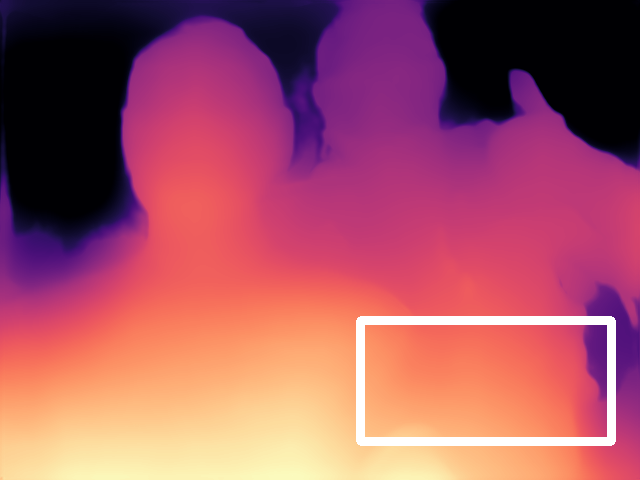} &
            \includegraphics[height=3cm,width=4cm]{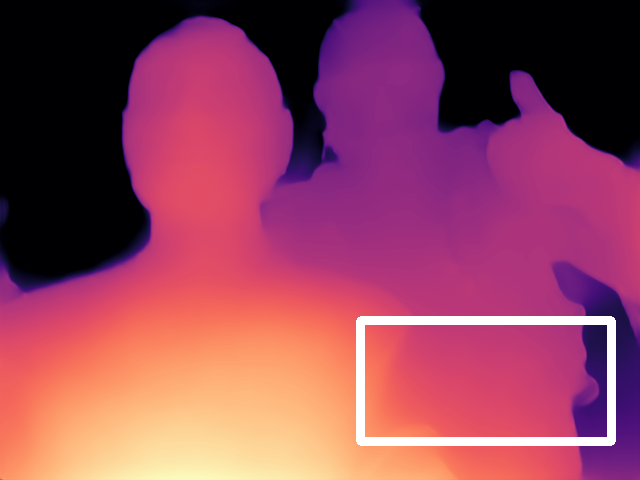} &
            \includegraphics[height=3cm,width=4cm]{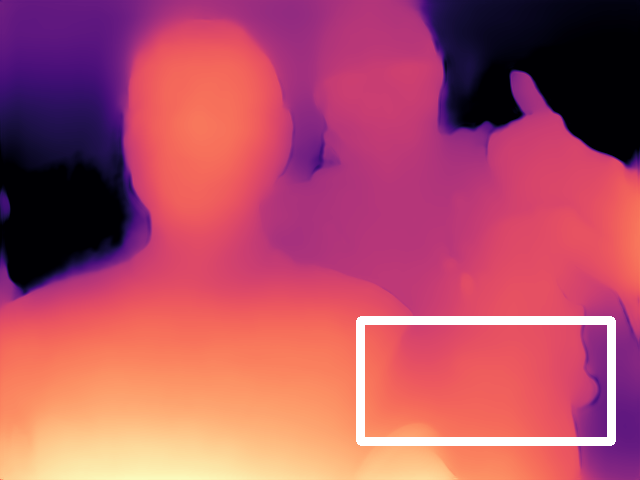} &
            \includegraphics[height=3cm,width=4cm]{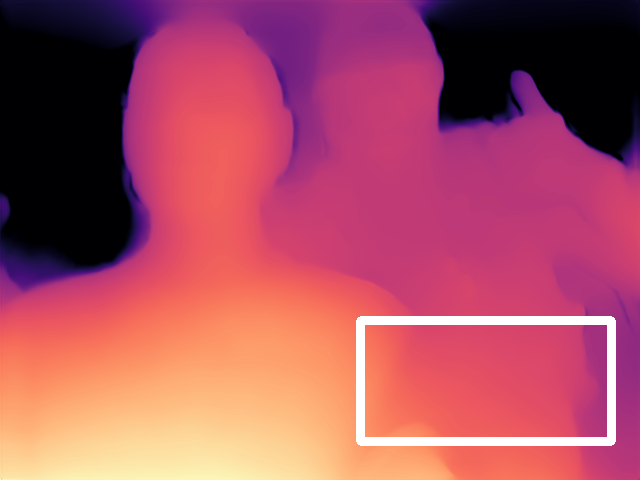} \\
            
            \includegraphics[height=2cm,width=4cm]{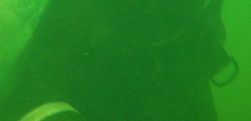} &
            \includegraphics[height=2cm,width=4cm]{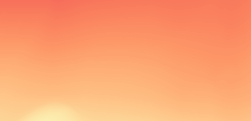} &
            \includegraphics[height=2cm,width=4cm]{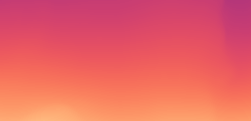} &
            \includegraphics[height=2cm,width=4cm]{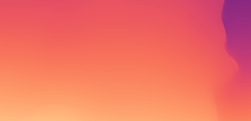} &
            \includegraphics[height=2cm,width=4cm]{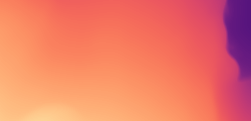} &
            \includegraphics[height=2cm,width=4cm]{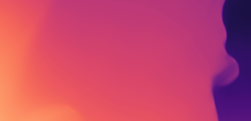} &
            \includegraphics[height=2cm,width=4cm]{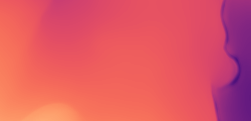} &
            \includegraphics[height=2cm,width=4cm]{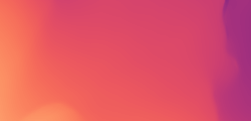} \\

            (a) Input & 
            (b) ResNet18 &
            (c) ResNet50 & 
            (d) ResNet101 & 
            (e) EfficientNet-B0 & 
            (f) EfficientNet-B2 & 
            (g) EfficientNet-B6 & 
            (h) EfficientNet-B7 \\

            \includegraphics[height=3cm,width=4cm]{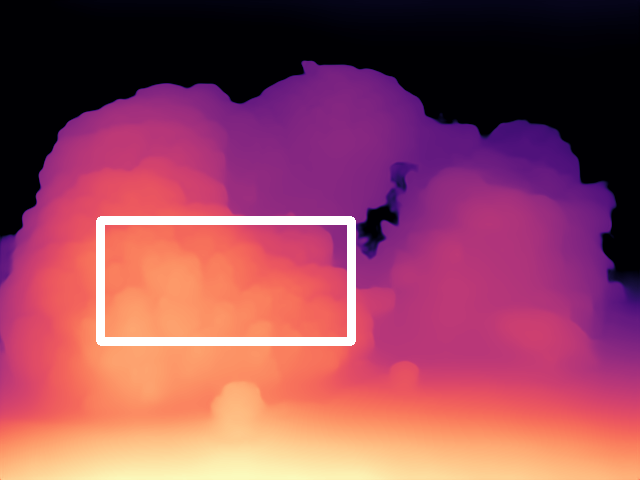} &
            \includegraphics[height=3cm,width=4cm]{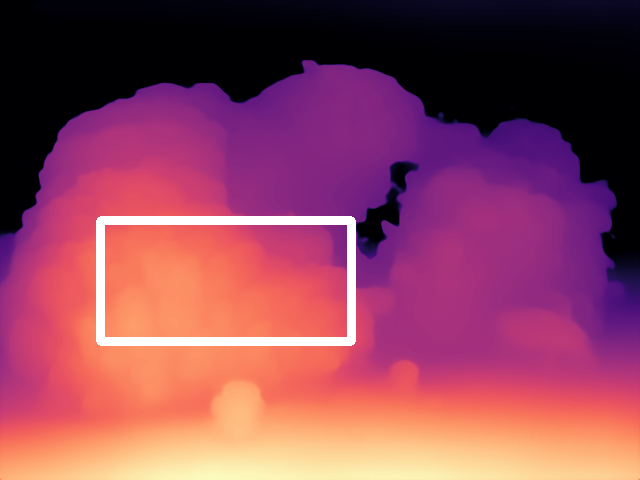} &
            \includegraphics[height=3cm,width=4cm]{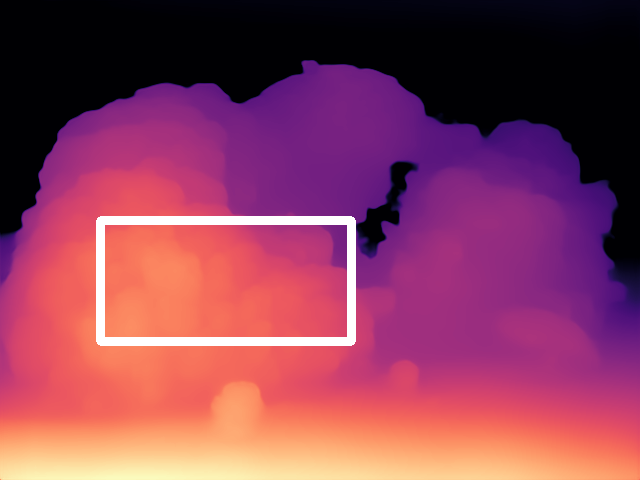} &
            \includegraphics[height=3cm,width=4cm]{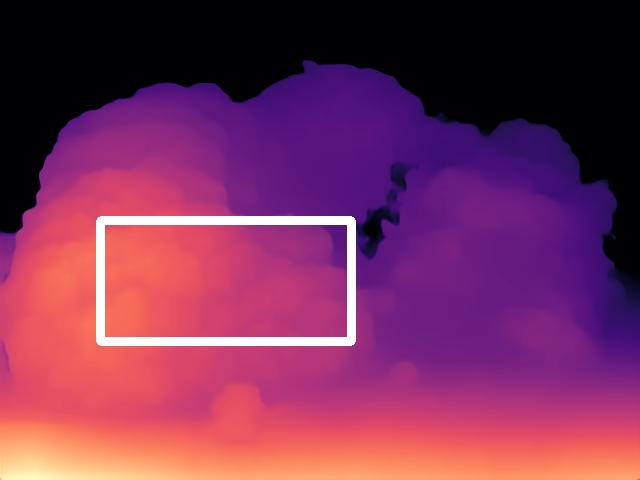} &
            \includegraphics[height=3cm,width=4cm]{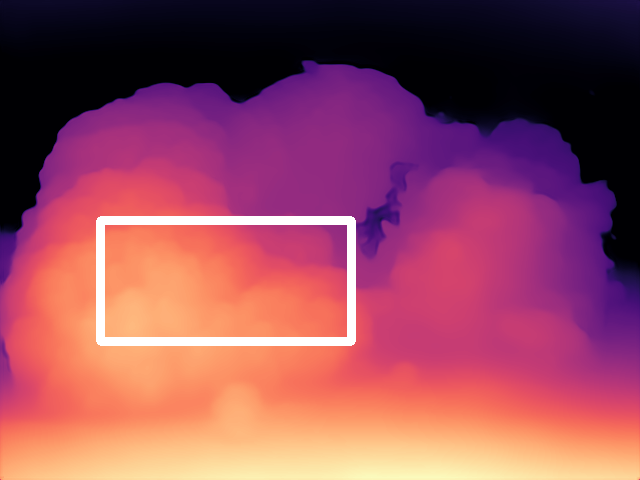} &
            \includegraphics[height=3cm,width=4cm]{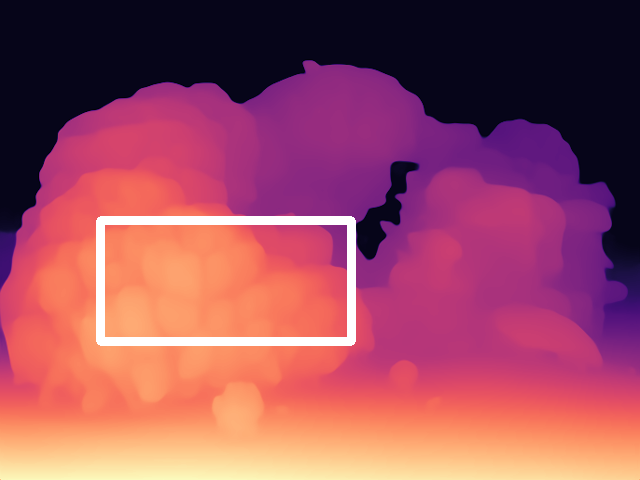} &
            \includegraphics[height=3cm,width=4cm]{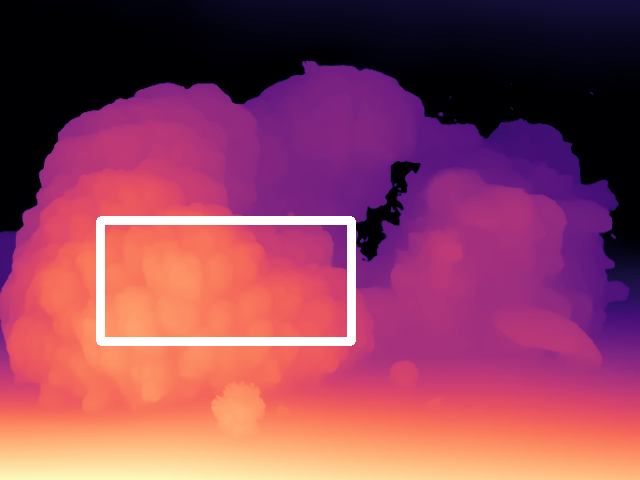} &
            \includegraphics[height=3cm,width=4cm]{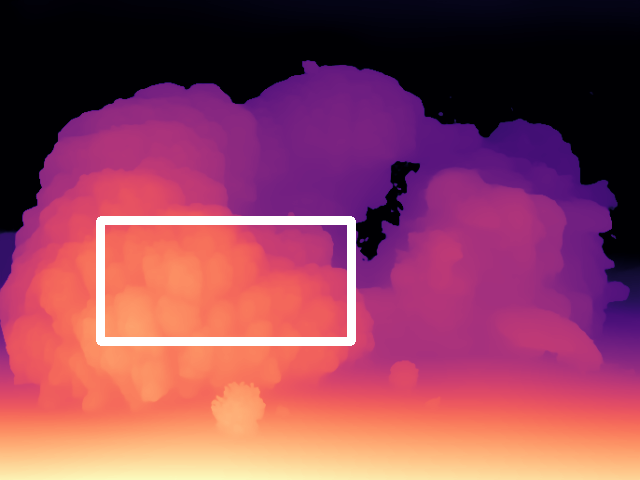} \\
            
            \includegraphics[height=2cm,width=4cm]{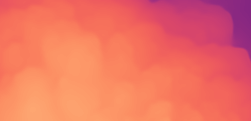} &
            \includegraphics[height=2cm,width=4cm]{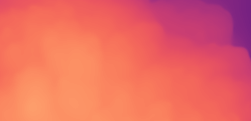} &
            \includegraphics[height=2cm,width=4cm]{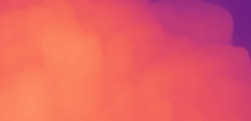} &
            \includegraphics[height=2cm,width=4cm]{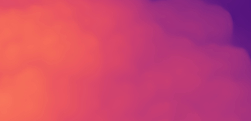} &
            \includegraphics[height=2cm,width=4cm]{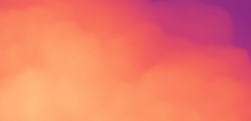} &
            \includegraphics[height=2cm,width=4cm]{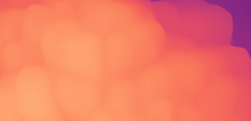} &
            \includegraphics[height=2cm,width=4cm]{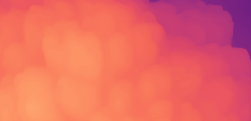} &
            \includegraphics[height=2cm,width=4cm]{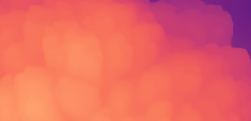} \\

            \includegraphics[height=3cm,width=4cm]{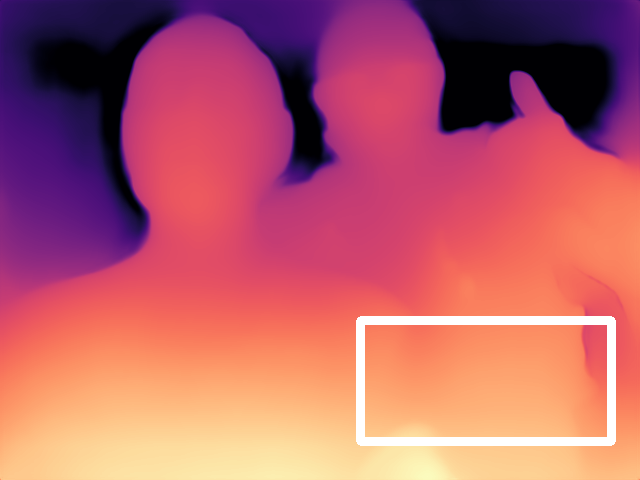} &
            \includegraphics[height=3cm,width=4cm]{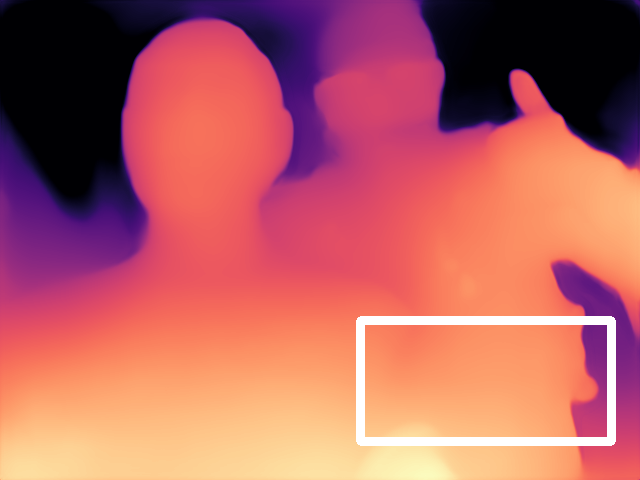} &
            \includegraphics[height=3cm,width=4cm]{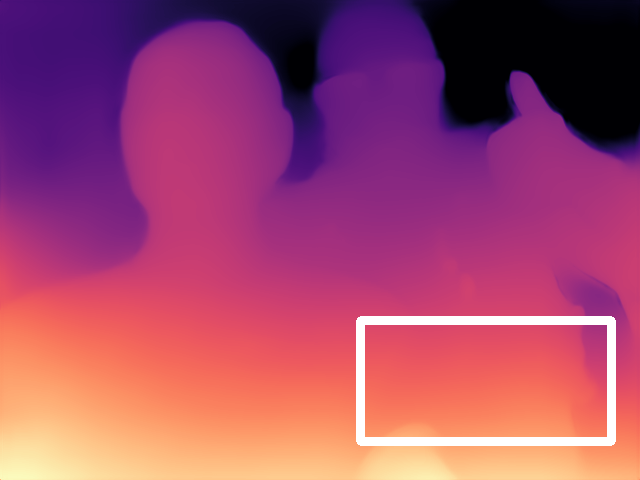} &
            \includegraphics[height=3cm,width=4cm]{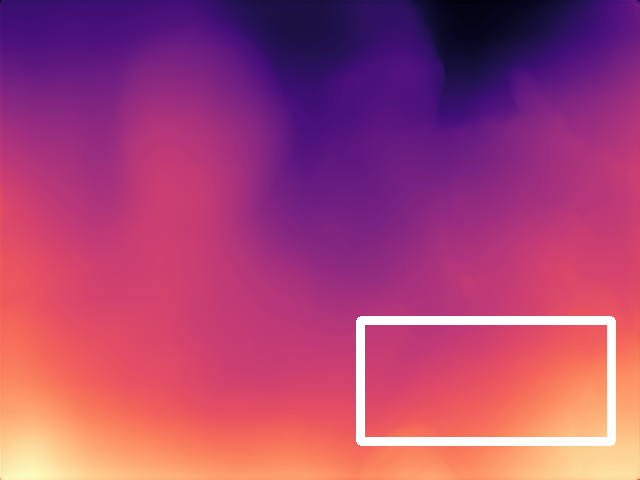} &
            \includegraphics[height=3cm,width=4cm]{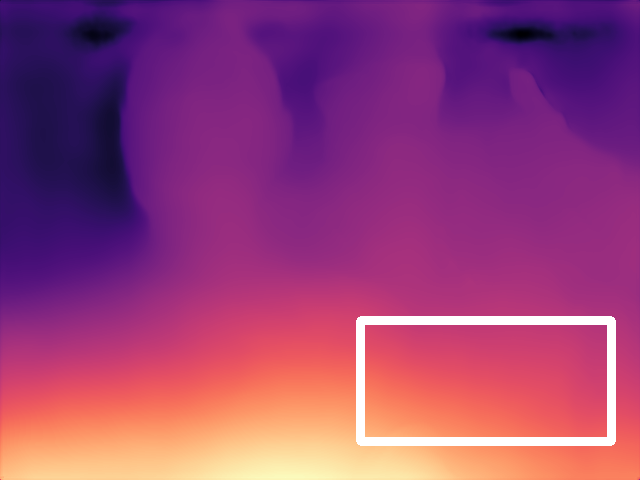} &
            \includegraphics[height=3cm,width=4cm]{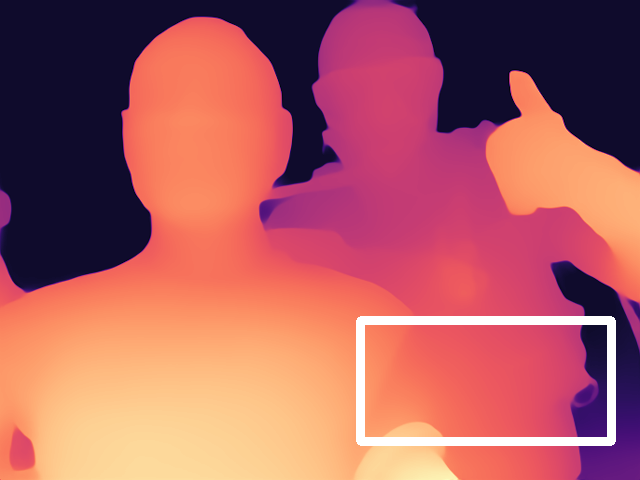} &
            \includegraphics[height=3cm,width=4cm]{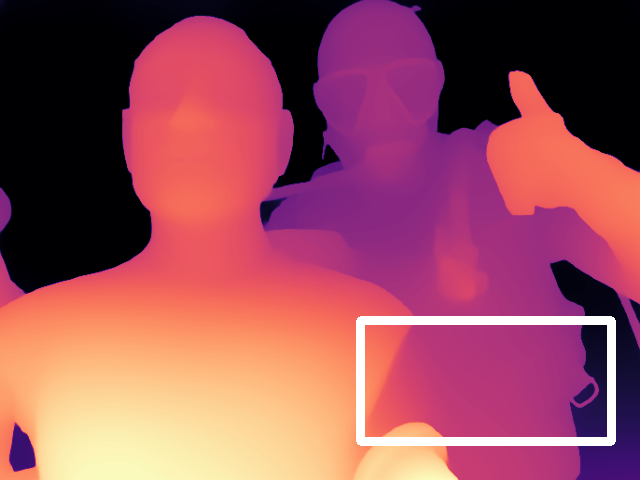} &
            \includegraphics[height=3cm,width=4cm]{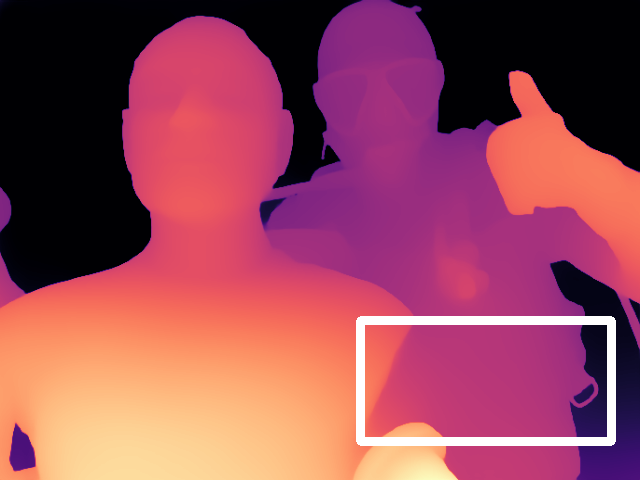} \\
            
            \includegraphics[height=2cm,width=4cm]{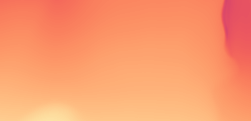} &
            \includegraphics[height=2cm,width=4cm]{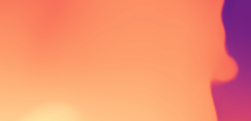} &
            \includegraphics[height=2cm,width=4cm]{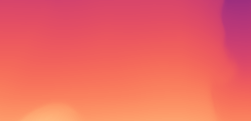} &
            \includegraphics[height=2cm,width=4cm]{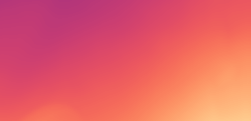} &
            \includegraphics[height=2cm,width=4cm]{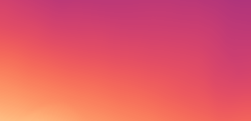} &
            \includegraphics[height=2cm,width=4cm]{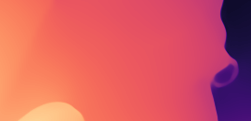} &
            \includegraphics[height=2cm,width=4cm]{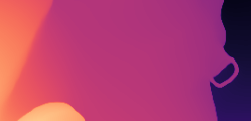} &
            \includegraphics[height=2cm,width=4cm]{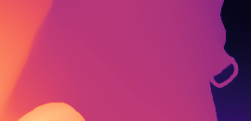} \\

            (i) DenseNet121 & 
            (j) DenseNet169 &
            (k) DenseNet201 & 
            (l) MobileNet-V2 & 
            (m) MobileNet-V3 & 
            (n) ResNeXt50 & 
            (o) \textcolor{red}{ResNeXt101} & 
            (p) Reference \\
             
        \end{tabular}
    }
    \caption{Visualization results of ablation study on different CNN-based encoders. When the encoder in our Tree-Mamba is ResNeXt101, predicted depth maps are closer to the reference images.}
    \label{Ablation_Encoders}
    
\end{figure*}

\begin{table}[t]
\centering
\caption{The average $\mathrm{A. Rel}$ and $\delta_1$ scores of ablation study of the tree-aware Mamba block and different CNN-based encoders on \textbf{Test-FR5691}. The complexity comparison of ablation study on an underwater image sized of 256 $\times$ 256. The \textcolor{red}{best} results are marked in red.}

\label{tab_Ablation}
    \addvspace{-6pt}  
\resizebox{0.9\linewidth}{!}{

\begin{tabular}{c|c|c|c|c}
\hline
\rowcolor[HTML]{FFCCC9} 
Methods & $\mathrm{A. Rel}$ ↓& $\delta_1$↑&FLOPs (G) ↓& Params (M) ↓\\ \hline \hline

\textit{w/ RS}&             8.82& 	0.19& 	\textcolor{red}{25.19}& 	\textcolor{red}{90.08}\\
\textit{w/ CS}&             4.71& 	0.35& 	25.33& 	90.27\\
\textit{w/ DS}&             7.42& 	0.30& 	25.27& 	90.23\\
\textit{w/ NSS}&            3.53& 	0.42& 	25.23& 	90.21\\
\textit{w/ Manhattan}&      5.79& 	0.39& 	\textcolor{red}{25.19}& 	\textcolor{red}{90.08}\\

\hline

\textit{full model}&        \textcolor{red}{0.29}& 	\textcolor{red}{0.77}&    \textcolor{red}{25.19}&    \textcolor{red}{90.08}\\

\hline \hline

ResNet18&               6.72& 	0.22& 	5.12& 	12.86\\
ResNet50&               6.10& 	0.22& 	8.99& 	26.85\\
ResNet101&              5.77& 	0.24& 	13.86&  45.84 \\

EfficientNet$\_$B0&     5.87 & 	0.21 & 3.14 & 5.96 \\
EfficientNet$\_$B2&     5.08 & 	0.22 & 3.52 & 9.73 \\
EfficientNet$\_$B6&     5.54 & 	0.23 & 7.26 & 43.34 \\
EfficientNet$\_$B7&     4.60 & 	0.22 & 9.71 & 66.56 \\

DenseNet121&             5.84 & 0.22 & 7.34 & 9.70 \\
DenseNet169&             7.08 & 0.24 & 8.10 & 15.75 \\
DenseNet201&             5.92 & 0.23 & 9.43 & 21.80 \\

MobileNet$\_$V2&        8.71 & 0.20 & 2.96 & \textcolor{red}{3.21} \\
MobileNet$\_$V3&        6.61 & 0.22 & \textcolor{red}{2.89} & 4.75 \\

ResNext50&              0.99 & 0.63 & 9.19 & 26.32 \\ \hline
ResNext101&             \textcolor{red}{0.29} & \textcolor{red}{0.77} & 25.19 & 90.08 \\

\hline

\end{tabular}

}
\end{table}

\vspace{-0.3cm}
\subsection{Ablation Study}

We conduct an ablation study on the Mamba block and different CNN-based encoders to show the effects of each key component.

\emph{1) Effects of the tree-aware Mamba block}:
We focus on the performance of the scanning strategy and feature similarity metric of Tree-Mamba.
The following ablation experiments are presented:
\begin{itemize}
\item \textit{w/ RS:} Using the raster scanning \cite{VMamba} instead of our tree-aware scanning. 
\item \textit{w/ CS:} Using the continuous scanning \cite{Zigma} instead of our tree-aware scanning.
\item \textit{w/ DS:} Replacing our tree-aware scanning with the diagonal scanning \cite{VmambaIR}.
\item \textit{w/ NSS:} Replacing our tree-aware scanning with the nested S-shape scanning \cite{MaIR}.
\item \textit{w/ Manhattan:} Using the Manhattan distance as the feature similarity metric.
\item \textit{full model:} Using our full tree-aware Mamba block.

\end{itemize}

The qualitative and quantitative results are shown in Fig. \ref{Ablation_SSM} and Table \ref{tab_Ablation}, respectively.
In Fig.~\ref{Ablation_SSM} (b), \textit{w/ RS} produces blurred underwater object edges and erroneous depth assignments at the image boundaries.
\textit{w/ CS} and \textit{w/ DS} alleviate the blurs of object edges, but their limited performances are shown in Figs. \ref{Ablation_SSM} (c) and (d).
\textit{w/ NSS} yields accurate depth results but lacks clear underwater object edges in Fig. \ref{Ablation_SSM} (e).
As shown in Fig. \ref{Ablation_SSM} (f), \textit{w/ Manhattan} produces clear object edges but loses depth details.
In contrast, our \textit{full model} not only achieves the best $\mathrm{A. Rel}$ and $\delta_1$ scores, but also possess the lowest computational complexity in Table \ref{tab_Ablation}.
Moreover, our depths are closer to the reference images in Fig. \ref{Ablation_SSM} (h).

\emph{2) Effects of different CNN-based encoders}:
The backbone encoder with strong feature extraction ability can improve the performance of our Tree-Mamba. 
Thus, we test a series of networks, including the ResNet family \cite{ResNet}, EfficientNet family \cite{EfficientNet}, DenseNet family \cite{DenseNet}, MobileNet family \cite{MobileNetV2, MobileNetV3}, and ResNeXt family \cite{ResNeXt}.
The qualitative and quantitative results are shown in Fig. \ref{Ablation_Encoders} and Table \ref{tab_Ablation}, respectively.
Within the ResNet family, ResNet18 has the lowest computational complexity but produces the poorest depth estimation performance, as shown in Fig.~\ref{Ablation_Encoders} (b).
As the network depth increases, ResNet101 achieves better performance at the cost of higher computational complexity, as shown in Table \ref{tab_Ablation}.
For EfficientNet and DenseNet families, EfficientNet-B2 and DenseNet169 possess a superior tradeoff between depth prediction performance and computational complexity in Table \ref{tab_Ablation}.
MobileNet-V2 and -V3 achieve the fewest parameters and lowest FLOPs, but their limited performances are significant in Figs. \ref{Ablation_Encoders} (l) and (m).
Although ResNeXt50 has fewer parameters and lower FLOPs, ResNeXt101 yields better results in depth details and object edges.
Therefore, we choose ResNeXt101 as our final encoder.

\vspace{-0.3cm}
\subsection{Computational Efficiency}

We compare the runtime of different depth estimation methods in Table \ref{tab_runtime_complexity}, and the average runtime of each method is executed one hundred times on images of different sizes. 
Traditional methods are running on an Intel Core i9-9900k CPU, where IBLA \cite{IBLA}, GDCP \cite{GDCP}, NUDCP \cite{NUDCP}, and ADPCC \cite{ADPCC} are implemented by Python 3.9, while HazeLine is implemented by Matlab R2020a.
The rest deep learning-based methods are implemented using PyTorch 2.1.0 and executed on an NVIDIA RTX 4090 GPU.
As shown in Table~\ref{tab_runtime_complexity}, compared with other competitors, our Tree-Mamba method yields the shortest running times on underwater images sized $512 \times 512$ and $1024 \times 1024$, and achieves the second-fastest speed on an underwater image sized $256 \times 256$.
These results show that our Tree-Mamba has a competitive-to-better running time on handling various sizes of underwater images while maintaining superior depth estimation performance, which can be primarily attributed to the proposed tree-aware scanning strategy that aggregates features from all channels in a single-scan manner to improve model processing speed.

\begin{table}[t]
\centering

\caption{Runtime comparison of different methods. The \textcolor{red}{best} results are marked in red.}
\label{tab_runtime_complexity}
    \addvspace{-6pt}  

\resizebox{0.9\linewidth}{!}{

\begin{tabular}{c|c|c|c|c}

\hline
\rowcolor{red!20} 
Methods  &Publication& 256×256 & 512×512 & 1024×1024 \\ 
\hline \hline

IBLA\cite{IBLA}                 &   TIP   2017 & 3.41 & 14.41 & 58.97\\
GDCP\cite{GDCP}                 &   TIP   2018 & 3.55 & 3.80 & 3.81 \\
UW-Net\cite{UW-Net}             &   ICIP  2019 & 2.61 & 6.35 & 10.60 \\
NUDCP\cite{NUDCP}               &   TOB   2020 & \textcolor{red}{0.02}& 0.10 & 0.40 \\
UW-GAN\cite{UW-GAN}             &   TIM   2021 & \textcolor{red}{0.02}& 0.05 & 0.10 \\
HazeLine\cite{HazeLine}         &   TPAMI 2021 & 2.58 & 10.88 & 40.40 \\
MiDas\cite{MiDas}               &   TPAMI 2022 & \textcolor{red}{0.02}& \textcolor{red}{0.04}& 0.09 \\
Lite-Mono\cite{Lite-Mono}       &   CVPR  2023 & 0.08 & 0.18 & 0.62 \\
UDepth\cite{UDepth}             &   ICRA  2023 & 0.08 & 0.16 & 0.54 \\
ADPCC\cite{ADPCC}               &   IJCV  2023 & 2.46& 3.13& 8.57\\
UW-Depth\cite{UW-Depth}         &   ICRA  2024 & 0.11 & 0.19 & 0.62 \\
WsUID-Net\cite{SUIM-SDA}        &   TGRS  2024 & 0.80 & 2.99 & 13.18 \\
WaterMono\cite{WaterMono}       &   TIM   2025 & 0.07 & 0.14 & 0.42 \\

\hline
Tree-Mamba                      &   -          & 0.03 & 
\textcolor{red}{0.04} & 
\textcolor{red}{0.08}\\ 
\hline
\end{tabular}

}

\end{table}

\vspace{-0.3cm}
\section{Conclusion}

We have presented a Tree-Mamba method with a tree-aware scanning strategy for UMDE.
The proposed tree-aware scanning strategy dynamically generates a minimum spanning tree based on the feature similarity to capture spatial topology structures of underwater images, and propagates multi-scale features among tree nodes in bottom-up and top-down traversals, enhancing the model's feature representation capability and facilitating our Tree-Mamba to generate fine-grained depth maps.
The proposed Tree-Mamba can estimate underwater image depths sized of 512 × 512 at 25 frames per second, thanks to our designed tree-aware Mamba blocks with linear computational complexity to avoid expensive computation.
Meanwhile, we have released a large-scale BlueDepth, which contains more precise depth labels than those of existing UMDE benchmarks \cite{SeaThru, NYU-U, HazeLine, Flsea, Atlantis, SUIM-SDA, USOD10K}.
The BlueDepth enables existing deep learning-based UMDE models to learn more accurate depth estimation capabilities in training.
Final experiments show the superior performance of our Tree-Mamba on depth prediction results of various underwater images, with particular advantages of generating fine details, holistic content, and accurate geometries.
\vspace{-0.3cm}


\vspace{-1.5cm}

\begin{IEEEbiographynophoto}{Peixian Zhuang}
is an Associate Professor at University of Science and Technology Beijing. His research interests focus on underwater image processing, sparse representation, Bayesian machine learning, and deep learning.
\end{IEEEbiographynophoto}

\vspace{-1.5cm}
\begin{IEEEbiographynophoto}{Yijian Wang}
is currently pursuing the M.S. degree in electronics and information engineering from Wenzhou Medical University, Wenzhou, China. His research interest lies in deep learning with applications in image processing and computer vision.
\end{IEEEbiographynophoto}

\vspace{-1.5cm}
\begin{IEEEbiographynophoto}{Zhenqi Fu}
is currently a Postdoctoral Researcher in the Department of Automation, Tsinghua University, Beijing, China. His current research interests include low-level vision, biomedical imaging, and deep learning.
\end{IEEEbiographynophoto}

\vspace{-1.5cm}
\begin{IEEEbiographynophoto}{Hongliang Zhang}
is a Senior Engineer at Deepinfar Ocean Technology Inc. His research areas include computer vision, automatic control and navigation, especially underwater visual analysis and measurement, as well as dynamic control and navigation for underwater robots.
\end{IEEEbiographynophoto}

\vspace{-1.5cm}
\begin{IEEEbiographynophoto}{Sam Kwong}
is the Associate Vice-President, J.K. Lee Chair Professor of Computational Intelligence, the Dean of the School of Graduate Studies and the Acting Dean of the School of Data Science of Lingnan University. His research focuses on evolutionary computation, artificial intelligence solutions, and image/video processing. 
He was elevated to IEEE Fellow in 2014 for his contributions to optimization techniques in cybernetics and video coding. He was the President of IEEE Systems, Man, and Cybernetics Society in 2021-2022. He is a Fellow of US National Academy of Inventors, Canadian Academy of Engineering, and the Hong Kong Academy of Engineering. 
\end{IEEEbiographynophoto}

\vspace{-1.5cm}
\begin{IEEEbiographynophoto}{Chongyi Li}
is a Professor and Ph.D. advisor at Nankai University. His research focuses on artificial intelligence, computer vision, and machine learning, especially computational imaging, image enhancement and restoration, image generation and editing, and multimodal large language models.
\end{IEEEbiographynophoto}

\end{document}